\RequirePackage[l2tabu, orthodox]{nag}
\documentclass[11pt, dvipsnames, DIV=12]{scrartcl}

\usepackage[english]{babel}
\usepackage{natbib}
\usepackage[utf8]{inputenc}
\usepackage[T1]{fontenc}
\usepackage{subcaption}
\usepackage{booktabs}
\usepackage{multirow}
\usepackage{url}
\usepackage{soul}
\usepackage{tikz}
\usetikzlibrary{calc}
\usetikzlibrary{decorations.pathmorphing}
\usetikzlibrary{positioning}
\usetikzlibrary{shapes}
\usetikzlibrary{arrows.meta,backgrounds,automata}
\usetikzlibrary{positioning,shapes,matrix,calc}
\tikzstyle{vertex}=[circle, draw, fill=gray!80!white,thick,scale=1.2]
\tikzstyle{edge}=[draw=black, thick,-]
\usetikzlibrary{hobby,arrows,decorations.pathmorphing,backgrounds,positioning,fit,petri,calc}
\pgfdeclarelayer{background}
\pgfsetlayers{background,main}
\usepackage{todonotes}

\usepackage{tikz}
\usepackage[edges]{forest}
\usetikzlibrary{arrows.meta,shadows.blur}
\usepackage{rotating}

\usepackage{capt-of}
\usepackage{multirow}
\usepackage{changepage,threeparttable}

\definecolor{purple}{RGB}{27, 158, 119}
\definecolor{blue}{RGB}{217, 95, 2}
\definecolor{orange}{RGB}{117, 112, 179}
\definecolor{gray}{RGB}{239,240,241}
\definecolor{pink}{RGB}{254,15,127}
\definecolor{green}{RGB}{231, 41, 138}
\definecolor{darkgreen}{rgb}{0, 0.5, 0}

\usepackage{paralist}
\usepackage[stable]{footmisc}
\usepackage{adjustbox}

\usepackage{ifthen}
\newcommand{\CC}[1][]{$\text{C\hspace{-.25ex}}^{_{_{_{++}}}}
\ifthenelse{\equal{#1}{}}{}{\text{\hspace{-.625ex}#1}}$}

\usepackage{bm}
\usepackage{bbm}
\usepackage{amsmath}
\usepackage{amssymb}
\usepackage{amsthm}
\usepackage{amsfonts}
\usepackage{thmtools}	
\usepackage{xfrac}
\usepackage{mleftright}
\usepackage{stmaryrd}
\usepackage{nicefrac}
\usepackage{algorithm}
\usepackage{algorithmicx}
\usepackage[noend]{algpseudocode}

\usepackage{dirtytalk}

\usepackage{thm-restate}

\let\originalleft\left
\let\originalright\right
\renewcommand{\left}{\mathopen{}\mathclose\bgroup\originalleft}
\renewcommand{\right}{\aftergroup\egroup\originalright}

\usepackage{pifont}

\usepackage{enumitem}
\setlist[enumerate]{itemsep=0.2ex, topsep=0.5\topsep}
\setlist[description]{itemsep=0.2ex, topsep=0.5\topsep}
\setlist[itemize]{itemsep=0.2ex, topsep=0.5\topsep}

\makeatletter
\def\thmt@refnamewithcomma #1#2#3,#4,#5\@nil{%
\@xa\def\csname\thmt@envname #1utorefname\endcsname{#3}%
\ifcsname #2refname\endcsname
\csname #2refname\expandafter\endcsname\expandafter{\thmt@envname}{#3}{#4}%
\fi
}
\makeatother

\usepackage[
pdfa,
hidelinks,
pdftex, 
pdfdisplaydoctitle,
pdfpagelabels,
pdfauthor={Christopher Morris},
pdftitle={},
pdfsubject={},
pdfkeywords={MPNNs, generalization, expressivity},
pdfproducer={Latex with the hyperref package},
pdfcreator={pdflatex}
]{hyperref}

\usepackage[capitalise,noabbrev]{cleveref}   
\theoremstyle{definition}
\newtheorem{theorem}{Theorem}

\newtheorem{proposition}[theorem]{Proposition}

\newtheorem{lemma}[theorem]{Lemma}

\newtheorem{definition}[theorem]{Definition}

\usepackage{thm-restate}
\usepackage[mathic=true]{mathtools}
\usepackage{fixmath}
\usepackage{siunitx}

\usepackage[edges]{forest}
\usetikzlibrary{arrows.meta,shadows.blur}
\usepackage{rotating}

\usepackage{microtype}
\usepackage{ellipsis}

\usepackage{anyfontsize}
\usepackage[scaled=0.86]{helvet}
\usepackage{lmodern}

\usepackage[auth-lg]{authblk}

\newcommand{\cA}{\mathcal{A}}

\newcommand{\cE}{\mathcal{E}}

\newcommand{\cG}{\mathcal{G}}
\newcommand{\cH}{\mathcal{H}}
\newcommand{\cM}{\mathcal{M}}
\newcommand{\cN}{\mathcal{N}}
\newcommand{\cO}{\mathcal{O}}

\newcommand{\cR}{\mathcal{R}}
\newcommand{\cS}{\mathcal{S}}
\newcommand{\cT}{\mathcal{T}}

\newcommand{\cV}{\mathcal{V}}
\newcommand{\cW}{\mathcal{W}}
\newcommand{\cX}{\mathcal{X}}
\newcommand{\cY}{\mathcal{Y}}
\newcommand{\cZ}{\mathcal{Z}}

\newcommand{\Bb}{\mathbb{B}}
\newcommand{\Eb}{\mathbb{E}}

\newcommand{\Nb}{\mathbb{N}}

\newcommand{\Rb}{\mathbb{R}}

\newcommand{\FNN}{\mathsf{FNN}}
\newcommand{\MPNN}{\textsf{MPNN}}

\newcommand{\wlone}{$1$\textrm{-}\textsf{WL}}

\newcommand{\conv}[1]{\textsf{conv}(#1)}

\newcommand{\hb}{\mathbold{h}}

\newcommand{\UPD}{\mathsf{UPD}}
\newcommand{\AGG}{\mathsf{AGG}}
\newcommand{\RO}{\mathsf{READOUT}}

\newcommand{\REL}{\mathsf{RELABEL}}

\newcommand{\nw}{d(2dL+L+1) + 1}

\newcommand{\new}[1]{\emph{#1}}

\newcommand*{\tran}{\intercal}

\renewcommand{\vec}[1]{\mathbold{#1}}
\newcommand{\oms}{\{\!\!\{}
\newcommand{\cms}{\}\!\!\}}
\newcommand{\tup}[1]{{(#1)}}
\DeclarePairedDelimiterX{\norm}[1]{\lVert}{\rVert}{#1}

\DeclareFontFamily{U}{mathx}{\hyphenchar\font45}
\DeclareFontShape{U}{mathx}{m}{n}{<-> mathx10}{}
\DeclareSymbolFont{mathx}{U}{mathx}{m}{n}
\DeclareMathAccent{\widebar}{0}{mathx}{"73}

\newcommand{\wmat}{\vec{W}}
\newcommand{\amat}{\vec{A}}
\newcommand{\lmat}{\vec{L}}
\newcommand{\dmat}{\vec{D}}

\usepackage[auth-lg]{authblk}

\title{\normalfont\huge \textbf{Survey on Generalization Theory for Graph Neural Networks}}

\author[1]{Antonis Vasileiou}
\author[2]{Stefanie Jegelka}
\author[3]{Ron Levie}
\author[1]{Christopher Morris}

\affil[1]{RWTH Aachen University}
\affil[2]{Technical University of Munich \& Massachusetts Institute of Technology}
\affil[3]{Technion - Israel Institute of Technology}

\date{\vspace{-30pt}}

\begin{document}

\maketitle

\begin{abstract}
	Message-passing graph neural networks (MPNNs) have emerged as the leading approach for machine learning on graphs, attracting significant attention in recent years. While a large set of works explored the expressivity of MPNNs, i.e., their ability to separate graphs and approximate functions over them, comparatively less attention has been directed toward investigating their generalization abilities, i.e., making meaningful predictions beyond the training data. Here, we systematically review the existing literature on the generalization abilities of MPNNs. We analyze the strengths and limitations of various studies in these domains, providing insights into their methodologies and findings. Furthermore, we identify potential avenues for future research, aiming to deepen our understanding of the generalization abilities of MPNNs.
\end{abstract}

\tableofcontents

\section{Introduction}

Graphs model interactions among entities across the life, natural, and formal sciences, such as atomistic systems~\citep{Duv+2023,Zha+2023b} or social networks~\citep{Eas+2010,Lov+2012}, underlining the need for machine learning methods to handle graph-structured data. Hence, neural networks designed for graph-structured data, mainly \new{message-passing graph neural networks} (MPNNs)~\citep{merkwirth05,gori05,Hamilton2017,Kip+2017,Gil+2017,Sca+2009}, have gained significant attention in the machine learning community, showcasing promising outcomes across diverse domains~\citep{Cor+2024}, spanning drug design~\citep{Won+2023}, global medium-range weather forecasting~\citep{Lam+2023}, and combinatorial optimization~\citep{Cap+2021,Gas+2019,Qia+2023}.\footnote{We use the term MPNNs to refer to graph-machine learning architectures that fit into the framework of~\citet{Gil+2017}, see~\cref{sec:MPNN}, and use the term \new{graph neural networks} (GNNs) in a broader sense, i.e., all neural network architectures capable of handling graph-structured inputs.}

While MPNNs are successful in practice and make a real-world impact, their theoretical properties are understood to a lesser extent~\citep{Mor+2024b}. Only MPNNs' \new{expressivity} is somewhat understood. Here, the expressivity of MPNNs is modeled mathematically based on two main approaches, algorithmic alignment with the~\new{$1$-dimensional Weisfeiler--Leman algorithm} (\wlone)  and universal approximation theorems~\citep{Azi+2020,Boe+2023,Che+2019,geerts2022,Mae+2019,Rac+2024}. Here, the \wlone~\citep{Wei+1968,Wei+1976,Mor+2022} is a well-studied heuristic for the graph isomorphism problem. Concretely, regarding the former,~\citet{Mor+2019,Xu+2018b} showed that the \wlone{} limits the expressivity of any possible MPNN architecture in distinguishing non-isomorphic graphs.

In contrast, less is known about MPNNs' abilities to make meaningful predictions beyond the training data. More precisely, the \new{generalization} abilities of MPNNs assess the architectures' ability to adapt effectively to new, previously unseen graphs originating from the same distribution as the training set. In addition, \new{extrapolation} or \new{out-of-distribution generalization} distinguishes itself by involving unseen graphs drawn from a (slightly) different distribution than the training set. Nowadays, there exist several analyses of MPNNs' generalization using different theoretical frameworks such as \new{Vapnik--Chervonenkis dimension}  (VC dimension)~\citep{Mor+2023,scarselli2018vapnik}, \new{Rademacher averages}~\citep{Gar+2020}, and related formalism. However, the resulting bounds are often hard to compare due to different assumptions, MPNN architectures, and parameters.

\subsection{Contributions}

Here, we survey theoretical results on MPNNs' generalization abilities, making it easier to compare different results and quickly get into the field. We survey generalization results based on VC dimension, Rademacher complexity, covering number bounds, stability-based generalization bounds, and PAC-Bayesian analysis. In addition, we also look into generalization analysis using graphon theory and generalization theory for node-level prediction tasks and out-of-distribution generalization. Our discussion covers the mathematical tools used to establish these bounds and the theoretical foundations upon which they rely. The main goal of this survey is to provide readers with a comprehensive overview of MPNNs' generalization abilities and offer insights into potential paths for extending the current theory to address gaps in the literature. In addition, we propose open problems and future challenges in the field to help spur future work.

\subsection{Related work}
As mentioned above, there exists a large set of work relating MPNNs' expressive power to the \wlone~\citep{Azi+2020,Boe+2023,geerts2022,Mar+2019c,Mor+2019, Xu+2018b} and its generalizations; see~\citet{Mor+2022} for a thorough overview. Regarding MPNNs' expressiveness and generalization, \cite{jegelka2022theory} provides a good overview of works until 2022. In addition, in their position paper, \citet{Mor+2024b} argue that the current emphasis on \wlone-based expressivity analysis is limited and outline critical challenges regarding a fine-grained analysis of expressive power, generalization, and optimization dynamics of MPNNs and related graph learning architectures.

\begin{table}[htbp]
	\centering
	\caption{Summary of results investigating  MPNNs' generalization abilities, categorized by graph-level or node-level tasks, the different frameworks employed, and the types of MPNN architectures, i.e., general MPNNs or graph convolutional neural networks.}
	\begin{tabular}{@{}lllp{4.5cm}@{}}
		\toprule
		\textbf{} & \textbf{Framework}    & \textbf{Architectures} & \textbf{References}                                                                                                                          \\
		\midrule

		\multirow{4}{*}{\rotatebox{90}{\hspace*{-55pt} \textbf{Node-level}}}
		          & Rademacher complexity & GCNs                   & \citep{DBLP:journals/corr/abs-2102-10234}                                                                                                    \\
		          & Stability-based       & GCNs                   & \citep{Ver+2019,Zhou2021}                                                                                                                    \\
		          & VC dimension          & MPNNs                  & \citep{scarselli2018vapnik}                                                                                                                  \\
		          & Transductive learning & MPNNs                  & \citep{DBLP:conf/nips/CongRM21,Ess+2021,Oono2020,Tan+2023}                                                                                   \\

		\midrule

		\multirow{8}{*}{\rotatebox{90}{\hspace*{-110pt}  \textbf{Graph-level}}}
		          & Rademacher complexity & GCNs                   & \citep{DBLP:journals/corr/abs-2102-10234}                                                                                                    \\
		          & PAC-Bayesian          & GCNs                   & \citep{Ju+2023,Lia+2021,DBLP:journals/corr/abs-2402-04038, Wu2023}                                                                           \\
		          & VC dimension          & MPNNs                  & \citep{dinverno2024vc,Fra+2024,Mor+2023,scarselli2018vapnik}                                                                                 \\
		          & Rademacher complexity & MPNNs                  & \citep{Gar+2020, Kar+2024}                                                                                                                   \\

		          & PAC-Bayesian          & MPNNs                  & \citep{Ju+2023,Lia+2021,DBLP:journals/corr/abs-2402-04038}                                                                                   \\
		          & Covering number       & MPNNs                  & \citep{                                                                                                                                      
		Lev+2023, Mas+2022, maskey2024generalization, Rac+2024, Vas+2024}                                                                                                                                         \\
		          & Out-of-distribution   & MPNNs                  & \citep{DBLP:conf/icml/BevilacquaZ021,DBLP:conf/nips/GuiLLLJ23, DBLP:journals/corr/abs-2202-07987,DBLP:conf/icml/LiGLJ24, xu2021how,Yeh+2021} \\
		\bottomrule
	\end{tabular}
	\label{tab:generalization_summary}
\end{table}

\section{Background}
\label{background}
Let $\Nb \coloneq \{ 1,2, \dots \}$ and $\Nb_0 \coloneq \Nb \cup \{ 0 \}$. The set $\Rb^+$ denotes the set of non-negative real numbers. For $n \in \Nb$, let $[n] \coloneq \{ 1, \dotsc, n \} \subset \Nb$. We use $\oms \dotsc \cms$ to denote multisets, i.e., the generalization of sets allowing for multiple, finitely many instances for each of its elements. For two non-empty sets $X$ and $Y$, let $Y^X$ denote the set of functions from $X$ to $Y$. Let $S \subset \Rb^d$, then the \new{convex hull} \conv{$S$} is the minimal convex set containing the set $S$. Let $\vec{M}$ be an $n \times m$ matrix, $n>0$ and $m>0$, over $\Rb$, then $\vec{M}_{i,\cdot}$,  $\vec{M}_{\cdot,j}$, $i \in [n]$, $j \in [m]$, are the $i$th row and $j$th column, respectively, of the matrix $\vec{M}$. For $\vec{x} \in \Rb^{1 \times d}$, we use $\|\cdot\|_1$ and $\|\cdot\|_2$  to refer to the \new{$1$-norm} $\|\vec{x}\|_1 \coloneq |x_1|+\cdots+|x_d|$ and \new{$2$-norm} $\|\vec{x}\|_2 \coloneq\sqrt{x_1^2+\cdots+x_d^2}$, respectively. For $\vec{p} \in \Rb^{1 \times d}, d > 0$, and $\varepsilon > 0$, the \new{ball} $B_{\norm{\cdot}}(\vec{p}, \varepsilon, d) \coloneq \{ \vec{s} \in \Rb^{1 \times d} \mid \lVert \vec{p} - \vec{s} \rVert \leq \varepsilon  \}$; when the norm is evident from the context, we omit the subscript. In what follows, $\vec{0}$ denotes an all-zero vector with an appropriate number of components.

\paragraph{Graphs} An \new{(undirected) graph} $G$ is a pair $(V(G),E(G))$ with \emph{finite} sets of
\new{vertices} or \new{nodes} $V(G)$ and \new{edges} $E(G) \subseteq \{ \{u,v\} \subseteq V(G) \mid u \neq v \}$. For ease of notation, we denote an edge $\{u,v\}$ in $E(G)$ by $(u,v)$ or $(v,u)$. The \new{order} of a graph $G$ is its number $|V(G)|$ of vertices. If not stated otherwise, we set $n \coloneq |V(G)|$ and call $G$ an \new{$n$-order graph}. We denote the set of all $n$-order graphs by $\cG$. For an $n$-order graph $G$, we denote its \new{adjacency matrix} by $\vec{A}(G) \in \{ 0,1 \}^{n \times n}$, where $\amat(G)_{vw} = 1$ if and only if $(v,w) \in E(G)$. We further define the Laplacian matrix $\vec{L}(G) \in \mathbb{R}^{n \times n}$ as $\lmat(G) \coloneq \dmat(G)-\vec{A}(G)$, where $\dmat(G) \in \mathbb{R}^{n \times n}$ is a degree diagonal matrix with diagonal element $D_{ii} = \sum_{j}a_{ij}$, for $i \in [n]$. The \new{neighborhood} of $v \in V(G)$ is denoted by $N(v) \coloneq  \{ u \in V(G) \mid (v, u) \in E(G) \}$ and the \new{degree} of a vertex $v$ is $|N(v)|$. A \new{(vertex-)labeled graph} is a pair $(G,\ell_G)$ with a graph $G = (V(G),E(G))$ and a (vertex-)label function $\ell_G \colon V(G) \to \Sigma$, where $\Sigma$ is an arbitrary label set. For a vertex $v \in V(G)$, $\ell_G(v)$ denotes its \new{label}. A Boolean \new{(vertex-)$d$-labeled graph} is a pair $(G,\ell_G)$ with a graph $G = (V(G),E(G))$ and a label function $\ell_G \colon V(G) \to \{0,1\}^d$. We denote the set of all $n$-order Boolean $d$-attributed graphs by $\cG_{n,d}^{\Bb}$. An \new{attributed graph} is a pair $(G,a_G)$ with a graph $G = (V(G),E(G))$ and an (vertex-)attribute function $a_G \colon V(G) \to \Rb^{1 \times d}$, for $d > 0$. The attribute or feature of $v \in V(G)$ is $a_G(v)$. We denote the class of all $n$-order graphs with $d$-dimensional, real-valued vertex features by $\cG_{n,d}^{\Rb}$. In addition, we denote the class of all graphs up to order $n$ with $d$-dimensional, real-valued vertex features by $\cG_{n,d}^{\Rb}$. The class of all graphs is denoted by $\cG$.

Two graphs $G$ and $H$ are \new{isomorphic} if there exists a bijection $\varphi \colon V(G) \to V(H)$ that preserves adjacency, i.e., $(u,v) \in E(G)$ if and only if $(\varphi(u),\varphi(v)) \in E(H)$. In the case of attributed graphs, we additionally require $\ell_G(v) = \ell_H(\varphi(v))$ for all $v \in V(G)$. Given two graphs $G$ and $H$ with disjoint vertex sets, we denote their disjoint union by $G \,\dot\cup\, H$.

\subsection{Message-passing graph neural networks}
\label{sec:MPNN}
Message-passing graph neural networks (MPNNs) are a neural network framework for graph inputs. Intuitively, MPNNs learn a vectorial representation, i.e., $d$-dimensional real-valued vector, representing each vertex in a graph by aggregating information from neighboring vertices. Formally, let $(G,\ell_G)$ be a labeled graph with initial vertex features $\hb_{v}^\tup{0} \in \Rb^{d}$ that are \emph{consistent} with $\ell$. That is, each vertex $v$ is annotated with a feature  $\hb_{v}^\tup{0} \in \Rb^{d}$ such that $\hb_{v}^\tup{0} = \hb_{u}^\tup{0}$ if and only if $\ell_G(v) = \ell_G(u)$. An example is a one-hot encoding of the labels $\ell(u)$ and $\ell(v)$. Alternatively,  $\hb_{v}^\tup{0}$ can be an attribute or a feature of the vertex $v$, e.g., physical measurements of atoms of chemical molecules. An MPNN architecture consists of a stack of neural network layers, i.e., a composition of permutation-equivariant parameterized functions. Each layer aggregates local neighborhood information, i.e., the neighbors' features around each vertex, and then passes this aggregated information on to the next layer. Following \citet{Gil+2017} and \citet{Sca+2009}, in each layer $t > 0$,  we compute vertex features
\begin{equation}
	\label{def:MPNN}
	\hb_{v}^\tup{t} \coloneq
	\UPD^\tup{t}\Bigl(\hb_{v}^\tup{t-1},\AGG^\tup{t} \bigl(\oms (\hb_v^\tup{t-1},\hb_{u}^\tup{t-1})
	\mid u\in N(v) \cms \bigr)\Bigr) \in \Rb^{d_t},
\end{equation}
for $v\in V(G)$, where $\UPD^\tup{t}$ and $\AGG^\tup{t}$ may be parameterized functions, e.g., neural networks.
In the case of graph-level tasks, e.g., graph classification, one uses a \new{readout layer}
\begin{equation}\label{def:readout}
	\hb_G \coloneq \RO\bigl( \oms \hb_{v}^{\tup{L}}\mid v\in V(G) \cms \bigr) \in \Rb^d,
\end{equation}
to compute a single vectorial representation based on learned vertex features after iteration $L$.

Finally, the output of the readout layer is passed through a parameterized function, typically a feed-forward neural network $\mathsf{FNN} \colon \Rb^d \rightarrow \Rb$, which performs the final classification or regression. The $\RO$ layer is not used for node-level tasks, and we apply the parameterized function $\mathsf{FNN}$ directly to the vertex features for predicting the node labels. The parameters of $\UPD$, $\AGG$, $\RO$, and $\mathsf{FNN}$ are optimized end-to-end, usually through a variant of \new{(stochastic) gradient-descent} (SGD), e.g., \citep{Kin+2015}, against a loss function.

\subsection{Example of MPNN layers}

Here, we define various MPNN layers used in this survey.

\paragraph{Simple MPNNs}
For a given $d$ and $L \in \Nb$, we define the class $\MPNN_{\mathsf{FNN}}(d,L)$ of simple MPNNs as $L$-layer MPNNs for which, in \cref{def:MPNN}, for each $t \in [L]$, the aggregation function $\AGG^\tup{t}$ is summation and the update function $\UPD^\tup{t}$ is a feed-forward neural network $\mathsf{FNN}^\tup{t} \colon \Rb^{2d}\to\Rb^d$ of width at most $d$. Similarly, the readout function in \cref{def:readout} consists of a feed-forward neural network $\mathsf{FNN} \colon \Rb^d\to\Rb^d$ applied on the sum of all vertex features computed in layer $L$.\footnote{For simplicity, we assume that all feature dimensions of the layers are fixed to $d \in \Nb$.} More specifically,
MPNNs in $\MPNN_{\mathsf{FNN}}(d,L)$ compute on an attributed graph $(G,\ell_G)$ with $d$-dimensional initial vertex features
$\hb_v^\tup{0} \coloneq a_G(v) \in\Rb^d$, for $v \in V(G)$, the following vertex features, for each $v\in V(G)$,
\begin{equation*}
	\hb_{v}^\tup{t} \coloneq
	\mathsf{FNN}^\tup{t}\Bigl(\hb_{v}^\tup{t-1},\sum_{u\in N(v)}\hb_{u}^\tup{t-1}\Bigr) \in \Rb^{d},
\end{equation*}
for $t \in [L]$, and
\begin{equation*}
	\hb_G \coloneq \mathsf{FNN}\Bigl(\sum_{v\in V(G)}\hb_{v}^{\tup{L}}\Bigr) \in \Rb^{d}.
\end{equation*}
The class $\MPNN_{\mathsf{FNN}}(d,L)$ encompasses the GNN architecture derived by \citet{Mor+2019} that has the same expressive power as the \wlone{} in distinguishing non-isomorphic graphs.  In addition, by $\mathsf{MPNN}(L)$, we denote the set of all $L$-layer simple MPNNs.

\paragraph{Graph convolutional networks} \new{Graph convolutional networks} (GCNs) are a special case of MPNNs~\citep{Kip+2017}. We consider an undirected graph $G \in \cG^{\Rb}_{n,d}$ with adjacency matrix $\amat$, graph Laplacian $\lmat$, and node features $\mathbf{h}_v \in \Rb^{d}$, for $v \in V(G)$. We further assume that $V(G)\coloneq[n]$, and we also add self-connections for each node, i.e., $A_{ii} = 1$, for $i \in [n]$. We define a graph filter $g(\lmat) \in \Rb^{n\times n}$ as a function of the graph Laplacian $\lmat$, where typically $g(\lmat)\coloneq\amat+\mathbf{I}_n$ with $\mathbf{I}_n$ being the $n\times n$ identity  matrix, or a Chebyshev polynomial of $\lmat$. Since the subsequent results focus solely on GCNs with a single layer and GCNs with a single hidden layer followed by an output layer with a single neuron, we exclusively present these two architectures. For the general architecture of $k$-layer GCNs, please refer to the \cref{sec:kGCNs}. In the single-layer architecture, for a vertex $v$, we compute
\begin{equation*}
	\label{def:singlelayergcn}
	\vec{h}_v \coloneq \sigma\left( \sum_{u \in N(v)} [g(\lmat)]_{\cdot u} \vec{h}_u^\tran \boldsymbol{\theta}  \right)
\end{equation*}
where $\sigma$ represents a component-wise non-linear activation function, $[g(\lmat)]_{\cdot u}$ denotes the $u$th column of the graph filter, and $\vec{\theta} \in \mathbb{R}$ is a learnable parameter. Considering the single hidden layer with an output layer with a single neuron architecture for a vertex $v$,
\begin{equation}
	\label{def:singlehiddenlayergcn}
	\vec{h}_v \coloneq \sigma \left( \sum_{t=1}^{k}w_t^{(2)}\sum_{u=1}^{n}[g(\lmat)]_{vu} \cdot \sigma \left(\sum_{l=1}^{d}w_{lt}^{(1)}\sum_{j=1}^{n}[g(\lmat)]_{uj} (\vec{h}_u)_l \right) \right),
\end{equation}
where $w_{t}^{(2)}$ and $w_{i,j}^{(1)}$ are learnable parameters and $(\textbf{x}_u)_l$ is the $l$-th entry of the vector $\textbf{x}_u$.

\subsection{Supervised learning on graphs and generalization}
\label{subsec:stats=isticallearningsetting}

In the following, we introduce the formal framework of supervised learning on graphs.

\subsubsection{The statistical learning setting}
\label{subsec:statisticallearningsetting}
In \new{supervised learning} on graphs, we are given a set of graphs, e.g., the set of graphs $\cG^{\Rb}_d$ with $d$-dimensional, real-valued vertex features, and a set of \new{labels} $\mathcal{Y}$. For example, in \new{binary classification}, the label set is typically represented by the set $\{0, 1\}$, signifying that each graph is categorized into one of two classes. In the case of \new{regression}, the label set is $\mathcal{Y} \coloneq \Rb$. The primary goal of using MPNNs for learning on graphs is to estimate a function $f \colon \mathcal{G}^{\Rb}_{d} \rightarrow \mathcal{Y}$, where we represent $f$ through an MPNN architecture and a feed-forward neural network as described in \cref{background}, i.e., $f(G) \coloneq \FNN(\hb_G)$. Such a function $f$  is commonly referred to as an \new{MPNN concept} or \new{concept}. We denote $\mathcal{H} \coloneq \{ f \colon \mathcal{G}^{\Rb}_{d} \rightarrow \mathcal{Y} \mid f \text{ is an MPNN concept} \}$ as the set of all possible MPNN concepts or the \new{hypothesis class}.

To find a suitable concept $f$, we access a finite set of \new{training points} $\cS \coloneq \{(X, y) \mid (X, y) \in \mathcal{G}^{\Rb}_{d} \times \mathcal{Y} \}$. A \new{learning algorithm} is a procedure that takes the training data as input and outputs a concept $f$. Additionally, we assume a joint probability distribution $P$ on $\mathcal{G}^{\Rb}_{d} \times \mathcal{Y}$, with training examples $(X, y) \in \cS$ sampled independently from the distribution $P$. This type of sampling is often denoted as \new{independent and identically distributed} (i.i.d.\@) sampling. Furthermore, to appropriately estimate an MPNN concept, we need a function measuring how close the predicted label is to the true label. To formally measure the goodness of our concept, we introduce a loss function $\ell \colon \mathcal{H} \times \mathcal{G}^{\Rb}_{d} \times \mathcal{Y} \rightarrow \mathbb{R}^{+}$. For example, in the case of regression, the \new{squared error}
\begin{equation*}
	\ell(f,G,y) \coloneq |f(G)-y|^2
\end{equation*}
is commonly used.  Now, for an MPNN concept $f$, and a loss function $\ell$ we define the \new{expected loss} as
\begin{equation*}
	\label{def:expected_loss}
	\ell_{\text{exp}}(f) \coloneq \Eb_{(G,y)\sim P}[\ell(f,G,y)].
\end{equation*}
While the loss function measures the error of a concept on individual data points, the \new{expected loss} is the average loss over data points generated according to distribution $P$. Therefore, a concept is better than another if it has a smaller \new{expected loss}. Hence, the objective is to minimize the \new{expected loss} to find a concept capable of accurately predicting labels of samples from the distribution $P$. However, since the distribution $P$ of the data is unknown to the learning algorithm, we rely on the \new{empirical loss} estimating the \new{expected loss}, i.e., the average loss over the training data,
\begin{equation*}
	\label{def:empirical_loss}
	\ell_{\text{emp}}(f) \coloneq \frac{1}{N}\sum_{i=1}^{n}\ell(f,X_i,y_i).
\end{equation*}
Our goal is to find a concept $f_{\cS}$ minimizing the empirical loss. Typically, the \new{empirical loss} $\ell_{\text{emp}}(f)$ for a function $f$ learned from a specific training set is relatively low, indicating its ability to capture the training data. Yet, it remains uncertain whether a function $f$ with small error on the training set also performs well across the rest of the input space $\cG$, as reflected in its expected loss $\ell_{\text{exp}}(f)$. Intuitively, an MPNN concept $f$ generalizes effectively when the difference $|\ell_{\text{exp}}(f) - \ell_{\text{emp}}(f)|$ is small. This difference is the \new{generalization gap} of the concept $f$.

\paragraph{Node-level prediction} Defining the generalization gap for node-level prediction tasks bears some issues. In this setting, as described in \cref{sec:MPNN}, we do not use a readout layer, and instead, we directly apply a feed-forward neural network to a node's output feature vector to predict the node's label from $\mathcal{Y}$. Therefore, defining the empirical loss using a loss function $\ell$ for node-level tasks, e.g., classification or regression, is straightforward. However, for the expected loss, it is necessary to define the space on which we assume a distribution under which the expectation in \cref{def:expected_loss} is computed. Different approaches have been proposed in the literature. For example, \citet{scarselli2018vapnik} described a framework where the underlying distribution $P$ is used for i.i.d.\@ sampling triplets $(G,u,y)$ consisting of a graph $G \in \cG_d$, a node $u \in V(G)$, and a label $y \in \mathcal{Y}$. Alternatively, in the framework developed by \citet{Ver+2019}, considering a fixed graph $G \in \cG^{\Rb}_d$, the underlying distribution $P$ is used for i.i.d.\@ sampling a pair $(v,y)$ consisting of a node $v \in G$ and a label $y \in \mathcal{Y}$. It is worth noting that in the node classification setting, while nodes are often assumed to be sampled in an i.i.d.\@ fashion from $V(G)$, this assumption is neither widely accepted nor particularly intuitive. Consequently, several works have avoided this assumption, instead adopting a transductive learning approach to define the generalization gap, as described in \cref{sec:transduction}.

\paragraph{Generalization and concentration inequalities}
At first glance, the difference we aim to bound, i.e., $|\ell_{\text{emp}}(f) - \ell_{\text{exp}}(f)|$, appears to be the deviation of a random variable $ \ell_{\text{emp}}(f) $ from its expectation $ ell_{\text{exp}}(f) $. Since the observations are i.i.d.\@, one might initially conjecture that a standard \emph{concentration inequality} (e.g., Hoeffding’s inequality, see \citep{Hoeffding1963}) could be applied to bound this difference. Specifically, for a fixed function $f$ with an appropriate domain and under mild assumptions, a simple variant of Hoeffding’s inequality provides an upper bound (with high probability) on the deviation $ \big| \sum_{i=1}^{n} f(X_i) - \Eb[f(X_1)] \big| $, where $ X_i $ are i.i.d.\@ random variables.  However, if we look more carefully, we observe that in $ \ell_{\text{emp}}(f) $, the terms in the summation are not independent. This comes from the fact that the function $f$ is selected based on the training sample, thereby introducing dependence between $f$ and $\cS$. Consequently, Hoeffding’s inequality is not directly applicable. The main two approaches in generalization theory either employ concentration inequalities to derive bounds that hold uniformly over all functions $f$ in a hypothesis class, e.g., VC bounds (see \cref{sec:VCbounds}) or Rademacher complexity bounds (\cref{rademacherbounds}), or endow the hypothesis class with probability distributions and derive bounds for the expected difference concerning the induced distribution (PAC-Bayesian approach, see \cref{sec:pacbayesian}).

\section{MPNN generalization bounds via VC dimension}
\label{sec:VCbounds}

The VC dimension, introduced in \cite{Vap+1964}, is one example for capturing
the capacity of the hypothesis class $\cH$ for binary classification problems. The key contribution of VC dimension theory~\citep{Vap+1964,Vap+95,Vap+1998} lies in bounding the generalization gap by the worst-case behavior across all MPNN concepts $f$ in a hypothesis class $\mathcal{H}$ that the learning algorithm could select, noting that the actual trained network must be one of these MPNNs. In other words, instead of bounding the difference $|\ell_{\text{emp}}(f) - \ell_{\text{exp}}(f)|$ for the trained classifier $f=f_{\cS}$, the VC dimension aims to bound the supremum
\begin{equation*}
	\label{eq:uniform}
	\sup_{f \in \mathcal{H}} |\ell_{\text{emp}}(f) - \ell_{\text{exp}}(f)|.
\end{equation*}
In the following, we introduce the VC dimension tailored to graphs and MPNNs. Let us consider a binary classification problem over the graph class $\cG^{\Rb}_d$, for $d \in \mathbb{N}$, and a hypothesis class $\cH$ of MPNN concepts from $\cG^{\Rb}_d$ to $\{0,1\}$. We say that the graphs $\{ G_1,\dotsc,G_N \} \subseteq  \cG^{\Rb}_d$ are shattered by the hypothesis class $\cH$ if for any $\mathbold{\tau} \in \{0,1\}^N$ there exists a classifier $f \in \cH$ such that for all $i \in [N]$,
\begin{equation*}
	f(X_i)=
	\begin{cases}
		0, & \text{if $\tau_i=1$,} \text{ and } \\
		1, & \text{if $\tau_i=1$.}
	\end{cases}
\end{equation*}
For a set $\mathcal{X}$ of graphs and a class of MPNN concepts $\cH$, the VC dimension, denoted by $\text{VC}_{\mathcal{X}}(\cH)$, is the maximal number $N$ of graphs $G_1,\dotsc,G_N \subseteq \mathcal{X}$ that can be shattered by $\cH$. The idea behind the VC dimension is that although $\cH$ may contain infinitely many functions, the different ways it can classify a training set of $N$ sample graphs is finite. Namely, for any given graph in the training sample, a Boolean function can only take the values $0$ or $1$. On a sample of $N$ graphs $G_1, \dotsc, G_N$, a function can act in at most $2^N$ different ways. \citet{Vap+1964,Vap+1998} showed the following bound for the expected error for the $0$-$1$ loss function.
\begin{theorem}[{\cite{Vap+1964,Vap+1998}, adapted to MPNNs}]
	\label{VCtheorem}
	Let $\cH$ be a hypothesis class of MPNNs over the set of graphs $\cX$, with $VC_{\mathcal{X}}(\cH)=d<\infty$. Then, for all $\delta \in (0,1)$, with probability at least $1-\delta$, the following holds for all MPNN $f\in\cH$,
	$$
		\ell_{\text{exp}}(f) \leq \ell_{\text{emp}}(f) + \sqrt{\frac{2d \log({\sfrac{eN}{d})}}{N}} + \sqrt{\frac{\log({\sfrac{1}{\delta})}}{2N}}.
	$$
\end{theorem}
The proof of the above result follows the following process. First, we use the so-called symmetrization lemma (see \citep[Lemma 2]{DBLP:conf/ac/BousquetBL03}) to bound $\sup_{f \in \mathcal{H}} |\ell_{\text{emp}}(f) - \ell_{\text{exp}}(f)|$ by a sum of finite terms, utilizing the fact that any hypothesis class $\cH$ can only act in at most $2^N$ different ways on an $N$-size sample. Then, we apply Hoeffding’s inequality \citep{DBLP:conf/ac/BoucheronLB03} to each term of the finite sum that arises.

It is worth mentioning that \cref{VCtheorem} generally holds for any input data space $\cX$ beyond graph data and any hypothesis class $\cH$. Hence, bounding the VC dimension of a hypothesis class of Boolean concepts directly implies an upper bound on the generalization gap. Note that the analysis is only meaningful in cases where the number of samples is much larger than the VC dimension.

\subsection{Bounding the VC dimension of MPNNs}

In one of the earliest studies aiming to bound the VC dimension of MPNNs, \cite{scarselli2018vapnik}, based on~\citet{Kar+1997}, derived bounds for the VC dimension for a specific class of MPNNs. They examined the influence of various factors, e.g., the number of parameters and the maximum number of vertices in the input graph, based on the choice of the activation function for the utilized feed-forward network, using the framework of Pfaffian function theory~\citep{Kar+1997}. It is worth noting that the bounds presented in~\citet{scarselli2018vapnik} closely resemble those established for recurrent neural networks. This similarity was not initially expected, given the complexity of MPNN architectures, which could have potentially affected the VC dimension differently. In more recent work, \citet{dinverno2024vc}, using similar assumptions on the activation functions and following similar proof techniques as \cite{scarselli2018vapnik}, extended the analysis of the VC dimension to modern MPNNs. Since \cite{scarselli2018vapnik} established their bounds for the earliest MPNN model~\citep{Sca+2009}, the architecture and notation in \citep{scarselli2018vapnik} slightly differs from the one introduced in \cref{sec:MPNN}. Therefore, we choose to present the results from \cite{dinverno2024vc}, adapted to the recent and most commonly used MPNN architecture described in~\cref{def:MPNN,def:readout}.

Specifically,~\citet{dinverno2024vc} provided upper bounds for the VC dimension of modern MPNN architectures under the assumption of Pfaffian activation functions; see~\cref{sec:pfaff}. The bounds depend on the total number of parameters $p$, the number of layers $L$, the feature dimension $d$, the size $q$ of the hidden vector representation of each node $v$ after the $L$th layer, and the maximal order $n$ of the graphs.

While Theorem 1 in \citet{dinverno2024vc} provides the analytical expression of the VC bound for MPNNs, we present the following result showing the dependence of the VC dimension on different MPNN architectures and graph parameters. The following result provides bounds for a restricted class $\cH$ of MPNNs, where the functions $\UPD$, $\AGG$, and $\RO$ are Pfaffian functions, and their number of parameters for each layer is in $\mathcal{O}(d)$.
\begin{theorem}
	\label{scarseli24}
	Let $\mathcal{F}$ be the hypothesis class described above. Then,
	\begin{align*}
		 & \text{VC}_{\mathcal{G}^{\Rb}_{\leq n,d}}(\cH) \leq  \mathcal{O}(p^{-4}), \\
		 & \text{VC}_{\mathcal{G}^{\Rb}_{\leq n,d}}(\cH) \leq \mathcal{O}(n^2),     \\
		 & \text{VC}_{\mathcal{G}^{\Rb}_{\leq n,d}}(\cH) \leq \mathcal{O}(L^4),     \\
		 & \text{VC}_{\mathcal{G}^{\Rb}_{\leq n,d}}(\cH) \leq \mathcal{O}(d^6),     \\
		 & \text{VC}_{\mathcal{G}^{\Rb}_{\leq n,d}}(\cH) \leq \mathcal{O}(q^2),
	\end{align*}
	where $p$ is the total number of parameters, $L$ is the depth of MPNNs, $q$ the dimension of the output representation vector, and $d$ the node feature dimension.
\end{theorem}

Furthermore, \citet{dinverno2024vc} established alternative bounds for the VC dimension of MPNNs based on the color count produced by the $\wlone$ algorithm. However, we choose not to delve into these findings as we will explore the exact correlation of the \wlone{} algorithm, see~\cref{subsec:1WL}, and the VC dimension in the following section.

\subsection{Connecting the \wlone{} and VC dimension bounds for MPNNs}

\citet{Mor+2023} also explored bounds for the VC dimension. What is particularly intriguing about their work is the connection they establish between the VC dimension of a subclass of MPNNs and the \wlone; see~\cref{subsec:1WL}. While it was previously known~\citep{Morris2019,Xu+2018b} that the \wlone{} is closely linked with the expressivity of MPNNs, little was understood about its relationship with the generalization capabilities of MPNNs. Their study is divided into deriving upper bounds for the VC dimension of MPNNs under two scenarios. One where an upper bound on the graph's order is known (\new{non-uniform}), and one without this assumption (\new{uniform}). In the former case, leveraging the equivalence between MPNNs and  \wlone~\citep{Morris2019,Xu+2018b}, they demonstrate that the VC dimension of MPNNs with $L$ layers equals the maximum number of graphs distinguishable in $L$ iterations of \wlone. By doing so, they established the \emph{first lower bound} for the VC dimension of MPNNs. It is essential to note that this result was established only for the case with Boolean vertex features. For scenarios where the graph's order is unbounded, they establish that the number of parameters, specifically the bit length of MPNNs' weights, tightly bounds their VC dimension; see also~\citep{Daniels+2024} for a refined result. Additionally, they provide an upper bound for MPNNs' VC dimension using the number of colors produced by the \wlone{} for each graph. We summarize the above results in the next theorems.

Concretely, they consider the following slightly modified definition of VC dimension. For a hypothesis class $\cH$ of MPNN concepts and $\cX \subset \cG$ of graphs, $\text{VC}_\cX(\cH)$ is the maximal number $N$ of graphs $G_1,\dotsc, G_N$ in $\cX$ that
can be shattered by $\cH$. Here, $G_1,\dotsc,G_N$ are \new{shattered} if for any $\mathbold{\tau} $ in $\{0,1\}^N$ there exists an MPNN concept $f \in \cH$ such that for all $i$ in $[N]$,
\begin{equation*}\label{threshold}
	f(G_i)
	\begin{cases}
		\geq 2/3 & \text{if $\tau_i=1$, and} \\
		\leq 1/3 & \text{if $\tau_i=0$.}
	\end{cases}
\end{equation*}

Note that the above two thresholds are arbitrary, and the below results also work with other constants. \citet{Mor+2023} first consider the VC dimension of MPNN concepts on the class $\cG_{n,d}^\mathbb{B}$ consisting of graphs of an order of at most $n$ with $d$-dimensional Boolean features. Let $m_{n,d,L}$ be the maximal number of graphs in $\cG_{\leq n,d}^\mathbb{B}$ distinguishable by \wlone{} after $L$ iterations. Then, $m_{n,d,L}$ is also the maximal number of graphs in $\cG_{\leq n,dn,d}^\mathbb{B}$  that can be shattered by $L$-layer MPNN concepts $\mathsf{MPNN}(L)$, as is stated next.
\begin{proposition}\label{thm:colorbound_up}
	For all $n$, $d$ and $L$, it holds that
	\begin{equation*}
		\text{VC}_{\cG_{\leq n,d}^{\mathbb{B}}}\bigl(\mathsf{MPNN}(L)\bigr) \leq m_{n,d,L}.
	\end{equation*}
\end{proposition}
This upper bound holds regardless of the choice of aggregation, update, and readout functions used in the MPNNs. They next show a matching lower bound for the VC dimension of MPNNs on graphs in $\cG_{\leq n,d n,d}^{\mathbb{B}}$.
\begin{proposition}\label{prop:matchingvc}
	For all $n$, $d$, and $L$,  all $m_{n,d,L}$ \wlone-distinguishable graphs of order at most $n$ with $d$-dimensional Boolean features can be shattered by sufficiently
	wide $L$-layer MPNN concepts. Hence,
	\begin{equation*}
		\text{VC}_{\cG_{\leq n,d}^\mathbb{B}}\!\!\bigl(\mathsf{MPNN}(L)\bigr)=m_{n,d,L}.
	\end{equation*}
\end{proposition}
The lower bound follows from the fact that MPNNs are as powerful as the \wlone{} \citep{Mor+2019,Xu+2018}.

In the uniform case,~\citet{Mor+2023} considered the class of graphs $\cG^{\Rb}_{d,\leq u}$ having $d$-dimensional real vertex features and at most $u$ colors under the \wlone, resulting in the following result.
\begin{theorem}\label{thm:bartlett}
	For $d$ and $L \in \Nb$, and MPNN concepts in  $\mathsf{MPNN}_{\mathsf{FNN}}(d,L)$, assume we are using piece-wise polynomial activation functions with $p > 0$ pieces and degree $\delta \geq 0$. Let $P = \nw$ be the number of parameters in the MPNNs. For all $u$ in $\Nb$,
	\begin{equation*}
		\text{VC}_{\cG_{d,\leq u}}(\MPNN_{\mathsf{FNN}}(d,L))\leq
		\begin{cases}
			\cO(P\log(puP))                   & \text{if $\delta=0$,} \\
			\cO(LP\log(puP))                  & \text{if $\delta=1$,} \\
			\cO(LP\log(puP)+L^2P\log(\delta)) & \text{if $\delta>1$}. \\
		\end{cases}
	\end{equation*}
\end{theorem}
The dependence on $u$ in these upper bounds cannot be improved by more than a constant factor.

\subsection{Margin-based VC bounds for MPNNs}

\cite{Fra+2024} investigated the VC dimension of MPNNs and graph kernels based on the \wlone{} from a data margin perspective, building on the theory of VC dimension of partial concepts outlined in~\cite{Alo+2021}. Formally, let $\cX$ be a non-empty set, following~\citet{Alo+2021}, they consider \new{partial concepts} $\mathcal{H} \subseteq \{0,1,\star\}^{\cX}$, where each concept $c \in \mathcal{H}$ is a \emph{partial} function. That is, if $x \in \cX$ such that $c(x) = \star$, then $c$ is \new{undefined} at $x$. The \new{support} of a partial concept $h \in \mathcal{H}$ is the set
\begin{equation*}
	\mathsf{supp}(h) \coloneq \{ x \in \cX \mid h(x) \neq \star \}.
\end{equation*}
\citet{Alo+2021} showed that the VC dimension of (total) concepts straightforwardly generalizes to partial concepts. As with the original definition of VC dimension, the VC dimension of a partial concept class $\mathcal{H}$ over a set of $\cX$, denoted $\text{VC}_{\cX}(\mathcal{H})$, is the maximum cardinality of a shattered set $U \coloneq \{ x_1, \dots, x_N \} \subseteq \cX$. Here, the set $U$ is \new{shattered} if for any $\pmb\tau \in \{0,1\}^N$ there exists $f \in \mathcal{H}$ such that for all $i\in[N]$,
\begin{equation*}
	f(x_i) = \tau_i.
\end{equation*}
In essence,~\citet{Alo+2021} showed that the standard definition of PAC learnability extends to partial concepts, recovering the equivalence of finite VC dimension and PAC learnability.

Following~\citet{Alo+2021}, a \new{sample} $(\vec{x}_1, y_1), \dotsc, (\vec{x}_N, y_N) \in \Rb^{d} \times \{ 0,1 \}$, for $d > 0$, is \emph{$(r,\lambda)$-separable} if (1) there exists $\vec{p} \in \Rb^d$, $r > 0$, and a ball $B(\vec{p}, r)$ such that $\vec{x}_1, \dotsc, \vec{x}_N \in B(\vec{p}, r)$ and (2) the Euclidean distance between $\conv{\{ \vec{x}_i \mid y_i = 0  \}}$ and $\conv{\{ \vec{x}_i \mid y_i = 1 \}}$ is at least $2\lambda$. Then, the sample $S$ is \new{linearly separable} with \new{margin} $\lambda$. We define the set of concepts
\begin{align*}
	\cH_{r,\lambda}(\Rb^d) \coloneq \Big\{ h \in \{0,1,\star\}^{\mathbb{R}^d} \mathrel{\Big|}\, \forall\, & \vec{x}_1, \dots, \vec{x}_N \in \mathsf{supp}(h) \colon                                                \\
	(                                                                                                     & \vec{x}_1, h(\vec{x}_1)), \dotsc, (\vec{x}_N, h(\vec{x}_N)) \text{ is $(r,\lambda)$-separable} \Big\}.
\end{align*}
\citet{Alo+2021} showed that the VC dimension of the concept class $\cH_{r,\lambda}(\Rb^d)$ is asymptotically lower- and upper-bounded by $\nicefrac{r^2}{\lambda^2}$. Importantly, the above bounds are independent of the dimension $d$, while standard VC dimension bounds scale linearly with $d$~\citep{Ant+2002}.

\citet{Fra+2024} lifted the result to MPNNs. Assuming a fixed but arbitrary number of layers $T \geq 0$, a number of vertices $n > 0$, and an feature dimension $d>0$, let
\begin{equation*}
	\cE_{\textsf{MPNN}}(n,d,T) \coloneq \{ G\mapsto f(G)   \mid G \in \cG_n \text{ and } f \in  \MPNN_{\mathsf{FNN}}(d,T) \},
\end{equation*}
i.e., the set of $d$-dimensional vectors computable by simple $T$-layer MPNNs over the set of $n$-order graphs. A (graph) sample $(G_1,y_1),\dotsc,(G_N,y_N) \in \cG_{n}\times\{0,1\}$ is \new{$(r,\lambda)$-$\cE_{\textsf{MPNN}}(n,d,T)$-separable} if there is an MPNN $f \in \cE_{\textsf{MPNN}}(n,d,T)$ such that $(f(G_1),y_1),\dotsc,(f(G_N),y_N)\in \Rb^d\times\{0,1 \}$ is $(r,\lambda)$-separable, resulting in the set of partial concepts
\begin{align*}
	\cH_{r,\lambda}(\cE_{\textsf{MPNN}}(n,d,T)) \coloneq \Big\{
	h\in \{ & 0,1,\star\}^{\cG_n} \mathrel{\Big|}\,\forall\, G_1,\dotsc,G_N\in\mathsf{supp}(h)\colon \\
	(       & G_1,h(G_1)),\dotsc,(G_N,h(G_N))
	\text{ is $(r,\lambda)$-$\cE_{\textsf{MPNN}}(n,d,T)$-separable} \Big\}.
\end{align*}
Based on this,~\citet{Fra+2024} showed that the VC dimension of MPNNs can be tightly bounded by the margin of linearly separable data, resulting in the following theorem.
\begin{theorem}\label{prop:matchingvc_mpnn}
	For any $n, T>0$ and sufficiently large $d>0$, we have,
	\begin{align*}
		 & \text{VC}(\mathcal{H})_{\sqrt{T+1}n,\lambda}(\cE_{\textsf{MPNN}}(n,d,T))) \in {\Theta}(\nicefrac{r^2}{\lambda^2}), \text{ for } r=\sqrt{T}n  \text{ and } n\geq \nicefrac{r^2}{\lambda^2}.
	\end{align*}
\end{theorem}
Moreover, they also showed the same bound for more expressive MPNNs incorporating subgraph information, following~\citet{Bou+2020}. In addition, they derived conditions for when such more expressive architectures lead to a larger margin, resulting in improved generalization performance.

\section{Rademacher complexity bounds}
\label{rademacherbounds}
Next, we introduce \new{Rademacher complexity}, see~\citet{MohriRostamizadehTalwalkar18}, a concept similar to VC dimension, measuring the capacity of a concept class. The fundamental distinction is that while the VC dimension remains independent of the data distribution, the Rademacher complexity relies on the underlying data distribution. Formally, let $\sigma_1, \dotsc, \sigma_N$ be independent random variables taking values of $-1$ and $1$ with equal probabilities of 0.5 each (i.e., Rademacher random variables). We define the Rademacher complexity $\mathcal{R}(\cH)$ of a hypothesis class $\cH$ consisting of binary classification concepts, i.e., mappings from $\cX\to\{-1,+1\}$, as
\begin{equation*}
	\mathcal{R}(\cH) \coloneq \Eb_{\sigma,P_X}\left[ \sup_{f \in \cH} \left( \frac{1}{n} \sum_{i=1}^{n} \sigma_i f(X_i) \right) \right].
\end{equation*}
Here, $\sigma$ are the Rademacher random variables, and $P_X$ is the marginal distribution of the initial underlying data distribution $P$, as described in \cref{subsec:statisticallearningsetting}. To understand the above expression, consider the values of $\sigma_i$ as fixed, where each $\sigma_i$ represents a label for the data point $X_i$. Since both $\sigma_i$ and the concept $f(X_i)$ take values of +1 or -1, their product $\sigma_i f(X_i)$ yields +1 if $\sigma_i$ equals $f(X_i)$, and -1 otherwise. Consequently, the summation becomes large when the predicted labels $f(X_i)$ match the $\sigma_i$ labels across many data points. Hence, the function $f$ fits well with the $\sigma_i$ labels. Considering the supremum, we examine not just one function $f$ but all functions $f \in \cH$. The supremum becomes large if a function in $\cH$ fits the given sequence of $\sigma_i$ labels effectively. As $\sigma_i$ represent random variables, they serve as random labels for the data points $X_i$. Thus, we assign a higher Rademacher complexity value to the class $\cH$ if it can effectively adapt to random labels, aligning with the understanding that a function space needs to be sufficiently large to accommodate diverse random labels across varied datasets.

Finally, it is worth mentioning that in the definition of the Rademacher complexity, the expectation is computed regarding both Rademacher random variables and the data marginal distribution $P_X$. This makes it non-computable since the underlying distribution $P$ is unknown. Thereto, we introduce the empirical Rademacher complexity to provide generalization bounds. The empirical Rademacher complexity differs because the expectation is computed only regarding the Rademacher random variables $\sigma_i$, while for the $X_i$, we use the training set. Therefore, the \new{empirical Rademacher complexity}
\begin{equation*}
	\hat{\mathcal{R}}(\cH) \coloneq \Eb_{\sigma}\left[ \sup_{f \in \cH} \left( \frac{1}{N} \sum_{i=1}^{N} \sigma_i f(X_i) \right) \right].
\end{equation*}
From a mathematical perspective, the (empirical) Rademacher complexity can be employed to bound the generalization error, utilizing the following result~\citep{MohriRostamizadehTalwalkar18}.
\begin{theorem}[\citep{DBLP:conf/ac/BousquetBL03}[Theorem 5]]
	\label{thm:rademachercomplexitybounds}
	Let $\cH$ be a hypothesis class mapping from a set of graphs $\cG$ (e.g., $\cG_{n,d}^{\Rb}$ ) to $\{ 0,1 \}$. Then, for any $\delta>0$, with probability at least $1-\delta$, over the draw of an i.i.d.\ sample of size $N$, the following holds:
	\begin{equation*}
		\ell_{\text{exp}}(f) \leq \ell_{\text{emp}}(f) + 2\mathcal{R}(\cH) + \sqrt{\frac{\log(1/\delta)}{2N}}.
	\end{equation*}
	and also, with probability at least $1-\delta$,
	\begin{equation*}
		\ell_{\text{exp}}(f) \leq \ell_{\text{emp}}(f) + 2\hat{\mathcal{R}}(\cH) + \sqrt{\frac{2\log(2/\delta)}{N}}.
	\end{equation*}
\end{theorem}
The important part of the above result is the second inequality, showing that one can obtain a bound that depends solely on the data. The proof of the above theorem again utilizes the symmetrization lemma and another concentration inequality (McDiarmid's inequality, see \citep{McDiarmid1989}) to bound the difference between the Rademacher complexity and the empirical Rademacher complexity.

\subsection{Rademacher complexity bounds for MPNNs}

\citet{Gar+2020} derived data-dependent bounds based on bounding the Rademacher complexity, tailored specifically to MPNN architectures, extending results from~\citet{Bar+2017} for feed-forward neural networks.  Their findings show that these bounds are significantly tighter compared to VC dimension-based bounds for MPNNs outlined in \citet{scarselli2018vapnik}, and they resemble Rademacher bounds for recurrent neural networks~\citep{Chen2020}.

They consider the following popular MPNN architecture \citep{Dai+2016,Jin2019} for computing vertex features. Let $G$ be a graph, for $v \in V(G)$ with initial feature vector $\hb_v^\tup{0} \in \mathbb{R}^d$, at layer $l > 0$, we set
\begin{equation*}
	\hb_{v}^\tup{l} \coloneq \phi \Bigl(   \hb_{v}^\tup{l-1} \wmat_1 +  \rho \Bigl( \sum_{u \in N(v)} g \left(\hb_{u}^\tup{l-1} \right) \Bigl) \wmat_2 \Bigr) \in \mathbb{R}^{1 \times d},
\end{equation*}
where $\phi$, $\rho$, and $g$ are pointwise nonlinear activation functions and $\wmat_1,\wmat_2 \in \Rb^{d\times d}$ are learnable parameters of the MPNN. They further assume that $\norm{\hb_v^\tup{0}}_2 \leq B$ for $v \in V(G)$, and that $\phi$ is bounded by a constant $b$. For example, $\phi$ could be the hyperbolic tangent function. Furthermore, they assume that $\phi$, $\rho$, and $g$ are Lipschitz continuous functions with constants $C_{\phi}$, $C_{\rho}$, and $C_{\text{g}}$, respectively, and that the weights $\wmat_1$ and $\wmat_2$ have bounded norms, i.e., $ \norm{\wmat_i}_2 \leq B_i$, for $i \in [2]$. Note that in this architecture, the weights $\wmat_1$ and $\wmat_2$ and the functions $\phi$, $\rho$, and $g$ are shared across all vertices and layers. Clearly, the above architecture is a special case of the general form of MPNNs described in~\Cref{def:MPNN}.

The final step in the MPNN computation involves the readout layer, which generates a single feature vector $\hb_G$ for the entire graph after the $L$th layer as in \Cref{def:readout}, namely,
\begin{equation*}
	f_c(\hb_{v}^\tup{L}) \coloneq \sigma \mleft(\mathbold{\beta}^\tran\hb_{v}^\tup{L} \mright), \text{ for all } v \in V(G), \quad \text{and} \quad \hb_G \coloneq \sum_{v \in V}  f_c \mleft(\hb_{v}^\tup{L} \mright),
\end{equation*}
where $\hb_{v}^\tup{L}$ is the feature embedding of node $v$ in the last layer $L$, $\mathbold{\beta} \in \Rb^{d_L}$ is a parameter of the readout function with bounded norm ($\norm{\mathbold{\beta}}_2 < B_{\mathbold{\beta}}$), and $\sigma$ denotes the sigmoid function. Since we are considering a binary classification problem, we can, without loss of generality, consider labels $\{0,1\}$ instead of $\{-1,+1\}$. The loss function for a graph $G$ with true label $y$ and a concept $f \colon \cG \rightarrow \{0,1\}$ is given by the margin $\gamma$ loss function,
\begin{equation*}
	\text{loss}_{\gamma}(\alpha) \coloneq \mathbf{1}[a>0] + (1+\alpha/\gamma)\mathbf{1}[\alpha \in [-\gamma,0]],
\end{equation*}
where, $\mathbf{1}_A$ denotes the indicator function of an event A, $\alpha$ is defined through the population risk $p(f(G),y) \coloneq y(2f(G)-1) + (y-1)(2f(G)-1)$ as $\alpha = -p(f(G),y)$. The margin loss function evaluates the classifier's output by considering whether the prediction is correct and how confident the prediction is. A larger margin indicates higher confidence in the correct classification. See \citet{Lin2004} for a detailed analysis of margin loss functions.

The following result shows the relation between the population risk $\mathbb{P}[p(f(G),y)\leq0]$, the empirical risk $\ell_{\text{emp}}(f) = \frac{1}{N}\sum_{j=1}^{N}\text{loss}_{\gamma}(-p(f(G_i),y_i))$ for the training set $\{ (G_i,y_i) \}_{i=1}^{N}$ and the empiricial Rademacher complexity.

\begin{lemma}[\citep{MohriRostamizadehTalwalkar18}]
	For any margin $\gamma >0$,  any prediction function $f$ in a hypothesis class $\cH$ and $\mathcal{J}_{\gamma} \in \{ (G,y) \mapsto \text{loss}_{\gamma}(-p(f(G),y)) \mid f \in \cH \}$, given $N$ samples $(G_j,y_j)$ sampled i.i.d.\@ from $P$, with probability at least $1-\delta$, the population risk for $P$ and $f$ is bounded as,
	\begin{equation*}
		\mathbb{P}[p(f(G),y)\leq0] \leq \ell_{\text{emp}}(f) + 2\hat{\mathcal{R}}(\mathcal{J}_{\gamma}) +3\sqrt{\frac{\log(2/\delta)}{2N}},
	\end{equation*}
	where $\hat{\mathcal{R}}(\mathcal{J}_{\gamma})$ is the empirical Rademacher complexity of the hypothesis class $\mathcal{J}_{\gamma}$.
\end{lemma}
Therefore, to bound the population risk, \citet{Gar+2020} derived bounds for the empirical Rademacher complexity. These bounds were derived in two steps. Firstly, they demonstrated that bounding the empirical Rademacher complexity can be achieved by bounding the Rademacher complexity of vertex-wise computation trees. Then, they bounded the complexity of a single computation tree via recursive spectral bounds, resulting in the following proposition.

\begin{proposition}
	If the maximum degree for any node is at most $\tilde{d}$, then the empirical Rademacher complexity $\hat{\mathcal{R}}(\mathcal{J}_{\gamma})$, is bounded by,
	\begin{align*}
		\hat{\mathcal{R}}(\mathcal{J}_{\gamma}) & \leq \frac{4}{\gamma N} + \frac{24dB_{\mathbold{\beta}}Z}{\gamma \sqrt{N}}\sqrt{3 \log{Q}}, \text{ where } \\
		                                        & Q \coloneq 24B_{\mathbold{\beta}}\sqrt{N}\max\{ Z,M\sqrt{d}\max\{ B_hB_1,\bar{R}B_2 \}  \},                \\& M \coloneq C_{\phi} \frac{(C\tilde{d})^L-1}{C\tilde{d}-1}, \ Z\coloneq C_{\phi}B_1B_h + C_{\phi}B_2 \bar{R}, \\
		                                        & C \coloneq C_{\rho} C_g C_{\phi}B_2,                                                                       \\
		                                        & \bar{R} \leq C_{\rho}C_g \tilde{d} \min \{ b\sqrt{d},B_1B_h M \}.
	\end{align*}
\end{proposition}

The most important parameters in the above result are the dimension of the feature space $d$, the size $N$ of the training data, and the number of layers $L$. All other constants in the bound stem from the model's parameters and the Lipschitz constants of the non-linear activation functions. These bounds exhibit the same dependence on dimension $d$, depth $L$, and sample size $N$ as the generalization bounds established for recurrent neural networks~\citep{Chen2020}. Additionally, the VC bounds established by \citet{scarselli2018vapnik} for the setting with tanh and sigmoid activation functions depend on the fourth order regarding the number of hidden units and quadratic in dimension $d$ and the maximum number of nodes $n$ in the input graph. However, in the above setting, the VC dimension scales as $\mathcal{O}(d^6n^2)$ and consequently, the generalization error as $\mathcal{O}(d^3n/\sqrt{N})$. Thus, the bounds established by \citet{Gar+2020} are significantly tighter. Recently,~\citet{Kar+2024} extended the above analysis to equivariant MPNNs.

\paragraph{Rademacher complexity bounds for GCNs}
\citet{DBLP:journals/corr/abs-2102-10234} studied the generalization power of single-layer GCN, providing tight upper bounds of Rademacher complexity for GCN models with a single hidden layer. Since these results refer to node-level tasks, as will be clarified, to define the Rademacher complexity in this work, it is assumed that pairs of vertices and labels have been sampled in an i.i.d.\@ fashion according to some underlying distribution $P$. The Rademacher bounds derived in this work explicitly depend on the largest eigenvalue of the graph (convolution) filter and the degree distribution of the graph. Another important difference from the work by \citet{Gar+2020} is that only the vertex-focused task involving a fixed adjacency matrix is considered in this study. Therefore, the data is no longer subject to the i.i.d.\@ assumptions. Furthermore, they provide a lower bound of the Rademacher complexity of the aforementioned models, demonstrating the optimality of their derived upper bound.

We consider an undirected graph $G \in \cG_{n,d}$ with $V(G)=[n]$ and $\amat, \lmat$   being the adjacency matrix of $G$ with added self-connections and the Laplacian matrix, respectively. We denote by $\Omega \subset V(G)$ the set of vertex indices with observed labels such that $m \coloneqq |\Omega| < n$. Additionally, for $d \in \Nb$, an each vertex $v \in \Omega$, we denote by $\vec{h}_v \in \Rb^{d}$ its feature vector and its label by $y_v \in \mathcal{Y} \subset \mathbb{R}$, while for each $u \in V(G) \setminus \Omega$ we only know the feature vectors in $\Rb^{d}$. The goal is to predict $y_u$ for every $u \in V(G)\setminus \Omega$. We consider the single-layer GCN architecture as described in \cref{def:singlehiddenlayergcn}.

The predicted label of vertex $v$ is denoted as $f_{G}(v)$. The expected loss is therefore defined as
\begin{equation*}
	\ell_{\text{exp}}(f_{G}) \coloneq \Eb_{P}[\ell(f_{G},v,y],
\end{equation*}
where $\ell$ is the loss function and the distribution $P$ is the underlying distribution on $\Rb^{d} \times \mathcal{Y}$. Similarly, given a training set through the node indices $\Omega$ as previously described, we can compute the empirical risk
\begin{equation*}
	\ell_{\text{emp}}(f_{G}) \coloneq \frac{1}{N}\sum_{u \in \Omega} \ell(f_{G}, u,y_i).
\end{equation*}
Now, consider the class $\mathcal{F}_{D,R}$ to be the class of two-layer graph neural networks over $\mathbb{R}^d$, under some technical assumptions analytically described in the original paper. The constants $D$ and $R$ are also defined according to these assumptions. Especially, $D$ bounds the $L_2$ norm of the weight vector of the output ($\norm{\vec{w}^{(2)}}_2 \leq D$), and $R$ bounds the Frobenius norm of the weight matrix in the first layer ($\norm{\vec{W}}^{(1)}_{F}<R$). Finally, an important assumption is that they consider all vertex having degree $q$ (e.g., regular graphs). Let $\hat{\mathcal{R}}(\mathcal{F}_{D,R})$ be the empirical Rademacher complexity of class $\mathcal{F}_{D,R}$ and let $\mathcal{F} \coloneq \{ \ell(y,f(\cdot)) \colon f \in \mathcal{F}_{D,R} \}$ where $\ell$ is a $\alpha_{\ell}$-Lipschitz continuous function. By \citet{DBLP:conf/colt/BartlettM01}, we have that
\begin{equation*}
	\ell_{\text{exp}}(f) \leq \ell_{\text{emp}}(f) + 2 \hat{\mathcal{R}}(\mathcal{F}) + \sqrt{\frac{2\log(2/\delta)}{N}}, \ \text{for all} \ f \in \mathcal{F},
\end{equation*}
and also $\hat{\mathcal{R}}(\mathcal{F}) \leq 2\alpha_{\ell}\hat{\mathcal{R}}(\mathcal{F}_{D,R})$.

Combining the above two results and bounding the empirical Rademacher complexity of $\mathcal{F}_{D,R}$, \citet{DBLP:journals/corr/abs-2102-10234} proved the following generalization bound.

\begin{theorem}
	Let $f$ be any given predictor of a class of GCNs with one hidden layer on a graph where all vertex degree $q$. For any $\delta >0$, with probability at least $1-\delta$, we have
	\begin{equation*}
		\ell_{\text{exp}}(f) \leq \ell_{\text{emp}}(f) + 16M^2BDR\alpha_{\ell}|\lambda_{\text{max}}(G)|\sqrt{\frac{q}{m}}\sum_{l=1}^{q}\max_{j \in [m]}|[g(\lmat)]_{jn_l(j)}| +  \sqrt{\frac{2\log(2/\delta)}{n}},
	\end{equation*}
	where $M$ is the Lipschitz constant of the activation function $\sigma$, $B$ bounds the $L_2$-norm of feature vectors in the input space, $\lambda_{\text{max}}$ is the maximum absolute eigenvalue of the graph filter $g(\lmat)$, and $n_l(j)$ refers to the $l$-th neighbor of node $v_j$ by a given order.
\end{theorem}

\section{Stability-based generalization bounds}

So far, we have only seen generalization bounds based on VC dimension theory and the Rademacher complexity of a hypothesis class. In all previous approaches, the bounds were based on bounding the capacity of the set of functions that the learning machine can implement. As nicely observed and described in \citet{Luxburg2011}, all the above bounds have the following form,
\begin{equation*}
	\ell_{\text{exp}}(f) \leq \ell_{\text{emp}}(f) + \text{capacity}(\mathcal{H}) + \text{confidence}(\delta), \text{ for all }  f \in \mathcal{H}.
\end{equation*}
In this context, $\mathcal{H}$ represents our hypothesis class, while the confidence term depends on the probability with which the bound should hold. It is important to recognize that all bounds structured this way are inherently pessimistic, as they apply universally to every possible learning algorithm that can be used to choose a function from $\mathcal{H}$. Consequently, the most extreme or problematic learning algorithm influences the bound's behavior. However, typically, real-world optimizers do not tend to pick the worst function in a model class.

\citet{Ver+2019}, based on \citet{stabilityoriginal}, followed a different approach to derive generalization bounds for single-layer GCN. Instead of uniformly bounding the capacity of the hypothesis class, they derive generalization bounds for learning algorithms satisfying useful properties. They also consider the inherent randomness in estimating the MPNN concept from sample data in the following setting. Intuitively, this randomness arises during the optimization of the MPNN concept using stochastic gradient descent. We first recall the notion of a learning algorithm, which is the optimization process for estimating the MPNN concept.

For an input space $\mathcal{X}$, an output space $\mathcal{Y}$, and a hypothesis class $\mathcal{H}$, a learning algorithm $\mathcal{A}$ maps an $N$-size sample from $\cZ^N$ to a concept $\mathcal{H}$, where $\cZ = \cX \times \cY$. Given the training set $\cS$ as $\cA_{\cS}$, we denote the concept learned. Here, $\cA_{\cS}$ represents an MPNN concept. Below, we define \new{uniform stability} for a randomized algorithm. Let $ \cS=\{z_1,z_2,\dots, z_N \} \in \cZ^{N}$ denote an $N$-sized sample. Additionally, we define $ \cS^{\setminus i} = \cS\setminus\{ z_i\} $, for $i \in [N]$.
\begin{definition}
	For a given sample size $N \in \Nb$, a randomized learning algorithm $\cA$ is \emph{$\beta$-uniformly stable} with respect to a loss function $\ell$ if it satisfies
	\begin{equation*}
		\sup_{\cS \in \cZ^N,z \in \cZ} \{ \left| \Eb_{\cA}[\ell(\cA_{\cS},z)] - \Eb_{\cA}[\ell(\cA_{\cS^{\setminus i}},z)] \right|  \} \leq 2 \beta, \quad \text{for all } i\in [N].
	\end{equation*}
\end{definition}
Here, the expectation regarding $\cA$ is over the implicit randomness in the algorithm. In our scenario, we can regard the algorithm $\cA$ as the optimization method used during the learning process (e.g., SGD).

Based on the stability properties of single-layer GCN models, \citet{Ver+2019} derived generalization bounds for node classification tasks under the assumption that the given graph can be decomposed into distinct $1$-hop subgraphs, each corresponding to the $1$-neighborhood of an individual node in the original graph. They further assume that these decomposed subgraphs are sampled independently and identically from an underlying distribution.

More precisely, following \citet[Theorem 12]{stabilityoriginal}, they first utilized the algorithmic uniform stability property to derive generalization error bounds in terms of the uniform stability parameter \citep[Theorem 2]{Ver+2019}. Subsequently, they showed that the algorithmic stability parameter is bounded by the largest absolute eigenvalue of the graph convolution filter \citep[Theorem 3]{Ver+2019}. By combining these results, they established the following generalization bound as a function of the largest eigenvalue of the graph convolution filter.

\begin{theorem}[\cite{Ver+2019}, Theorem 1]
	Let $\cA_{\cS}$ be a single-layer GCN equipped with the graph convolution filter $g(\vec{L})$ and trained on a dataset $S$ using the SGD algorithm for $T$ iterations. Assume the loss and activation functions are Lipschitz continuous and smooth. Then, for all $\delta \in (0,1)$ with probability at least $1-\delta$,
	\begin{equation*}
		\Eb_{\text{SGD}} [\ell_{\text{exp}}(\cA_\cS) - \ell_{\text{emp}}(\cA_\cS)] \leq \frac{1}{n} \mathcal{O} \left( \left( \lambda_{G}^{\text{max}}\right)^{2T}\right) + \left( \mathcal{O} \left( \left( \lambda_{G}^{\text{max}}\right)^{2T}\right) + M \right)\sqrt{\frac{\log(1/\delta)}{2N}},
	\end{equation*}
	where $\lambda_{G}^{\text{max}}$ is the absolute largest eigenvalue of $g(\vec{L})$, $N$ represents the number of training samples, and $M$ stands for a constant depending on the loss function.
\end{theorem}
Building on the work of \citet{stabilityoriginal,Mukherjee2006}, \citet{Zhou2021} established stability-based generalization bounds dependent on graph filters and their product with node features for graph convolutional networks with multiple layers. They demonstrated that the generalization error of GCNs tends to grow with more layers, providing an explanation for why GCNs with deeper layers exhibit relatively poorer performance on test datasets.

\section{PAC-Bayesian bounds}
\label{sec:pacbayesian}
In this section, we present the PAC-Bayesian approach \citep{DBLP:conf/colt/McAllester99,DBLP:conf/colt/McAllester03,DBLP:conf/nips/LangfordS02}, which is an alternative framework for deriving generalization bounds by assuming a prior distribution of the hypothesis class instead of having a deterministic model as in common learning formulations.

We slightly modify the previous notation here. For multiclass classification, $f_{\vec{\omega}}$ is a function from the hypothesis class, i.e., $f_{\vec{\omega}} \in \mathcal{H} \subseteq \{ f \colon \mathcal{G}_d \rightarrow \mathbb{R}^K \}$, where $\vec{\omega}$ is the vectorization of all model parameters and $K$ the number of classes for the multiclass classification problem. For a given loss function $\ell \colon \mathcal{H} \times \mathcal{G}_d \times \mathbb{R}^K \rightarrow \mathbb{R}$, as previously, we define the expected and empirical loss of the function $f_{\vec{\omega}}$ as follows,
\begin{equation*}
	\ell_{\text{exp}}(f_{\vec{\omega} }) \coloneq \Eb_{(G,y)\sim P}[\ell(f_{\vec{\omega}},G,y)]
	\quad
	\text{and}
	\quad
	\ell_{\text{emp}}(f_{\vec{\omega} }) \coloneq \frac{1}{N}\sum_{i=1}^{N}\ell(f_{\vec{\omega} },X_i,y_i).
\end{equation*}
Here, $N$ is the total number of i.i.d.\@ observations $(G_i,y_i) \sim P$.

In the PAC-Bayes approach, we assume a prior distribution $\pi$ over the hypothesis class $\mathcal{H}$, and we obtain a posterior distribution $Q$ over the same support through the learning process. Under this Bayesian view, we define the expected and the empirical error for a distribution $Q$ over $\mathcal{H}$, respectively, as
\begin{equation*}
	L_{\text{exp}}(Q) \coloneq \Eb_{\vec{\omega}  \sim Q}[\ell_{\text{exp}}(f_{\vec{\omega} })],
\end{equation*}
and
\begin{equation*}
	L_{\text{emp}}(Q) \coloneq \Eb_{\vec{\omega}  \sim Q}[\ell_{\text{emp}}(f_{\vec{\omega} })].
\end{equation*}
Following the above notation, \citet{DBLP:conf/colt/McAllester03} provided a general tool for deriving generalization bounds for any pair of distributions $(\pi, Q)$ based on the \say{distance} between $\pi$ and $Q$. This result is obtained using tools from information theory, which leverage the \new{Kullback–Leibler divergence} (KL) divergence to measure the discrepancy between the prior and posterior distributions. Formally, we have the following theorem.

\begin{theorem}
	\label{pacbayesianthm}
	For any prior distribution $\pi$ over $\mathcal{H}$ and any $\delta \in (0,1)$, with probability at least $1-\delta$, the following bound holds for all distributions $Q$ over $\mathcal{H}$,
	\begin{equation*}
		L_{\text{exp}}(Q) \leq L_{\text{emp}}(Q) + \sqrt{\frac{D_{\text{KL}}(Q \parallel \pi) + \log\mleft(\sfrac{2N}{\delta}\mright)}{2(N-1)}}.
	\end{equation*}
\end{theorem}

Here, $D_{\text{KL}}(Q \parallel \pi) \coloneq \sum_{x \in \mathcal{X}} P(x) \log\mleft(\frac{P(x)}{Q(x)}\mright)$ denotes the Kullback–Leibler divergence, which measures the discrepancy between the distributions $Q$ and $\pi$. Specifically, it quantifies the information loss incurred when using $Q$ as an approximation of $\pi$. A lower KL divergence indicates that $Q$ is closer to $\pi$, while a higher value suggests significant deviation.

While the above theorem provides a theoretical framework for generalization bounds, its direct application remains unclear, as it does not specify which choices of $\pi$ and $Q$ lead to a \say{good} bound. The key advantage of this theorem lies in its flexibility, allowing us to select a prior $\pi$ and a posterior $Q$ that yield a meaningful and computable bound. Therefore, as we will discuss later, a careful selection of $\pi$ and $Q$ is essential to ensure that the resulting bound is both tight and practically useful.

\subsection{PAC-Bayesian bounds for MPNNs and GCNs} \citet{Lia+2021}, following a PAC-Bayesian approach, established generalization bounds for multilayer GCNs and MPNNs. Their generalization bounds for both architectures depend on the maximum node degree of the graphs and the spectral norm of the weights of each model. Their bounds for GCNs can be seen as a natural generalization of the results developed in \citet{Neyshabur2018} for fully-connected and convolutional neural networks, while for MPNNs, the PAC-Bayesian bounds improve over the Rademacher complexity bounds by \citet{Gar+2020}.

We focus on the $K$-class classification problem and make the following assumptions.
\begin{itemize}
	\item [(A1)] For both MPNNs and GCNs, the maximum hidden dimension across all layers is denoted as $h$.
	\item [(A2)] The node features of every node on each graph are contained in a $\ell_2$-ball with radius $B$.
	\item [(A3)] The maximum node degree in all graphs is $\tilde{d}-1$.
\end{itemize}
In the following, we use the multi-class $\gamma$-margin loss for some $\gamma >0$ as our loss function. That is, the expected loss is defined as,
\begin{equation*}
	\ell_{\mathrm{exp},\gamma}(f_{\vec{\omega} }) \coloneq \mathbb{P} \left( f_{\vec{\omega} }(G)[y] \leq \gamma + \max_{j \neq y}{f_{\vec{\omega} }(G)[j]} \right),
\end{equation*}
and the empirical loss,
\begin{equation*}
	\ell_{\mathrm{emp},\gamma}(f_{\vec{\omega} }) \coloneq \frac{1}{n} \sum_{i=1}^{n} \mathbf{1}\left(f_{\vec{\omega} }(G_i)[y_i] \leq \gamma + \max_{j \neq y}{f_{\vec{\omega} }(G_i)[j]} \right),
\end{equation*}
where, $f_{\vec{\omega} }(G)[j]$ is the $j$-th entry of $f_{\vec{\omega} }(G) \in \mathbb{R}^{K}$, and $y_i$ is the label of $G_i$. Before presenting the main results of \citet{Lia+2021}, it is crucial to note that in this approach, the PAC-Bayes framework serves only as a tool to derive a generalization bound, and we do not assume randomness in $\vec{\omega} $. Hence, the objective is to establish a bound applicable to every deterministic $\vec{\omega} $. Nonetheless, to employ PAC-Bayes theory, we extend the deterministic $\vec{\omega} $ to a distribution by introducing a random perturbation $u$. Moreover, the prior distribution $\pi$ on the hypothesis class must remain independent of $\vec{\omega} $. Since $\pi$ cannot rely on $\vec{\omega} $, we aim to minimize the Kullback-Leibler divergence between $\pi$ and $Q$ (where $Q$ is defined by $\vec{\omega}  +u$) by selecting a prior $\pi$ and a random perturbation $u$ with sufficiently high variance allowing $\pi$ and $\vec{\omega}  +u$ being similar. Finally, \cite{Neyshabur2018}, combining the above idea with \cref{pacbayesianthm}, proved the following Lemma, which converts PAC-Bayes bounds to generalization bounds on deterministic $\vec{\omega} $ for the margin loss.
\begin{lemma}[\cite{Neyshabur2018}]
	\label{PACbayeslemma}
	Let $f_{\vec{\omega} } \colon \cX \rightarrow \Rb^{K}$ be any model with parameters $\vec{\omega} $, and let $\pi$ be any prior distribution on the parameters that is independent of the training data. For any $\vec{\omega} $, we construct a posterior $Q(\vec{\omega} +u)$ defined by the perturbation $u$ to $\vec{\omega} $, s.t.,  $\mathbb{P} \left( \max_{x \in \cX} |f_{\vec{\omega} +u}(x) - f_{\vec{\omega} }(x)|_{\infty} < \sfrac{\gamma}{4} \right)>\sfrac{1}{2}$. Then, for any $\gamma, \delta>0$, with probability at least $1-\delta$ over an i.i.d. size-$N$ training set, for any $\vec{\omega} $, we have,
	\begin{equation*}
		\ell_{\text{exp},0}(f_{\vec{\omega} }) \leq \ell_{\text{emp},\gamma}(f_{\vec{\omega} }) + \sqrt{\frac{2D_{\text{KL}}
				(Q(\vec{\omega} +u)\parallel \pi)+\log{(\sfrac{8N}{\delta})} }{2(N-1)}}.
	\end{equation*}
\end{lemma}
By assuming spectral norm bounds on all layers, we can constrain the change in the value of $f_{\vec{\omega} }$ as $\vec{\omega} $ is perturbed, thereby controlling $|f_{\vec{\omega} }(x)-f_{\vec{\omega} +u}(x)|$ independently of $x$ and enabling the utilization of \cref{{PACbayeslemma}}.

\begin{theorem}[GCN generalization bound]
	Let $f_{\vec{\omega} } \in \mathcal{H} \subset \{ h \colon \mathcal{G} \rightarrow \mathbb{R}^K \}$ be an $L$-layer GCN under the assumptions (A1)-(A3). Then, for any $\delta, \gamma > 0$, with probability at least $1-\delta$ over the choice of an i.i.d.\@ size $N$ training set, for any $\vec{\omega} $, we have,
	\begin{align*}
		\ell_{\text{exp},0}(f_{\vec{\omega} }) & \leq \ell_{\text{emp},\gamma}(f_{\vec{\omega} }) \\ &+ \mathcal{O}\mleft( \sqrt{\frac{B^2 \tilde{d}^{L-1}L^2h\log{(Lh)} \prod_{i=1}^{L}\parallel  \vec{W}_i\parallel ^2_2\sum_{i=1}^{L}\mleft( \parallel \vec{W}_i\parallel _{F}^2/\parallel \vec{W}_i\parallel _2^2 \mright) + \log{(\sfrac{NL}{\delta})}}{\gamma^2 N}} \mright),
	\end{align*}
	where $\tilde{d}, h$ as defined in assumptions (A1)-(A3), and $\vec{W}_i$, are the parameter matrices of the GCN, as described in \cref{sec:kGCNs}.
\end{theorem}

\begin{theorem}[MPNN generalization bound]
	Let $f_{\vec{\omega} } \in \mathcal{H} \subset \{ h \colon \mathcal{G} \rightarrow \mathbb{R}^K \}$ be an $L$-layer MPNN under the assumptions (A1)-(A3). Then, for any $\delta, \gamma > 0$, with probability at least $1-\delta$ over the choice of an i.i.d.\ size $N$ training set, for any $\vec{\omega} $, we have,
	\begin{align*}
		\ell_{\text{exp},0}(f_{\vec{\omega} }) & \leq \ell_{\text{emp},\gamma}(f_{\vec{\omega} }) \\ &+ \mathcal{O}\left( \sqrt{\frac{B^2 \left( \max(\zeta^{-(L+1)},(\lambda \xi)^{(L+1)/L} \right)^2 L^2 h \log{(Lh)}\parallel \vec{\omega}  \parallel ^2_2+\log{(\sfrac{N(L+1)}{\delta})} }{\gamma^2 N}} \right)
		,
	\end{align*}
	where $\zeta, \lambda$ depend on the spectral norm of the parameter matrices of the MPNN, $\|\vec{\omega} \|_2^{2}$ depends on the Forbenius norm of the parameter matrices, and $\xi$ is a constant depending on the Lipschitz constants of the activation functions and the maximum node degree $\tilde{d}-1.$
\end{theorem}

Comparing the bounds established by \citet{Liao+2019} to those in \citet{Gar+2020}, \citet{scarselli2018vapnik} in terms of their dependency on the maximum node degree and maximum hidden dimension, we can derive the following conclusions. Firstly, the bound proposed by \citet{Gar+2020} scales as $\mathcal{O}(d^{L-1} \sqrt{p \log(d^{2L-3})})$, whereas \citet{Liao+2019} scales as $\mathcal{O}(d^{L-1})$, indicating a slight improvement. Additionally, the bound by \citet{Liao+2019} scales as $\mathcal{O}(\sqrt{h \log h})$, tighter than the Rademacher complexity bound $\mathcal{O}(h \sqrt{\log h})$ by \citet{Gar+2020} and the VC-dimension bound $\mathcal{O}(h^4)$ by \citet{scarselli2018vapnik}.

\citet{DBLP:journals/corr/abs-2402-04038}, in a similar work to \citet{Lia+2021}, extended the establishment of generalization bounds for GCNs and MPNNs using a PAC-Bayesian framework. They also established bounds depending on the spectral norm of the diffusion matrix and the spectral norm of the weights. Specifically, they improve upon the bounds proposed by \citet{Lia+2021} by eliminating the exponential dependence on the maximum node degree. Additionally, \citet{Wu2023} derived non-trivial extensions for the above generalization bounds on GCNs avoiding the exponential dependence on the maximum node degree. Finally, in a recent work, \citet{DBLP:conf/icml/LeeHW24} applied a PAC-Bayesian approach to derive generalization bounds for knowledge graph representation learning, providing a first theoretical contribution in this domain.

\subsection{Graph diffusion matrix dependent bound}
\citet{Ju+2023} followed a PAC Bayesian approach and measured the stability of GNNs against noise perturbations through the Hessian matrices. They derived generalization bounds for a more general architecture of GNNs, including message passing graph neural networks \citep{Gil+2017}, graph convolution networks \citep{Kip+2017}, and graph isomorphism networks \citep{Xu2019}.

The main difference from the previously described PAC-Bayesian bounds is that in this work, the generalization bounds scale with the largest singular value of the GNN feature diffusion matrix (e.g., the adjacency matrix) instead of the maximum node degree. In their work, they also constructed a lower bound of the generalization gap, which asymptotically matches the upper bound. In the following, we briefly present their framework considering a very general set of GNN models for graph-level prediction tasks, i.e.,~\citet{Dai+2016, Gar+2020, Gil+2017, Jin2018}, as well as the derived generalization bounds.

Since in the main result, the generalization bound depends on the graph diffusion matrix, we present the matrix notation of GNNs described in \citet{Ju+2023}. Let $L$ be the number of layers (the $L$-th layer is the pooling layer), and $d_t$ denotes the width of each layer for $t \in [L]$. We use $\phi_{t}$, $\rho_{t}$, $\psi_{t}$ as nonlinear functions centered at zero. Additionally, we assume weight matrices $\vec{W}^{(t)}$ of dimension $d_{t-1} \times d_{t}$ for transforming neighboring nodes and another weight matrix $\vec{U}^{(t)}$ with the same dimension. For the first $L-1$ layers, the node embeddings  are recursively computed from the input feature matrix $\vec{H}^{(0)} \coloneq \vec{X} \in \Rb^{n \times d}$ for any graph $G \in \cG_{n,d}$ with diffusion matrix $\vec{P}_G$ as follows,
\begin{equation*}
	\vec{H}^{(t)} \coloneq \phi_{t} \mleft(  \vec{XU}^{(t)}+\rho_{t} \left(  \vec{P}_G \psi_{t} \left( \vec{H}^{(t-1)} \right)  \right) \vec{W}^{(t)} \mright).
\end{equation*}
For the last layer $L$, we aggregate using the vector $\mathbf{1}_n$ containing ones everywhere as,
\begin{equation*}
	H^{(L)} \coloneq \frac{1}{n} \mathbf{1}_n^{T} \vec{H}^{(L-1)} \vec{W}^{(L)}.
\end{equation*}
Possible common choices for the graph diffusion matrix $\vec{P}_G$ include the adjacency matrix $\vec{A}(G)$ or the normalized adjacency $\vec{D}(G)^{-1}A(G)$. Note that the above architecture can be considered a special case of the broader MPNN architecture described in \cref{def:MPNN}.
We arrive at the following bound for the generalization error.
\begin{theorem}
	Suppose all of the nonlinear functions $\phi_{t}$, $\rho_{t}$, $\psi_{t}$, and the loss function $\ell(\cdot, y)$ are twice-differentiable, Lipschitz-continuous, and their first and second-order derivatives are both Lipschitz continuous. With probability at least $1-\delta$, for any $\delta>0$ and any $\epsilon>0$, any model $f$ following the previously described architecture satisfies,
	\begin{equation*}
		\ell_{\text{exp}}(f) \leq \ell_{\text{emp}}(f) + \mathcal{O}\mleft( \frac{\log\left(\delta^{-1} \right)}{N^{\frac{3}{4}}} \mright) + \sum_{i=1}^L \sqrt{ \mleft( \frac{CBd_i \mleft( \max \norm{ X} ^2 \norm{\vec{P}_G}^{2(L-1)}\mright) \left( r_i^2 \prod_{j=1}^{l}s_j^2 \right)}{N} \mright) },
	\end{equation*}
	where $N$ is the size of the i.i.d.\ sample, $B$ is an upper bound of the loss function $\ell$, and $C$ is a constant depending on the Lipschitz constants of $\phi_{t}$, $\rho_{t}$, $\psi_{t}$, and $\ell(\cdot, y)$. The constants $r_i,s_j$ are constants used for bounding the spectral and the Forbenius norm of the weight matrices $\vec{W}^{(t)}$ and $\vec{U}^{(t)}$.
\end{theorem}
The bounds proposed in \citet{Ju+2023} scale with the spectral norm of the graph diffusion matrix $P_G$, while the bounds established in \citep{Gar+2020, Liao+2019} scale with the graph's maximum degree. By utilizing appropriate diffusion matrices \citep{Hamilton2017,Fen+2022}, it can be shown that the spectral norm of $\vec{P}_G$ is strictly smaller than the maximum degree of the graph, suggesting that the bounds presented in \citet{Ju+2023} in many cases might be tighter.

\section{Covering number-based generalization bounds}
\label{sec:coveringnumberbounds}

In this section, we present generalization bounds for MPNNs by endowing the space of graphs with a (pseudo-)metric that satisfies two key conditions.
\begin{enumerate}
	\item MPNNs are Lipschitz-continuous functions concerning this (pseudo-)metric.
	\item The induced (pseudo-)metric space of graphs is compact.
\end{enumerate}
If these conditions hold, then bounding the covering number of this space leads to a generalization bound for MPNNs. Before presenting the two main theorems that motivate using graph-metric spaces for deriving generalization bounds, we introduce some notations for covering numbers and metric spaces.

\paragraph{Pseudo-metric spaces and covering numbers}
Let $\cX$ be an arbitrary set. A function $d \colon \cX \times \cX \to \Rb^+$ is called a \emph{pseudo-metric} on $\cX$ if it satisfies:
(i) $d(x,x) = 0$, (ii) $d(x,y) = d(y,x)$ for all $x,y \in \cX$, and (iii) $d(x,y) \leq d(x,z) + d(z,y)$, for all $x,y,z \in \cX$. The pair $(\cX,d)$ is called a \emph{pseudo-metric space}. If, in addition, $d(x,y) = 0 \Rightarrow x=y$ for all $x,y \in \cX$, then $d$ is a \emph{metric}, and $(\cX,d)$ is called a \emph{metric space}.

Given a pseudo-metric space $(\cX,d)$ and $\varepsilon > 0$, an \emph{$\varepsilon$-cover} of $\cX$ is a subset $C \subseteq \cX$ such that for every $x \in \cX$, there exists a $y \in C$ satisfying $d(x,y) \leq \varepsilon$. The \emph{covering number} of $\cX$ is then defined as
\begin{equation*}
	\cN(\cX, d, \varepsilon) \coloneq \min \left\{ m \mid \text{there exists an $\varepsilon$-cover of $\cX$ with cardinality $m$} \right\},
\end{equation*}
i.e., the smallest number of elements required to form an $\varepsilon$-cover of $\cX$ concerning $d$. Given a pseudo-metric space $(\cX,d)$, one can construct a corresponding metric space $\tilde{\cX}$ by considering the equivalence relation $\sim$ on $\cX$, where $x \sim y$ if and only if $d(x,y)=0$. The quotient space $\tilde{\cX} = \cX /_{\sim} = \{[x] \mid x \in \cX\}$ is then endowed with the metric
\begin{equation*}
	\tilde{d}([x],[y]) = d(x,y).
\end{equation*}

Finally, given a pseudo-metric space $(\cX,d)$, we say that a subset $U \subset \cX$ is \emph{open} if for every $x \in U$, there exists $\varepsilon > 0$ such that for all $y \in \cX$ with $d(x,y) < \varepsilon$, we have $y \in U$. A pseudo-metric space $(\cX,d)$ is \emph{compact} if for every $\varepsilon > 0$ and every $\varepsilon$-cover $\cup_{i \in I} U_i$ of $\cX$ consisting of open sets $U_i \subset \cX$, there exists a finite subcover $\cup_{i \in J} U_i$ with $J \subset I$, and $|J| < \infty$. It follows that for any compact metric space $(\cX,d)$, we have $\cN(\cX,d,\varepsilon) < \infty$ for all $\varepsilon > 0$. Moreover, if the completion,\footnote{For any pseudo-metric space $\mathcal{X}$ there is unique pseudo-metric space $\overline{\mathcal{X}}$, called the \emph{completion} of $\mathcal{X}$, in which every Cauchy sequence converges, and such that $\mathcal{X}$ is a dense subset of $\overline{\mathcal{X}}$. An example is $\mathbb{R}$ being the completion of the rationals $\mathbb{Q}$ with the metric $|x-y|$.} of a pseudo-metric space is compact, the pseudo-metric space has finite covering.

\paragraph{Covering number-based generalization bounds}
We now present two closely related theorems that justify using compact metric spaces for deriving generalization bounds in graph learning. The following theorem from \citet[Theorem G.3]{Lev+2023} is an extension of \citet[Lemma 2]{Mas+2022}.

\begin{theorem}
	\label{thm:mpnn_generalization}
	Let $(\cZ,d)$ be a metric space with finite covering, $\mu$ a Borel distribution on $\cZ$, and $\{Z_i\}_{i=1}^{N}$ be i.i.d.\@ samples from $\mu$ for some $N \in \Nb$. Suppose $\cH \subset \{h \colon\cZ \to \Rb\}$ consists of bounded and $L$-Lipschitz functions for some $L \in \Rb$. Then, the following holds for any $\delta > 0$, with probability at least $1-\delta$. For every $f \in \cH$,
	\begin{equation*}
		\left| \frac{1}{N} \sum_{i=1}^{N} f(Z_i) - \Eb_{\mu}[f(Z_1)] \right| \leq 2 \xi^{-1}(N)L + \frac{1}{\sqrt{2}} \xi^{-1}(N) M (1 + \sqrt{\log(\sfrac{2}{\delta})}),
	\end{equation*}
	where $\Eb_{\mu}[f(Z_1)] \coloneq \int f(x) \, \mu(dx)$, $\xi(\epsilon) \coloneq\frac{2 \kappa(\epsilon)^2 \log(\kappa(\epsilon))}{\epsilon^2}$,  $\kappa(\epsilon) \coloneq \cN(\cZ,d,\epsilon)$,
	and $\xi^{-1}$ is the inverse function of $\xi$.
\end{theorem}

This theorem provides a method for deriving generalization bounds given a data distribution over a compact metric space, with a proof based on Hoeffding’s inequality. It is the key idea behind the approaches taken by \citet{Lev+2023, Rac+2024}. An alternative similar framework, used by \citet{Vas+2024}, involves a learning algorithm's notion of \emph{robustness}.

\begin{definition}
	We say that a (graph) learning algorithm for a hypothesis class $\cH$, is \new{$(K,\varepsilon(\cdot))$-robust} if $\cZ$ can be partitioned into $K>0$ sets, $\{C_i\}_{i=1}^{K}$, such that for all samples $\cS$ and $(G,y)\in\cS$, the following holds. If $(G,y)\in C_i$, for some $i\in\{1,\ldots,K\}$, then for all $(G',y')\in C_i$,
	\begin{equation*}
		\big|\ell\bigl(h_{\cS}(G),y\bigr) - \ell\bigl(h_{\cS}(G'),y'\bigr) \big|<\varepsilon(\cS).
	\end{equation*}
	Recall that $h_{\cS}$ is the function returned by the learning algorithm regarding the data sample $\cS$, and $\ell$ is a loss function.
\end{definition}
Intuitively, the above definition requires that the difference of the losses of two data points in the same cell is small. \citet{Xu+2012} showed that a \new{$(K,\epsilon(\cdot))$-robust} learning algorithm implies a bound on the generalization error.
\begin{theorem}[\citet{Xu+2012}, Theorem 3]
	\label{robustness_bound}
	If we have a $(K,\epsilon(\cdot))$-robust (graph) learning algorithm for $\cH$ on $\cZ$, then for all $\delta>0$, with probability at least $1-\delta$, for every sample $\cS$,
	\begin{equation*}
		\label{eq:xumannorbound}
		|\ell_{\mathrm{exp}}(h_{\cS})-\ell_{\mathrm{emp}}(h_{\cS})| \leq  \epsilon(\cS) + M \sqrt{\frac{2K\log(2)+2\log(\sfrac{1}{\delta})}{|\cS|}},
	\end{equation*}
	where $h_{\cS}$, as before, denotes a function from the hypothesis class $\cH$ returned by the learning algorithm given the data sample $\cS$ of $\cZ$. Additionally, $M$ denotes a bound on the loss function $\ell$.
\end{theorem}
\citet{Kaw+2022} improved the above generalization bound by establishing a data-dependent bound that reduces the dependency on $K$ from $\sqrt{K}$ to $\log{(K)}$, as shown in \citep[Theorem 1]{Kaw+2022}. Even though $\epsilon$-covers and partitions of a set are closely related notions, as each $\epsilon$-cover trivially induces a partition, it may not yet be entirely clear how compactness can be leveraged to derive generalization bounds using the notion of robustness. This will become clearer later with \cref{proposition:robustnessfortreedistance(class)}.

The key insight in the following results is using more fine-grained expressivity properties of MPNNs compared to the $\wlone$ distinguishability that the VC dimension-based analyses used; see~\cref{sec:VCbounds}. Specifically, while \citet{Mor+2023} leveraged the property that MPNN outputs for two graphs are identical whenever these graphs are \wlone-equivalent, as shown in \citep{Mor+2019}, covering number approaches exploit a more \say{fine-grained} property of MPNNs. Namely, the MPNN outputs for two graphs are close in Euclidean distance whenever the graphs are close regarding some appropriately defined distance pseudo-metric between graphs.

\paragraph{Recipe for covering number-based MPNN generalization bounds}
Theorems \ref{thm:mpnn_generalization} and \ref{robustness_bound} provide generalization bounds for any hypothesis space of Lipschitz continuous functions $f$, mapping data from a compact metric space $(\mathcal{M}, d_{\mathcal{M}})$ to $\Rb^d$, given an i.i.d.\@ sample from any Borel distribution $\mu$ on $(\mathcal{M}, d_{\mathcal{M}})$. In the case of MPNNs and graph learning, the function $f$ corresponds to the composition of the loss function $\ell$ with the MPNN's output $h$ (after the readout layer).

\begin{figure}
	\centering
	\resizebox{0.45\textwidth}{!}{\begin{tikzpicture}[transform shape, scale=1]

\node[] at (0,0) {\scalebox{1}{$\mathcal{G}$}};
\node[] at (0,1.7) {\scalebox{1}{$\mathcal{G}$}};
\node[] at (2.2,0) {\scalebox{1}{$(\mathsf{\mathcal{M},d_\mathcal{M}})$}};
\node[] at (2.2,1.7) {\scalebox{1}{$(\mathsf{\mathcal{M},d_\mathcal{M}})$}};
\node[] at (1.1,-1.4) {\scalebox{1}{$\mathbb{R}^{d}$}};

\node[rotate=-49] at (0.25,-0.75) {\scalebox{0.55}{\textsf{readout}}};
\node[rotate=55] at (1.8,-0.75) {\scalebox{0.55}{\textsf{readout}}};

\draw[thick,-stealth] (0.3,0) to[bend left=0] (1.4,0);
\draw[thick,-stealth] (0,1.4) to[bend left=0] (0,0.3);
\draw[thick,-stealth] (2.2,1.4) to[bend left=0] (2.2,0.3);
\draw[thick,-stealth] (0.3,1.7) to[bend left=0] (1.4,1.7);
\draw[thick,-stealth] (0.05,-0.3) to[bend right=0] (0.85,-1.15);
\draw[thick,-stealth] (1.95,-0.3) to[bend left=0] (1.35,-1.15);

\node[black!80] at (4.5,2.2) {\scalebox{0.8}{$(\mathsf{\widetilde{\mathcal{WS}}_r,\text{$\delta$}_{\square}})$}};

\node[black!80] at (4.6,1.25) {\scalebox{0.8}{$(\mathsf{IDM,d_{IDM}})$}};
\draw[black!80,thick,dashed,dash pattern=on 3pt off 2pt] (3.8,2.1) to[bend left=0] (3,1.9);
\draw[black!80,thick,dashed,dash pattern=on 3pt off 2pt] (3.8,1.35) to[bend left=0] (3,1.5);

\node[black!80] at (0.85,1.9) {\scalebox{1}{$\pi$}};
\node[black!80] at (0.85,0.2) {\scalebox{1}{$\pi$}};

\node[black!80,rotate=90] at (1.9,0.85) {\scalebox{0.5}{\textsf{message}}};
\node[black!80,rotate=90] at (2.05,0.85) {\scalebox{0.5}{\textsf{passing}}};

\node[black!80,rotate=90] at (-0.29,0.85) {\scalebox{0.5}{\textsf{message}}};
\node[black!80,rotate=90] at (-0.14,0.85) {\scalebox{0.5}{\textsf{passing}}};
\node[black!80] at (0.2,0.85) {\scalebox{0.7}{$\Theta$}};
\node[black!80] at (2.4,0.85) {\scalebox{0.7}{$\Theta$}};
\end{tikzpicture}}
	\caption{Commutative diagram illustrating the graph compact embedding process. In this diagram, the message-passing phase is represented as a function that updates node features, producing a graph with the same structure but updated node features.}
	\label{fig:commutative_diagram}
\end{figure}
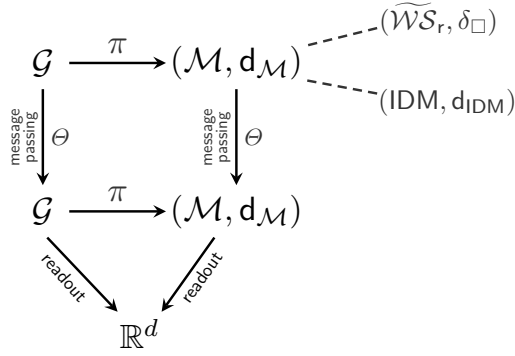

There are two approaches in the literature for applying the theorems above in graph learning tasks. The first approach involves directly endowing the space of graphs $\cG$ with a pseudometric $d_{\cG}$ such that 1) MPNNs are Lipschitz continuous and 2) the completion of the space $\cG$ under $d_{\cG}$ gives a compact space. The fact that compact spaces always have a finite covering number leads to a generalization bound. Since the graphs' order is not bounded, we call this the \emph{uniform regime}.  The following technique for the uniform regime, which we call \textit{the GNN Lipschitz-compact covering bounds}, was first introduced in \citet{Lev+2023}. The commutative diagram \cref{fig:commutative_diagram} illustrates the following steps.

\begin{enumerate}
	\item Define a compact space $(\cM, d_{\cM})$.
	\item Define a mapping $\pi:\cG\rightarrow \cM$ that associates each graph $G \in \cG$ with an element $\pi(G) \in \cM$.
	\item Consider the \emph{pullback} pseudometric $\pi^* d_{\cM}$ in $\cG$ defined by
	      \[\pi^* d_{\cM}(G_1,G_2)= d_{\cM}(\pi(G_1),\pi(G_2)).\]
	      Note that $\cG$ is precompact with respect to $\pi^*(d_{\cM})$, namely, the completion of $\cG$ is compact.
	\item For every $\text{MPNN}_{\Theta}:\cG\rightarrow \Rb$ with parameters $\Theta$, define a corresponding Lipschitz continuous mapping $\overline{\text{MPNN}}_{\Theta}: \cM \to \Rb$, also called an MPNN,  satisfying the equivariance relation
	      \begin{equation*}
		      \pi(\text{MPNN}_{\Theta}(G)) = \overline{\text{MPNN}}_{\Theta}(\pi(G)).
	      \end{equation*}
	      Note that the above construction assures that $\text{MPNN}_{\Theta}$ are Lipschitz continuous with respect to $\pi^*(d_{\cM})$.
\end{enumerate}
Then, since precompact spaces have finite covering (the same covering number as their compact completions), an upper bound on $\cN(\cG, \pi^* d_{\cM}, \epsilon)$ is guaranteed, leading to a generalization bound through \cref{thm:mpnn_generalization}. This approach was introduced by \citet{Lev+2023}, where $\cM$ was the space of graphon-signals and $d_{\cM}$ was the cut distance. \citet{Rac+2024} later utilized this approach, where $\cM$ was defined as the space of iterated degree measures (IDMs) and $d_{\cM}$ the DIDM-mover's distance.

We note that making the space of graphs compact and MPNNs Lipschitz continuous can be challenging. Compactness requires ``all graphs to be close to each other,'' and Lipschitz continuity requires the graphs to be ``sufficiently far apart for MPNNs to have bounded slopes.''  To avoid this complication, the second approach for GNN covering number generalization bounds involves restricting the focus to a finite space of graphs (e.g., $\cG_n$), where finiteness immediately implies compactness for any pseudo-metric. In addition, various combinatorial techniques, such as the set-covering problem, can be employed on finite spaces to derive tight upper bounds or even compute the covering number exactly. This methodology is followed by \citet{Vas+2024}, and since their bounds apply only to $n$-order graphs and not uniformly across the space of all graphs, we refer to these bounds as the \textit{non-uniform regime}.

\paragraph{MPNNs hypothesis class} The following works consider MPNNs with the following architecture. Each layer is defined using \new{order-normalized sum aggregation}, and we further use a mean readout layer, where
\begin{equation}\label{def:unlabeledmpnnsgraphs}
	\vec{h}_{v}^{(t)} \coloneq \varphi_{t}\Bigl(\vec{h}^{(t-1)}_u, \tfrac{1}{|V(G)|} \sum_{u \in N(v)} \vec{h}^{(t-1)}_u \Bigr) \quad \text{and} \quad \vec{h}_{G} \coloneq \psi \Bigl( \tfrac{1}{|V(G)|} \sum_{u \in V(G)} \vec{h}^{(L)}_u \Bigr),
\end{equation}
for $v \in V(G)$, where $\varphi_{t} \colon \Rb^{d_{t-1}+1} \to \Rb^{d_t}$ is an $L_{\varphi_t}$-Lipschitz continuous function with respect to the $2$-norm-induced metric, for $d_t \in \Nb$ and $t \in [L]$. Additionally, $\psi \colon \Rb^{d_L} \to \Rb^{d}$ is an $L_{\psi}$-Lipschitz continuous function.\footnote{As implied by~\citet{Mor+2019,xu2021how}, choosing the function $\varphi_t$ appropriately leads to a \wlone-equivalent MPNN layer.} For some graph space $\cX$ they analyze the class of $L$-layer MPNNs, based on~\cref{def:unlabeledmpnnsgraphs}, where $\psi$ is represented by a feedforward neural network (FNN) with Lipschitz constant $L_{\mathsf{FNN}}$ and bounded by $M' \in \Rb$, denoted as $\MPNN^{\mathrm{ord}}_{L,M',L_{\mathsf{FNN}}}(\cX)$:
\begin{equation*}
	\MPNN^{\mathrm{ord}}_{L,M',L_{\mathsf{FNN}}}(\cX) \coloneq \Bigl\{
	h \colon \cX \rightarrow \Rb \mathrel{\Big|} h(G) \coloneq \FNN_{\mathbold{\theta}} \circ \vec{h}_{G}, \text{ where } \mathbold{\theta} \in \Theta  \Bigr\}.
\end{equation*}
Additionally, in the case of unattributed graphs, they simplify the architecture to the following:
\begin{equation*}
	\vec{h}_{v}^{(t)} \coloneq \varphi_{t}\Bigl(\sfrac{1}{|V(G)|} \sum_{u \in N(v)} \vec{h}^{(t-1)}_u \Bigr),
\end{equation*}
which remains $\wlone$-equivalent in distinguishing non-isomorphic graphs.

\subsection{The uniform regime}
Below, we present two works deriving generalization bounds via the uniform regime, as illustrated in
\cref{fig:commutative_diagram}. We provide a detailed discussion of the work of \citet{Lev+2023}, where the space $\cM$ is chosen as the space of graphon-signals. Additionally, we briefly mention the approach taken by \citet{Rac+2024}, where $\cM$ is the space of iterated degree measures.

\paragraph{Graphon-signals}
A graphon can be viewed as a generalization of a graph, where instead of discrete sets of nodes and edges, there is an infinite set indexed by the sets $ V(W) \coloneq [0,1] $ for nodes and $ E(W) \coloneq V(W)^2 = [0,1]^2 $ for edges. A graphon is defined as a measurable symmetric function $ W \coloneq E(W) \to [0,1] $, i.e., $ W(x,y) = W(y,x) $. Each value $ W(x,y) $ represents the probability or intensity of a connection between points $x$ and $y$. Graphons are used to study the limit behavior of large graphs \citep{Lov+2012}. A graphon-signal, introduced by \citet{Lev+2023}, is a pair $(W, \vec{f})$, where $W$ is a graphon and $\vec{f} \colon V(W) \to \Rb^d$ is a measurable function. We denote by $ \mathcal{WS}^d $ the space of all graphon-signals with signals $\vec{f} \colon V(W) \to \Rb^d $. Additionally, for $r>0$, we define the space of graphon-signals with $r$-bounded signal functions as $\mathcal{WS}^d_{r} = \mathcal{W} \times \mathcal{L}^{\infty}_r[0,1]$, where
\begin{align*}
	\label{n:Linfr1}
	\mathcal{L}^{\infty}_r[0,1] \coloneqq \{ \vec{f} \colon [0,1] \to \Rb^d \mid \|\vec{f}(x)\|_{\infty} \leq r  \}.
\end{align*}

Any labeled graph can be identified with a corresponding graphon-signal as follows. Let $ (G, \ell_G) $ be a labeled graph with node set $ \{1, \dots, n\} $ and adjacency matrix $ \vec{A}(G) = \{a_{i,j}\}_{i,j \in [n]} $. Let $ \{I_k\}_{k=1}^N $ with $ I_k = \left[ \frac{k-1}{n}, \frac{k}{n} \right) $ be the equipartition of $ [0,1] $ into $ n $ intervals. The graphon-signal $ (W, f)_{(G,\ell_G)} = (W_G, f_{\ell_G}) $ induced by $ (G, \ell_G) $ is defined by
\begin{equation*}
	W_G(x, y) \coloneq \sum_{i,j=1}^n a_{ij} \, \mathbf{1}_{I_i}(x) \, \mathbf{1}_{I_j}(y) \quad \text{and} \quad f_{\ell_G}(z) \coloneq \sum_{i=1}^n \ell_G(i) \, \mathbf{1}_{I_i}(z),
\end{equation*}
where $ \mathbf{1}_{I_i} $ is the indicator function of the set $ I_i \subset [0,1] $. We write $ (W, \vec{f})_{(G,\ell_G)} = (W_G, \vec{f}_{\ell_G}) $ and identify any labeled graph with its induced graphon-signal.

\paragraph{Cut metrics.}
For any graphon $W$, we define the cut norm of $W$ as
\begin{equation*}
	\|W\|_{\square} \coloneq \sup_{U,V \subset [0,1]} \left| \int_{U \times V} W(x,y) \, dx \, dy \right|,
\end{equation*}
where the supremum is over measurable sets  $U, V \subset [0,1]$. The cut metric between two graphons $W, W' \in \mathcal{W}$ is defined as
\begin{equation*}
	d_{\square}(W, W') \coloneq \|W - W'\|_{\square}.
\end{equation*}
This measures the difference between the average edge weights of $W$ and $W'$ on the cut $U \times V$ that maximizes this difference. The \emph{signal cut norm} on $\mathcal{L}^{\infty}_r[0,1]$ is defined by
\begin{equation*}
	\label{signalnormr}
	\|f\|_{\square} \coloneq \sup_{S \subseteq [0,1]} \left| \int_{S} f(x) \, d\mu(x) \right|,
\end{equation*}
where the supremum is taken over the measurable subsets $S \subset [0,1]$. For simplicity of notation, we assume that signals map to $\Rb$ (i.e., $d = 1$), and we omit $d$ from the notation. However, all results hold also for $d$-dimensional signals. The \emph{graphon-signal cut norm} on $\mathcal{WS}_r$ is defined by
\begin{equation*}
	\|(W, f)\|_{\square} \coloneq \|W\|_{\square} + \|f\|_{\square}.
\end{equation*}
The \emph{graphon-signal cut metric} and \emph{graphon-signal cut distance} are defined  by
\begin{equation*}
	d_{\square}\big((W, f), (V, g)\big) \coloneq \|(W, f) - (V, g)\|_{\square},
\end{equation*}
and
\begin{equation*}
	\delta_{\square}\big((W, f), (V, g)\big) \coloneq \inf_{\psi \in S_{[0,1]}} d_{\square}\big((W, f), (V, g)^{\psi}\big),
\end{equation*}
where $(V, g)^{\psi} \coloneq (V^{\psi}, g^{\psi})$ and $V^{\psi}(x,y)\coloneq V(\psi(x),\psi(y))$ and $ V^{\psi}(x)\coloneq g(\psi(y))$, where $S_{[0,1]}$ is the set of all Lebesgue measure-preserving bijections on $[0,1]$ up to null sets.

\paragraph{Compactness and covering number of the graphon-signal space}
At this point, note that $(\mathcal{WS}_r, \delta_{\square})$ is a pseudo-metric space. We denote the induced metric space on the space of $\delta_{\square}$-equivalence classes as $(\widetilde{\mathcal{WS}_r}, \delta_{\square})$. \citet{Lev+2023}, using a variant of the weak regularity lemma by \citet{weakReg}, showed that the metric space $(\widetilde{\mathcal{WS}_r}, \delta_{\square})$ is compact and derived an upper bound on its covering number.

\begin{theorem}[{\citep[Theorem 3.6]{Lev+2023}}]
	\label{lem:compact00}
	Let $r > 0$.
	The metric space $(\widetilde{\mathcal{WS}_r}, \delta_{\square})$ is compact.
	Moreover,  for every $c > 1$ and every sufficiently small $\epsilon > 0$, the space $\widetilde{\mathcal{WS}_r}$ can be covered by
	\begin{equation*}
		\label{eq:WLrCover0}
		\kappa(\epsilon) \leq 2^{k^2}
	\end{equation*}
	balls of radius $\epsilon$, where $\kappa(\epsilon) \coloneq \cN(\widetilde{\mathcal{WS}_r}, \delta_{\square}, \epsilon)$ and $k \coloneq \lceil 2^{2c/\epsilon^2} \rceil$.
\end{theorem}

\paragraph{Lipschitz continuity of MPNNs on graphon-signals}

To leverage the compactness result for analyzing MPNNs, \citet{Lev+2023} extended the definition of MPNNs to graphon-signals so that MPNNs on attributed graphs become a special case. In the context of the diagram in \cref{{fig:commutative_diagram}}, $\pi(G,\ell_G)=(W_{G},\vec{f}_{\ell_G})$ is the operator that induces graphon-signals from a graph signal or every MPNN without readout $h$, and for any labeled graph $(G,\ell_G)$, we have $\pi(h_{(G,\ell_G)}) = h_{\pi(G,\ell_G)}$.
As a result, for MPNNs with readout $h \in \MPNN^{\mathrm{ord}}_{L,M',L_{\mathsf{FNN}}}(\cX)$,\footnote{To be precise, \citet{Lev+2023} defined MPNNs on graphons through a slightly more general family of MPNNs.}
\begin{equation*}
	h_{(G,\ell_G)} = h_{(W_{G},\vec{f}_{\ell_G})}.
\end{equation*}
See \cref{sec:extended_graphon-theory} for a detailed description of MPNNs applied to graphon-signals and an extended analysis of graphon theory.

One of the main results of \citet{Lev+2023} is that MPNNs defined on graphon-signals are Lipschitz continuous concerning the graphon-signal cut distance. Furthermore, the Lipschitz constant of an MPNN can be bounded by the Lipschitz constants of its message functions.
It is important to note that in the extension from graphs to graphons, the aggregation step in MPNNs must always be normalized by order of the graph (i.e., $\MPNN^{\mathrm{ord}}_{L,M',L_{\mathsf{FNN}}}(\cX)$); otherwise, the commutative diagram (\cref{fig:commutative_diagram}) cannot be satisfied.

\paragraph{Generalization bound for MPNNs}
As a consequence of the Lipschitz continuity of MPNNs, the compactness of the space of graphon-signals, and the bound on its covering number, the following generalization theorem is established for the binary classification of attributed graphs.

\begin{theorem}[{\citet[Theorem 4.2]{Lev+2023}}]
	\label{thm:graphons_classification_generalization}
	Consider the hypothesis class $\cH$ of Lipschitz continuous MPNNs with Lipschitz constant $L_1$. Suppose the loss function is also Lipschitz continuous with Lipschitz constant $L_2$. Let $X_1,\ldots,X_N$ be independent random samples from a distribution $\mu$ on $\widetilde{\mathcal{WS}_r}$. Then, for every $\delta > 0$, with probability at least $1-\delta$, the following holds: for all $h \in \cH$
	\begin{align}
		\label{eq:gen00}
		\Big| \ell_{\text{emp}}(h) - \ell_{\text{exp}}(h) \Big| \leq \xi^{-1}(N/4) \Big( 2L + \frac{1}{\sqrt{2}} \big(L+M\big) \big(1+\sqrt{\log(2/\delta)}\big) \Big),
	\end{align}
	where $L \coloneq L_1L_2$, $M$ is an upper bound for the loss function, $\xi(\epsilon) \coloneq \frac{\kappa(\epsilon)^2\log(\kappa(\epsilon))}{\epsilon^2}$, $\kappa(\epsilon)$ is an upper bound on the covering number $\cN(\widetilde{\mathcal{WS}_r},\delta_{\square},\epsilon)$, and $\xi^{-1}$ is the inverse function of $\xi$.
\end{theorem}

Note that the term $\xi^{-1}(N/2C)$ in \cref{eq:gen00} decreases to zero as the size of the training set $N$ goes to infinity, but very slowly. The advantage of this theorem is its generality: it does not depend on any assumption on the data distribution. For example, the data distribution may be any arbitrary probability distribution supported on the subset of attributed graphs. This makes \cref{thm:graphons_classification_generalization} a general uniform generalization bound for MPNNs on attributed graphs. The disadvantage is that in such a general setting, the decay concerning the size of the training set $N$ is very slow.

\paragraph{Comparison to iterated degree measures approach}
\citet{Rac+2024} followed a similar approach by mapping the space of graphs into the space of distributions of iterated degree measures (DIDMs). They defined a metric, called the DIDM-mover's distance, that renders this space compact and introduced an MPNN framework on DIDMs in which MPNNs are Lipschitz continuous while satisfying the commutative diagram (\cref{fig:commutative_diagram}). We omit the details since this construction parallels the graphon-based approach and involves slightly complicated notation. However, we briefly compare the results of these two works. \citet{Lev+2023} derived an explicit bound on the covering number using the weak regularity lemma, whereas \citet{Rac+2024} only established the finiteness of the covering number through compactness. On the other hand, the graphon-signal cut distance underlies a finer topology than the metric used in \citet{Rac+2024},\footnote{Every open set w.r.t.\@ DIDM-mover's distance is also open w.r.t.\@ cut distance, but not vice-versa. Hence, every open covering w.r.t.\@ DIDM-mover's distance is also a covering w.r.t.\@ cut distance, but not vice-versa.} which may potentially mean that the covering number w.r.t.\@\@ the DIDM-mover's distance is smaller than that of cut distance.

\subsection{The non-uniform regime}
Below, we present an approach for deriving generalization bounds for MPNNs by restricting our focus to the space of $n$-order graphs and leveraging the notion of robustness from \cref{robustness_bound}. We begin by demonstrating how \cref{robustness_bound} can be transformed into a generalization bound that depends on the covering number of a finite space and the corresponding radius. We then present the covering number bounds derived in this framework, followed by the final form of the generalization bounds for MPNNs. For simplicity, we show only the case of graphs without node features, where the corresponding pseudometrics are more straightforward to define.

\paragraph{Unlabeled Graph Metrics} Following the works of \citet{Boe+21,Boe+2023}, we define the \new{tree distance} as
\begin{equation*}
	\delta_{\norm{\cdot}}^{\cT}(G,H) \coloneq \min_{\vec{S} \in D_n} \norm{\vec{A}(G) \vec{S} - \vec{S} \vec{A}(H)}_{\square},
\end{equation*}
where $ D_n $ denotes the set of $ n \times n $ doubly stochastic matrices (i.e., matrices with non-negative entries where each row and each column sums to 1), and the cut-norm on $ \Rb^{n \times n} $ is defined as
\begin{equation*}
	\norm{\vec{M}}_{\square} \coloneq \max_{S \subset [n], T \subset [n]} \Bigl| \sum_{i \in S, j \in T} m_{ij} \Bigr|.
\end{equation*}

\paragraph{Continuity of MPNNs on graphs} A slight difference with previous works is that in \citet{Vas+2024}, the MPNNs are required to satisfy a weaker notion of continuity than Lipschitz continuity, namely equicontinuity. The following result by \citet{Vas+2024} shows that the MPNN class $ \MPNN^{\mathrm{ord}}_{L,M',L_{\mathsf{FNN}}}(\cG_n)$ satisfies the equicontinuity property concerning the tree distance.

\begin{lemma}
	\label{thm:uniformcontinuityofmpnns}
	For all $ \varepsilon > 0 $, $ L \in \Nb $, there exists a function $ \gamma(\varepsilon) > 0 $ such that for $ n \in \mathbb{N} $, $ M' \in \Rb $, and $ G, H \in \cG_n $, and for all MPNN architectures in $ \MPNN^{\mathrm{ord}}_{L,M',L_{\mathsf{FNN}}}(\cG_n) $, the following implication holds,
	\begin{equation*}
		\delta_{\square}^{\cT}(G,H) < n^2 \cdot \gamma(\varepsilon) \implies \| \vec{h}_{G} - \vec{h}_{H} \|_2 \leq \varepsilon.
	\end{equation*}
\end{lemma}

Based on the above result, they define the following non-decreasing function $ \widebar{\gamma} \colon \Rb^{+} \cup \{ +\infty \} \to \Rb^{+} \cup \{ +\infty \} $ as $ \gamma(+\infty) = +\infty $, and
\begin{align*}
	\widebar{\gamma}(\epsilon) \coloneq \sup \left\{ \delta > 0 \mid \delta_{\square}^{\cT}(G,H) \leq n^2 \delta \implies \| h_G - h_H \| \leq \epsilon,  h \in \MPNN^{\mathrm{ord}}_{L,M',L_{\mathsf{FNN}}}(\cG_n), \ \forall G, H \in \cG_n \right\},
\end{align*}
and define its generalized inverse function $ \widebar{\gamma}^{\leftarrow} \colon \Rb^{+} \cup \{+\infty \} \to \Rb^{+} \cup \{+\infty \} $, where $ \Rb^{+} = (0, +\infty) $, as
\begin{equation*}
	\widebar{\gamma}^{\leftarrow}(y) \coloneq \inf \left\{ \epsilon > 0 \mid \gamma(\epsilon) \geq y \right\}.
\end{equation*}

Finally, by establishing that robustness is guaranteed by the equicontinuity property, they derive the following generalization bounds for the class $ \MPNN^{\mathrm{ord}}_{L,M',L_{\mathsf{FNN}}}(\cG_n) $, showing the connection between the covering number and the parameters in the definition of robustness in the binary classification setting using the binary cross entropy loss with the sigmoid function. That is $\ell \colon \Rb \times \{0,1\} \to \Rb$, $\ell(x,y)=y\log(\sigma(x)) + (1-y)\log(1-\sigma(x))$.
\begin{lemma}
	\label{proposition:robustnessfortreedistance(class)}
	For $ \varepsilon > 0 $, $ n, L \in \Nb $, and $ M' \in \Rb $, any graph learning algorithm for the class $ \MPNN^{\mathrm{ord}}_{L,M',L_{\mathsf{FNN}}}(\cG_n) $ is
	\begin{equation*}
		\Bigl( 2 \cN(\cG_n, \delta_{\square}^{\cT}, \varepsilon), L_{\ell} \cdot L_{\mathsf{FNN}} \cdot \widebar{\gamma}^{\leftarrow}\left( \frac{2\varepsilon}{n^2} \right) \Bigr)\text{-robust}.
	\end{equation*}
	Hence, for any sample $ \cS $ and $ \delta \in (0,1) $, with probability at least $ 1 - \delta $, we have
	\begin{equation*}
		|\ell_{\mathrm{exp}}(h_{\cS}) - \ell_{\mathrm{emp}}(h_{\cS})| \leq L_{\ell} \cdot L_{\mathsf{FNN}} \cdot \widebar{\gamma}^{\leftarrow}\left( \frac{2\varepsilon}{n^2} \right) + M \sqrt{\frac{4 \cN(\cG_n, \delta_{\square}^{\cT}, \varepsilon) \log(2) + 2 \log\left( \frac{1}{\delta} \right)}{|\cS|}},
	\end{equation*}
	where $ M $ is an upper bound for the loss function $ \ell $ and $ L_{\ell} $ is the Lipschitz constant of $ \ell(\cdot,y), y \in \{0, 1\} $.
\end{lemma}

\paragraph{Generalization bound}
Since computing the covering number as a function of $ \varepsilon $ is challenging due to the complex structure of this pseudo-metric space, \citet{Vas+2024} bounded the covering number by a function of $ \varepsilon $ and $m_n$ (i.e., the number of $ \wlone $-equivalence classes in $ \cG_n $) for a given $n$. This leads to a generalization bound that depends solely on $\varepsilon$, enabling the computation of the optimal radius by solving a minimization problem. Specifically, for the set of $ n $-order graphs $\cG_n$ and a radius of $\varepsilon = 4k$, they constructed a cover of $\cG_n$ with cardinality $\frac{m_n}{k+1}$, for all $k \in \Nb$, yielding the following family of generalization bounds.

\begin{theorem}
	\label{prop:extendedtreeconstruction}
	For $ n, L \in \Nb $, $ M' \in \Rb $, and any graph learning algorithm for $ \MPNN^{\mathrm{ord}}_{L,M',L_{\mathsf{FNN}}}(\cG_n) $, and for any sample $ \cS $ and $ \delta \in (0,1) $, with probability at least $ 1 - \delta $, we have
	\begin{equation*}
		\label{extendedtreeconstructionbounds}
		|\ell_{\mathrm{exp}}(h_{\cS}) - \ell_{\mathrm{emp}}(h_{\cS})| \leq L_{\ell} \cdot L_{\mathsf{FNN}} \cdot \widebar{\gamma}^{\leftarrow}\left( \frac{8k}{n^2} \right) + M \sqrt{\frac{ 4\frac{m_n}{k+1}\log(2) + 2\log\left( \frac{1}{\delta} \right)}{|\cS|}},
	\end{equation*}
	for $ k \in \Nb $, where $ M $ is an upper bound on the loss function $ \ell $.
\end{theorem}

The above bound shows a linear decrease in the covering number as a function of $ \varepsilon $. Furthermore, it is demonstrated that restricting the graph family makes it possible to achieve a bound that decreases exponentially with the radius. Additionally, they identify graph families for which the bounds can achieve a non-constant improvement compared to those in \citet{Mor+2023} (i.e., $ \frac{m_n}{n} $ instead of $ m_n $).

\paragraph{Vertex-attributed graphs and other Aggregation Functions}
To account for vertex-attributed graphs and consider the number of layers in an MPNN architecture, \citet{Vas+2024} define the \new{forest distance}. They show that this distance is an alternative but equivalent variation of the Tree Mover's distance introduced in \citet{chuang2022tree}, allowing them to devise an analogous version of \cref{proposition:robustnessfortreedistance(class)} for vertex-attributed graphs, taking layer information into account. Additionally, they derive a variant of $ \wlone $ that precisely captures the computation of the MPNN layers using mean aggregation, i.e., dividing by the number of neighbors after summing over the neighboring features, and they extend their generalization bounds to mean aggregation MPNNs.

Overall, the covering number bounds in this work are significantly tighter than those in \citet{Lev+2023}, which is expected due to the considerably more restricted space of graphs considered here. In practice, they experimentally show that these bounds closely approximate the actual test error observed on different datasets. These results also account for mean-aggregation MPNNs, which were not analyzed in the other works.

\subsection{Generalization bounds on mixtures of generating graphons}

The uniform convergence rate of \cref{thm:mpnn_generalization} is very slow under no assumption on the data distribution. To improve the asymptotics and obtain a bound, which can be meaningful in some practical settings,  \cite{Mas+2022,maskey2024generalization} introduces assumptions on the data distribution.

\paragraph{Graphons as generative graph models}
A graphon-signal can serve as a generative model of graphs. Given a graphon-signal $(W, \vec{f})$, a random graph with $n$ nodes is generated by sampling uniformly and independently the points $\{u_i\}_{i=1}^n$ from $[0,1]$, and connecting each pair $u_i, u_j$ with probability $W(u_i, u_j)$ to obtain the edges of the graph. The node feature of node $u_i$ is $\vec{f}(u_i)$, for $i \in [n]$.

\cite{Mas+2022} assumed that there is a finite set of template graphon-signals $(W_j,f_j)_{j=1}^\Gamma$, such that each graph from the data distribution is sampled independently by choosing randomly a template $j$, a degree $n$ (from a distribution of degrees $\nu$), and sampling a graph of $n$ nodes from $(W_j,f_j)$.
In fact, by the Aldous-Hoover theorem \citep{ALDOUS81,ALDOUS85,Hoover79,Hoover82}, any data distribution on graphs can be modeled this way if one allows a general ``infinite'' distribution of graphon-signal templates instead of a finite set.

In a classification setting, it is assumed that graphs sampled from the same template belong to the same class but not necessarily vice versa. \cite{maskey2024generalization} extends the analysis by considering sparse graphs sampled with noise from the template graphon-signals. This generative model of the data is called \emph{mixture of graphons}. A generalization theorem can then be derived from this setting. Let $\mu^N$ be the probability space of randomly sampled training sets of $N$ attributed graphs, $\mathrm{MPNN}_{L_{\varphi}}$ the hypothesis class of MPNN with Lipschitz message passing functions with Lipschitz constant $L_{\varphi}$,  $\nu$  the probability measure of the choice of the number of nodes $n$ of the graphs, $\ell$ the loss function, bounded by $M>0$ with Lipschitz constant $C_{\ell}$, $\varepsilon$ the noise rate when generating graphs, and $\alpha$ the sparsity level of the sampled graphs, where $n^{1-\alpha}$ is proportional to the average node degree; see \citep{maskey2024generalization} for more details.

\begin{theorem}[\citep{maskey2024generalization} Generalization of MPNN on mixtures of graphons]
	\label{thm:main_gen_bound_deformed_graphon}
	There exist constants $C, C'>0$ that describe the complexity of the hypothesis class of MPNNs in terms of their depth and the Lipschitz constants of their message functions, such that
	\begin{equation}
		\label{eq:main_gen_bound_deformed_graphon}
		\begin{aligned}    & \Eb_{\mu^m}\left[\sup_{f \in \mathrm{MPNN}_{L_{\varphi}}} \Big(\ell_{\text{emp}}(f)   - \ell_{\text{exp}}(f) \Big)^2  \right]   \leq   \frac{2^{\Gamma} M^2\pi}{N}
                  \\ &   + \frac{2^{\Gamma} C_{\ell}^2   }{N}  \left( C \cdot \Eb_{n\sim\nu} \left[ \frac{1+\log(n)}{n}n^{2\alpha} + \frac{1 + \log(n)}{n^{1/(D_{\chi} + 1)}}n^{2\alpha} + \mathcal{O}\left( \exp(-n) \right) \right] + C'\varepsilon\right),
		\end{aligned}
	\end{equation}
	where $D_{\chi}$ is a constant that characterizes the (Minkowski) dimension of the geometries described by the template graphons.
\end{theorem}

The proof of \cref{thm:main_gen_bound_deformed_graphon} is based on a covering number construction, a special case of \cref{robustness_bound} first proven in \citep[Appendix C.2]{maskey2024generalization}.
Note that the first term in the right-hand side of \cref{eq:main_gen_bound_deformed_graphon}  is typically much smaller than the second term, as it does not depend on the complexity $C$ of the hypothesis class.
Hence, one insight that can be derived from \cref{eq:main_gen_bound_deformed_graphon} is that MPNNs generalize better the larger the graphs in the data distribution are. Namely, even if the complexity of the MPNNs $C$ is higher than the sample size $m$, one can still generalize well if the average number of nodes $n$ is high. The disadvantage of this bound is that it is specific for a mixture of a graphon model, which may not capture the behavior of some real-life data distributions if the number of classes $K$ is small.

\section{Transduction in graph learning}
\label{sec:transduction}
Up to this point, our analysis of the generalization performance of MPNNs has primarily focused on graph-level prediction tasks. However, node-level prediction problems are equally important in real-world applications, such as social networks, recommendation systems, and more \citep{DBLP:journals/csur/WuSZXC23}. \citet{scarselli2018vapnik, Ver+2019, Gar+2020, Zhou2021} describe how their generalization bounds can be extended to node classification tasks by converting graphs into subgraphs via local node-wise computation trees and treating them as i.i.d.\ samples. This assumption does not align
with the commonly seen graph-based semi-supervised learning setting. Therefore, more recent research has derived generalization bounds for MPNNs in node classification under a \new{transductive learning approach}. In the first (to our knowledge) work on a transductive learning approach to node classification~\citep{Oono2020}, they provide generalization bounds via the transductive Rademacher complexity for \new{multi-scale MPNNs}. Multi-scale MPNNs are MPNN architectures designed to mitigate the over-smoothing issue commonly observed in MPNNs. Similarly, \citet{DBLP:conf/nips/CongRM21} established generalization bounds for GNNs by leveraging transductive uniform stability under the full-batch gradient descent. Further, \citet{Ess+2021} analyzed the generalization properties of graph convolutional networks via transductive Rademacher complexity, highlighting the interplay of graph structure and the predictive performance of MPNNs. \citet{Tan+2023} also developed generalization bounds in the transductive setting, applicable to models trained with the SGD. Before we define the statistical framework of transductive learning and present some well-established generalization bounds, let us briefly highlight the core distinction between transductive and inductive learning. Inductive learning, described so far, aims to learn a function $h$ from a training dataset and then applies $h$ to new, unseen data, emphasizing the generalization beyond the training examples. In contrast, transductive learning is tailored to make predictions specifically for a fixed set of unseen data available during training, focusing on optimizing predictions only for this set.

In the following, we present the transductive learning framework for graphs, defining the transductive Rademacher complexity as a modification of the traditional Rademacher complexity introduced in \cref{rademacherbounds} and explaining the key motivations behind this extension. We also present the complexity bounds from \citet{Ess+2021} as a specific example. We choose to highlight this work due to the importance of the transductive Rademacher complexity, a fundamental concept in transductive learning, as described in \citep{DBLP:journals/tit/Koltchinskii01,Yan+2009}.

\paragraph{Transductive learning statistical framework} Let $G \in \cG_{n,d}^{\Rb}$. For ease of notation, we assume $V(G) = [n]$, and we define the matrix $\vec{X} \in \Rb^{n \times d}$, with rows $\vec{x}_i$ representing the feature vectors of nodes $i \in [n]$. Without loss of generality, for some $m \in [n]$, we set $u=n-m$, we assume that the labels $y_1, \ldots, y_m \in \{-1, +1\}$ are known, and the goal is to predict $y_{m+1}, \ldots, y_n$. Usually, in the transductive framework, we do not assume that the data is sampled from an unknown underlying distribution. Instead, we introduce randomness in the data generation process as follows. We assume that the labeled set $\{(\vec{x}_i, y_i)\}_{i=1}^{m}$ is randomly sampled from all possible $m$-size subsets of $\{(\vec{x}_i, y_i)\}_{i=1}^{n}$. Or equivalently, we may assume that a random permutation of $\{(\vec{x}_i, y_i)\}_{i=1}^{n}$ is chosen, and the first $m$ pairs are selected as the training set, while the remaining pairs form the test set. As a result, the training data is no longer independently sampled, and the underlying distribution is known. Note that the learner still has access to the feature vectors $\{\vec{x}_i\}_{i=m+1}^{n}$. See \citep[Section 10.1]{DBLP:books/sp/Vapnik06}, \citep{DBLP:phd/il/Pechyony08}, for diﬀerent mechanisms of training set generation and diﬀerent learning goals. For a given loss function $\ell$ defined as previously, we define the test error of a predictor $h$ in some hypothesis class $\cH$ as $\ell_u(h) \coloneq \frac{1}{n-m} \sum_{i=m+1}^{n} \ell(h, \vec{x}_i, y_i)$ and the empirical error as $\ell_{\text{emp}}(h) \coloneq \frac{1}{m} \sum_{i=1}^{m} \ell(h, \vec{x}_i, y_i)$. The goal is then to bound their difference, i.e., deriving bounds of the form
\begin{equation*}
	\ell_u(h) \leq \ell_{\text{emp}}(h) + \text{complexity}(\cH,G).
\end{equation*}

\paragraph{Comparison to inductive learning} Before proceeding to the transductive Rademacher complexity bounds derived by \citet{Ess+2021}, we provide some remarks on key similarities and differences between transductive and inductive learning.  First, note that the data is assumed to be sampled independently in the inductive scenario, which is not valid in the transductive scenario. In both settings, the training points are sampled from some distribution.  However, in the inductive case, this unknown distribution is defined over the arbitrary product space $\mathcal{X} \times \mathcal{Y}$. In contrast, in the transductive setting, the distribution is known and is defined over a finite subset $\{(\vec{x}_i,y_i)\}_{i=1}^{n}$. In transductive learning, we also evaluate the error between the empirical loss and the loss over a fixed test set, denoted by $\ell_{u}$, which strongly depends on the training sample. In contrast, in the inductive setting, the error is measured between the empirical loss and the expected loss, which is an unknown but constant quantity.  Thus, in the transductive setting, we aim to bound the difference between two random variables, i.e., $\ell_{\text{emp}} - \ell_{u}$, while in the inductive setting, the objective is to bound the difference between a random variable and its expectation, i.e., $\ell_{\text{emp}} - \ell_{\text{exp}}$.

\paragraph{Transductive Rademacher complexity bounds} Assume the binary classification node level setting and, with a slight abuse of notation, that the hypothesis class consists of functions from $\vec{X}$ to $\{-1, 1\}^n$. Furthermore, since typically neural networks output a soft predictor in $\Rb$ transformed to labels via a sign function, we focus on predictors $h \colon \Rb^{n \times d} \to \Rb^{n}$. Following \citet{Ess+2021}, for some $L \in \Nb$, we consider the class of $L$-layer MPNNs with the following architecture. In each layer $t$, we compute the updated node features $\vec{g}_t \in \mathbb{R}^{n \times d_t}$, where $d_t$ is the dimension of layer $t$ as follows. For $\vec{S} \coloneq \vec{A}(G) + \vec{I}_{n \times n}$, with $\vec{A}(G)$ being the adjacency matrix of the graph $G$, and $\phi$ denote point-wise activation function, which we assume to be a Lipschitz function with Lipschitz constant $L_\phi$, we define the hypothesis class over all $L$-layer MPNNs as
\begin{equation*}
	\cH^\phi_G \coloneq \left\{ h^\phi(\vec{X}) = \vec{g}_L \colon \mathbb{R}^{n \times d} \rightarrow \Rb^{n} \right\}
\end{equation*}
with
\begin{equation*}
	\vec{g}_t \coloneq \phi \left( \vec{b}_t + \vec{S} \vec{g}_{t-1}\vec{W}_t \right), \quad t \in [L], \quad \vec{g}_0 \equiv \vec{X},
\end{equation*}
where $\vec{W}_t \in \mathbb{R}^{d_{t-1} \times d_t}$ and $\vec{b}_t \in \mathbb{R}^{n}$ are the trainable weight matrix and bias term, respectively. Through the message-passing process, it becomes clear that the graph structure is treated as part of the hypothesis class through the adjacency matrix. By further assuming that the norms $\|\vec{b}_t\|_1$ and $\|\vec{W}_t\|_{\infty}$ of the trainable parameters are bounded by $\beta$ and $\omega \in \mathbb{R}$, respectively, for all $t \in [L]$, we denote our hypothesis class with $\cH^{\phi,\beta,\omega}_G$.

We now define the \new{transductive Rademacher complexity} and describe the differences between the inductive Rademacher complexity introduced in \cref{rademacherbounds}.

\begin{definition}[TRC \citep{Yan+2009}]
	Let $\cV \subset \mathbb{R}^n$, $p \in [0,0.5]$, and $m$ be the number of labeled points. Let $\vec{\sigma} = (\sigma_1, \ldots, \sigma_n)^\tran$ be a vector of independent and identically distributed random variables, each taking the values $+1$ and $-1$ with probability $p$, and $0$ with probability $1-2p$. The \new{transductive Rademacher complexity} (TRC) of $\cV$ is defined as
	\begin{equation*}
		\cR_{m,n}(\cV) \coloneq \left( \frac{1}{m} + \frac{1}{n-m} \right) \cdot \Eb_{\sigma} \left[ \sup_{\vec{v} \in \cV} \vec{\sigma}^T \vec{v} \right].
	\end{equation*}
\end{definition}

Some key differences from inductive Rademacher complexity include that transductive Rademacher complexity depends on both the training and the test points. In contrast, the inductive complexity only depends on training points. Additionally, the transductive complexity is an empirical quantity that does not depend on any underlying distribution, only on the known distribution of $\sigma$.

\citet{Yan+2009} established a generalization error bound through the Transductive Rademacher Complexity (TRC) that is similar to \cref{thm:rademachercomplexitybounds}. Subsequently, \citet{Ess+2021} bounded the TRC of $\cH^{\phi,\beta,\omega}_G$ and, using this result, showed the following generalization bound:

\begin{theorem}[\citep{Ess+2021}]
	For the transductive Rademacher complexity of the hypothesis class $\cH^{\phi,\beta,\omega}_G$, we have the following bound,
	\begin{equation*}
		\cR_{m,n}(\cH^{\phi,\beta,\omega}_G) \leq \frac{c_1 n^2}{m(n-m)}\left( \sum_{t=0}^{L-1} c_2^{t}\|\vec{S}\|_{\infty}^{t}\right) + c_3 c_2^{L} \|\vec{S}\|_{\infty}^{L} \|\vec{S}\vec{X}\|_{2\rightarrow\infty}\sqrt{\log(n)},
	\end{equation*}
	where $c_1 \coloneq 2L_{\phi}\beta$, $c_2 \coloneq 2L_{\phi}\omega$, $c_3 \coloneq L_{\phi}\omega\sqrt{\sfrac{2}{d}}$, and $\| \cdot \|_{2\rightarrow\infty}$ denotes the maximum 2-norm of any column. Following \citet{Yan+2009}, the generalization error bound becomes, for any $\delta > 0$, with probability at least $1-\delta$,
	\begin{equation*}
		\ell_{\text{u}}(h) - \ell_{\text{emp}}(h) \leq \cR_{m,n}(\cH^{\phi,\beta,\omega}_G) + c_4 \frac{n\sqrt{\min\{m,n-m\}}}{m(n-m)} + c_5 \sqrt{\frac{n}{m(n-m)}\ln\left(\frac{1}{\delta}\right)},
	\end{equation*}
	where $c_4,c_5$ are absolute constants such that $c_4<5.05$, and $c_5<0.8$.
\end{theorem}

The remainder of this work focuses on understanding how the graph structure influences the generalization performance by analyzing the dependence of the above bound on $\vec{S}$ and $\vec{X}$.

\section{Various frameworks}

So far, we have reviewed works analyzing the generalization abilities of MPNNs through common theoretical frameworks such as VC dimension, margin-based bounds, Rademacher complexity, PAC-Bayesian framework, stability-based analysis, graphon theory, and transductive learning. However, novel approaches beyond these established frameworks also yield significant generalization results.

In this section, we focus on research that investigates the generalization properties of GNNs, specifically using methods that diverge from the ones previously mentioned. For instance, \citet{Zhang2020} approaches generalization by assuming the existence of a ground truth MPNN model with zero generalization error. They propose a gradient-based optimizer to learn this model and guarantee MPNN generalizability. One limitation of this work is that it only considers models with a single hidden layer. \citet{DBLP:conf/icml/LiWLCX22} introduce a different framework, combining SGD with graph topology sampling techniques. They provide generalization bounds for GCNs with up to three layers, primarily in the context of node classification tasks. \citet{Ma2021} employ a PAC-Bayesian approach in the semi-supervised setting, deriving generalization bounds for node classification under non-i.i.d.\@ assumptions. Similarly, \citet{DBLP:conf/icml/LeeHW24} apply the PAC-Bayesian framework to derive generalization bounds for a wide range of knowledge graph representation learning methods \citep{DBLP:conf/nips/BordesUGWY13, DBLP:conf/kdd/ChungLW23}. In a novel study, \citet{DBLP:conf/icml/SuzukiNS00Y23} modify the Rademacher complexity to derive generalization bounds for GNNs using non-Euclidean metric spaces for graph embeddings. In another recent work, \citet{Shi2024} take a step beyond the standard statistical learning framework by utilizing random matrix theory to account for phenomena like the double descent, establishing generalization bounds for GNNs. Furthermore, \citet{DBLP:journals/corr/abs-2402-03818} analyze the generalization performance of single-layer GCNs in the high-dimensional setting, using data generated by stochastic block models. \cite{Li+2024} build on the margin-based generalization framework proposed by \cite{Chuang+2021}, which is based on $k$-Variance and the Wasserstein distance. They provide a method to analyze how expressiveness affects graph embeddings' inter- and intra-class concentration. \citet{Pel+2024} investigate node-individualized MPNNs via the covering number framework~\citep{Xu+2012} outlined in~\cref{sec:coveringnumberbounds}. \citet{Kri+2018} leverage results from graph property testing~\citep{Gol2010} to study the sample complexity of learning to distinguish various graph properties, e.g., planarity or triangle freeness, using graph kernels~\citep{Borg+2020,Kri+2019}. \cite{xu20reasoning} formulate an early framework for learning algorithmic tasks with intermediate supervision. Finally, \citet{keriven2020convergence} investigate the generalization capabilities of GNNs on large random graphs.

\section{Out-of-distribution generalization}

So far, we have discussed the generalization analysis of GNNs in scenarios where both the training and test graphs are sampled from the same distribution, known as \new{in-distribution generalization}. However, in practice, it is often the case that the training graphs are drawn from a distribution that slightly differs from the distribution of the graphs where the GNN is expected to perform well. This challenge arises due to a distribution shift between the training and test sets, emphasizing the importance of studying \new{out-of-distribution} (OOD) generalization. Hence, in this section, we highlight key works that address the OOD generalization capabilities of GNNs. We first describe the statistical learning setting, emphasizing the differences between in- and out-of-distribution generalization problems.

\paragraph{Statistical learning setting}
Let $\cG$ be a graph input space and $\mathcal{Y}$ be the label space. We focus on predictors $f_{\theta} \colon \cG \to \cY$, with learnable parameters $\theta$ and a loss function $\ell$. While we primarily consider graph-level tasks, it is straightforward that the input space could consist of nodes or edges for node-level or edge-level tasks, respectively. Given a training set $\cS = \{(G_i, y_i)\}_{i=1}^{N}$ of $N$ i.i.d.\@ samples from a distribution $P_{\text{train}}$ on $\cG \times \cY$, the goal is to learn an optimal predictor $f_{\theta}^{*}$ that minimizes the expected loss on data drawn from a different test distribution $P_{\text{test}}$, where $P_{\text{train}} \neq P_{\text{test}}$, i.e.,
\begin{equation*}
	f_{\theta}^{*} \coloneq \arg \min_{f_{\theta}} \Eb_{P_{\text{test}}} \left[ \ell(f_{\theta}(G), y) \right].
\end{equation*}

Since the assumption that the training and test data are i.i.d.\@ no longer holds, traditional methods for bounding the generalization gap using concentration inequalities are not applicable. Consequently, less theoretical work has been conducted on out-of-distribution (OOD) generalization than in-distribution generalization. Even defining the generalization gap in this setting is unclear due to two distinct underlying distributions.  \citet{DBLP:conf/nips/YeXCLLW21} developed a rigorous and quantitative framework for defining OOD generalization using tools from information theory.
\citet{DBLP:journals/corr/abs-2202-07987} provide a comprehensive overview of various OOD generalization approaches in graph learning. By viewing graph data as a distribution over computation graphs and using optimal transport, \citet{chuang2022tree} quantified the effect of input perturbations through a Lipschitz constant and a domain generalization bound. Following a causal machine learning approach, \citet{DBLP:conf/nips/GuiLLLJ23} analyzed the out-of-distribution generalization of GNNs. Other works analyzing OOD generalization include \citet{xu2021how}, which provided sufficient conditions (on the architecture and task) under which GNNs generalize well, and \citet{DBLP:conf/icml/LiGLJ24}, which investigated data augmentation approaches.

A concept closely related to, but distinct from, out-of-distribution generalization is \new{size-generalization} or \new{size-transferabilitya}. Size-transferability analyzes the ability of GNNs to generalize across different graph sizes. It addresses the question: \say{When do GNNs generalize to graphs of sizes not seen during training?} Continuous extensions of graphs, such as graphons and graphops, appear to be appropriate tools for analyzing the size-transferability of GNNs. These continuous counterparts of graphs retain structural information without being affected by size variations. Some of the most detailed analyses of size-transferability include \citet{DBLP:journals/corr/abs-1901-10524,Ruisz+2020,Levie+2022,le23transfer}. \cite{herbst25higher} analyze transferability of higher-order graph neural networks. Taking a different approach, \citet{Yeh+2021} studied graph distributions in which local structures (which GNNs capture) significantly depend on graph size. Specifically, they used $d$-patterns (a notion closely related to the $k$-neighborhood of a node) to show that a $d$-pattern discrepancy between the training graphs and the test set can lead to poor global minima, ultimately failing to generalize to larger graphs. Furthermore, \citet{DBLP:conf/icml/BevilacquaZ021} investigated size generalization using a causal machine learning approach, assuming independence between cause and mechanism, i.e., $P_{\text{train}}(Y|X) = P_{\text{test}}(Y|X)$, drawing on principles from \citet{DBLP:conf/nips/LouizosSMSZW17}.

\section{Future directions and open problems}

The present work shows that the last years resulted in many works studying MPNNs' generalization properties. However, many open problems and challenges remain. First, it is valuable to understand if tighter generalization can be obtained when restricting the analysis to specific graph classes, e.g., trees or planar graphs. As a first step, one could tailor the results of~\cite{Mor+2023} to specific graph classes. Moreover, most generalization bounds, e.g., based on VC dimension theory or Rademacher averages, only consider simplistic graph parameters such as the maximum degree. Hence, it is essential to include more structural information in the graph. In addition, only some works study the generalization properties of more expressive GNNs beyond \wlone's limitations. While~\cite{Fra+2024} took a first step in that direction by studying simple, more expressive MPNNs, including subgraph information, the paper only studies the linearly separable case. Hence, it would also be valuable to understand under which conditions adding more expressive power to MPNNs leads to better generalization performance. Since most of the discussed bounds are based on VC dimension theory or a similar framework, they are vacuous,  not implying much for practical considerations. Hence, it is essential to adapt learning-theoretic generalization bounds, such as information theoretic~\citep{Chu+2023} or compression-based frameworks~\citep{Aro+2018}, to the graph domain and understand to what extent one can incorporate graph structure information in such bounds. We also refer the reader to~\citet{Mor+2024b}, which provides an insightful resource for open problems and challenges in the foundations of graph learning.

\section{Conclusion}
Here, we reviewed the current state of generalization theory for graph learning. Specifically, we overviewed generalization theory for MPNNs using the VC dimension, Rademacher complexity, PAC-Bayesian theory, and covering numbers. Moreover, we reviewed generalization bounds based on the stability of MPNNs on graphs. In addition, we surveyed recent work on out-of-distribution generalization. Finally, we discussed open problems and challenges to advance our understanding of MPNNs' generalization. We hope that this paper presents a valuable handbook for the generalization theory of MPNNs and helps advance our understanding further.

\section*{Acknowledgements}
Antonis Vasileuou is supported by the German Research Foundation (DFG) within Research Training Group 2236/2 (UnRAVeL). Christopher Morris is partially funded by a DFG Emmy Noether grant (468502433) and RWTH Junior Principal Investigator Fellowship under Germany’s Excellence Strategy.
Ron Levie is supported by a grant from the United States-Israel Binational Science Foundation (BSF), Jerusalem, Israel, and the United States National Science Foundation (NSF), (NSF-BSF, grant No. 2024660), and by the Israel Science Foundation (ISF grant No. 1937/23). Stefanie Jegelka is supported by an Alexander von Humboldt professorship.

\bibliography{bibliography}

\begin{thebibliography}{144}
\providecommand{\natexlab}[1]{#1}
\providecommand{\url}[1]{\texttt{#1}}
\expandafter\ifx\csname urlstyle\endcsname\relax
  \providecommand{\doi}[1]{doi: #1}\else
  \providecommand{\doi}{doi: \begingroup \urlstyle{rm}\Url}\fi

\bibitem[Aldous(1981)]{ALDOUS81}
D.~J. Aldous.
\newblock Representations for partially exchangeable arrays of random
  variables.
\newblock \emph{Journal of Multivariate Analysis}, 11\penalty0 (4):\penalty0
  581--598, 1981.

\bibitem[Aldous(1985)]{ALDOUS85}
D.~J. Aldous.
\newblock Exchangeability and related topics.
\newblock In \emph{{\'E}cole d'{\'E}t{\'e} de Probabilit{\'e}s de Saint-Flour
  XIII --- 1983}, pages 1--198, 1985.

\bibitem[Alon et~al.(2021)Alon, Hanneke, Holzman, and Moran]{Alo+2021}
N.~Alon, S.~Hanneke, R.~Holzman, and S.~Moran.
\newblock A theory of {PAC} learnability of partial concept classes.
\newblock In \emph{FOCS}, 2021.

\bibitem[Anthony and Bartlett(2002)]{Ant+2002}
M.~Anthony and P.~L. Bartlett.
\newblock \emph{Neural Network Learning - Theoretical Foundations}.
\newblock Cambridge University Press, 2002.

\bibitem[Arora et~al.(2018)Arora, Ge, Neyshabur, and Zhang]{Aro+2018}
S.~Arora, R.~Ge, B.~Neyshabur, and Y.~Zhang.
\newblock Stronger generalization bounds for deep nets via a compression
  approach.
\newblock In \emph{ICML}, 2018.

\bibitem[Arvind et~al.(2015)Arvind, K{\"{o}}bler, Rattan, and
  Verbitsky]{Arv+2015}
V.~Arvind, J.~K{\"{o}}bler, G.~Rattan, and O.~Verbitsky.
\newblock On the power of color refinement.
\newblock In \emph{FCT}, pages 339--350, 2015.

\bibitem[Azizian and Lelarge(2021)]{Azi+2020}
W.~Azizian and M.~Lelarge.
\newblock Characterizing the expressive power of invariant and equivariant
  graph neural networks.
\newblock In \emph{ICLR}, 2021.

\bibitem[Babai and Kucera(1979)]{Bab+1979}
L.~Babai and L.~Kucera.
\newblock Canonical labelling of graphs in linear average time.
\newblock In \emph{FOCS}, 1979.

\bibitem[Bartlett and Mendelson(2001)]{DBLP:conf/colt/BartlettM01}
P.~L. Bartlett and S.~Mendelson.
\newblock Rademacher and gaussian complexities: Risk bounds and structural
  results.
\newblock In \emph{COLT}, 2001.

\bibitem[Bartlett et~al.(2017)Bartlett, Foster, and Telgarsky]{Bar+2017}
P.~L. Bartlett, D.~J. Foster, and M.~Telgarsky.
\newblock Spectrally-normalized margin bounds for neural networks.
\newblock In \emph{NeurIPS}, 2017.

\bibitem[Bevilacqua et~al.(2021)Bevilacqua, Zhou, and
  Ribeiro]{DBLP:conf/icml/BevilacquaZ021}
B.~Bevilacqua, Y.~Zhou, and B.~Ribeiro.
\newblock Size-invariant graph representations for graph classification
  extrapolations.
\newblock In \emph{ICML}, 2021.

\bibitem[Bianchini and Scarselli(2014)]{DBLP:journals/tnn/BianchiniS14}
M.~Bianchini and F.~Scarselli.
\newblock On the complexity of neural network classifiers: {A} comparison
  between shallow and deep architectures.
\newblock \emph{{IEEE} Transactions on Neural Networks and Learning Systems},
  25\penalty0 (8):\penalty0 1553--1565, 2014.

\bibitem[B{\"{o}}ker(2021)]{Boe+21}
J.~B{\"{o}}ker.
\newblock Graph similarity and homomorphism densities.
\newblock In N.~Bansal, E.~Merelli, and J.~Worrell, editors, \emph{ICALP},
  2021.

\bibitem[Bordes et~al.(2013)Bordes, Usunier, Garc{\'{\i}}a{-}Dur{\'{a}}n,
  Weston, and Yakhnenko]{DBLP:conf/nips/BordesUGWY13}
A.~Bordes, N.~Usunier, A.~Garc{\'{\i}}a{-}Dur{\'{a}}n, J.~Weston, and
  O.~Yakhnenko.
\newblock Translating embeddings for modeling multi-relational data.
\newblock In \emph{NeurIPS}, 2013.

\bibitem[Borgwardt et~al.(2020)Borgwardt, Ghisu, Llinares{-}L{\'{o}}pez,
  O'Bray, and Rieck]{Borg+2020}
K.~M. Borgwardt, M.~E. Ghisu, F.~Llinares{-}L{\'{o}}pez, L.~O'Bray, and
  B.~Rieck.
\newblock Graph kernels: State-of-the-art and future challenges.
\newblock \emph{Foundations and Trends in Machine Learning}, 13\penalty0
  (5–6), 2020.

\bibitem[Boucheron et~al.(2003)Boucheron, Lugosi, and
  Bousquet]{DBLP:conf/ac/BoucheronLB03}
S.~Boucheron, G.~Lugosi, and O.~Bousquet.
\newblock Concentration inequalities.
\newblock In \emph{Advanced Lectures on Machine Learning, {ML} Summer Schools
  2003}, 2003.

\bibitem[Bouritsas et~al.(2020)Bouritsas, Frasca, Zafeiriou, and
  Bronstein]{Bou+2020}
G.~Bouritsas, F.~Frasca, S.~Zafeiriou, and M.~M. Bronstein.
\newblock Improving graph neural network expressivity via subgraph isomorphism
  counting.
\newblock \emph{arXiv preprint}, 2020.

\bibitem[Bousquet and Elisseeff(2002)]{stabilityoriginal}
O.~Bousquet and A.~Elisseeff.
\newblock Stability and generalization.
\newblock \emph{Journal of Machine Learning Research}, 2:\penalty0 499--526, 06
  2002.

\bibitem[Bousquet et~al.(2003)Bousquet, Boucheron, and
  Lugosi]{DBLP:conf/ac/BousquetBL03}
O.~Bousquet, S.~Boucheron, and G.~Lugosi.
\newblock Introduction to statistical learning theory.
\newblock In \emph{Advanced Lectures on Machine Learning, {ML} Summer Schools
  2003}, 2003.

\bibitem[Böker et~al.(2023)Böker, Levie, Huang, Villar, and Morris]{Boe+2023}
J.~Böker, R.~Levie, N.~Huang, S.~Villar, and C.~Morris.
\newblock Fine-grained expressivity of graph neural networks.
\newblock In \emph{NeurIPS}, 2023.

\bibitem[Cai et~al.(1992)Cai, F{\"{u}}rer, and Immerman]{Cai+1992}
J.~Cai, M.~F{\"{u}}rer, and N.~Immerman.
\newblock An optimal lower bound on the number of variables for graph
  identifications.
\newblock \emph{Combinatorica}, 12\penalty0 (4):\penalty0 389--410, 1992.

\bibitem[Cappart et~al.(2021)Cappart, Ch{\'{e}}telat, Khalil, Lodi, Morris, and
  Veli\v{c}kovi\'{c}]{Cap+2021}
Q.~Cappart, D.~Ch{\'{e}}telat, E.~B. Khalil, A.~Lodi, C.~Morris, and
  P.~Veli\v{c}kovi\'{c}.
\newblock Combinatorial optimization and reasoning with graph neural networks.
\newblock In \emph{IJCAI}, 2021.

\bibitem[Chen et~al.(2020)Chen, Li, and Zhao]{Chen2020}
M.~Chen, X.~Li, and T.~Zhao.
\newblock On generalization bounds of a family of recurrent neural networks.
\newblock In \emph{AISTATS}, 2020.

\bibitem[Chen et~al.(2019)Chen, Villar, Chen, and Bruna]{Che+2019}
Z.~Chen, S.~Villar, L.~Chen, and J.~Bruna.
\newblock On the equivalence between graph isomorphism testing and function
  approximation with {GNNs}.
\newblock In \emph{NeurIPS}, pages 15868--15876, 2019.

\bibitem[Chu and Raginsky(2023)]{Chu+2023}
Y.~Chu and M.~Raginsky.
\newblock A unified framework for information-theoretic generalization bounds.
\newblock In \emph{NeurIPS}, 2023.

\bibitem[Chuang and Jegelka(2022)]{chuang2022tree}
C.-Y. Chuang and S.~Jegelka.
\newblock Tree mover's distance: Bridging graph metrics and stability of graph
  neural networks.
\newblock \emph{NeurIPS}, 2022.

\bibitem[Chuang et~al.(2021)Chuang, Mroueh, Greenewald, Torralba, and
  Jegelka]{Chuang+2021}
C.~Y. Chuang, Y.~Mroueh, K.~Greenewald, A.~Torralba, and S.~Jegelka.
\newblock Measuring generalization with optimal transport.
\newblock In \emph{NeurIPS}, volume~10, 2021.

\bibitem[Chung et~al.(2023)Chung, Lee, and Whang]{DBLP:conf/kdd/ChungLW23}
C.~Chung, J.~Lee, and J.~J. Whang.
\newblock Representation learning on hyper-relational and numeric knowledge
  graphs with transformers.
\newblock In \emph{KDD}, 2023.

\bibitem[Cong et~al.(2021)Cong, Ramezani, and Mahdavi]{DBLP:conf/nips/CongRM21}
W.~Cong, M.~Ramezani, and M.~Mahdavi.
\newblock On provable benefits of depth in training graph convolutional
  networks.
\newblock In \emph{NeurIPS}, 2021.

\bibitem[Corso et~al.(2024)Corso, Stark, Jegelka, Jaakkola, and
  Barzilay]{Cor+2024}
G.~Corso, H.~Stark, S.~Jegelka, T.~Jaakkola, and R.~Barzilay.
\newblock Graph neural networks.
\newblock \emph{Nature Reviews Methods Primers}, 4\penalty0 (1):\penalty0 17,
  2024.

\bibitem[Dai et~al.(2016)Dai, Dai, and Song]{Dai+2016}
H.~Dai, B.~Dai, and L.~Song.
\newblock Discriminative embeddings of latent variable models for structured
  data.
\newblock In \emph{ICLR}, 2016.

\bibitem[Daniëls and Geerts(2024)]{Daniels+2024}
N.~Daniëls and F.~Geerts.
\newblock A note on the {VC} dimension of $1$-dimensional {GNN}s.
\newblock \emph{arXiv preprint}, 2024.

\bibitem[D'Inverno et~al.(2024)D'Inverno, Bianchini, and
  Scarselli]{dinverno2024vc}
G.~A. D'Inverno, M.~Bianchini, and F.~Scarselli.
\newblock {VC} dimension of graph neural networks with {Pfaffian} activation
  functions.
\newblock \emph{arXiv preprint}, 2024.

\bibitem[Duranthon and
  Zdeborov{\'{a}}(2024)]{DBLP:journals/corr/abs-2402-03818}
O.~Duranthon and L.~Zdeborov{\'{a}}.
\newblock Asymptotic generalization error of a single-layer graph convolutional
  network.
\newblock \emph{arXiv preprint}, 2024.

\bibitem[Duval et~al.(2023)Duval, Mathis, Joshi, Schmidt, Miret, Malliaros,
  Cohen, Lio, Bengio, and Bronstein]{Duv+2023}
A.~Duval, S.~V. Mathis, C.~K. Joshi, V.~Schmidt, S.~Miret, F.~D. Malliaros,
  T.~Cohen, P.~Lio, Y.~Bengio, and M.~M. Bronstein.
\newblock A hitchhiker's guide to geometric {GNNs} for {3D} atomic systems.
\newblock \emph{arXiv preprint}, 2023.

\bibitem[Easley and Kleinberg(2010)]{Eas+2010}
D.~Easley and J.~Kleinberg.
\newblock \emph{Networks, Crowds, and Markets: Reasoning About a Highly
  Connected World}.
\newblock Cambridge University Press, 2010.

\bibitem[El{-}Yaniv and Pechyony(2007)]{Yan+2009}
R.~El{-}Yaniv and D.~Pechyony.
\newblock Transductive rademacher complexity and its applications.
\newblock In \emph{COLT}, 2007.

\bibitem[Esser et~al.(2021)Esser, Vankadara, and Ghoshdastidar]{Ess+2021}
P.~M. Esser, L.~C. Vankadara, and D.~Ghoshdastidar.
\newblock Learning theory can (sometimes) explain generalisation in graph
  neural networks.
\newblock In \emph{NeurIPS}, 2021.

\bibitem[Feng et~al.(2022)Feng, Chen, Li, Sarkar, and Zhang]{Fen+2022}
J.~Feng, Y.~Chen, F.~Li, A.~Sarkar, and M.~Zhang.
\newblock How powerful are k-hop message passing graph neural networks.
\newblock In \emph{NeurIPS}, 2022.

\bibitem[Franks et~al.(2023)Franks, Morris, Velingker, and Geerts]{Fra+2024}
B.~J. Franks, C.~Morris, A.~Velingker, and F.~Geerts.
\newblock Weisfeiler--leman at the margin: When more expressivity matters.
\newblock \emph{arXiv preprint}, 2023.

\bibitem[Frieze and Kannan(1999)]{weakReg}
A.~M. Frieze and R.~Kannan.
\newblock Quick approximation to matrices and applications.
\newblock \emph{Combinatorica}, 19:\penalty0 175--220, 1999.

\bibitem[Garg et~al.(2020)Garg, Jegelka, and Jaakkola]{Gar+2020}
V.~K. Garg, S.~Jegelka, and T.~S. Jaakkola.
\newblock Generalization and representational limits of graph neural networks.
\newblock In \emph{ICML}, 2020.

\bibitem[Gasse et~al.(2019)Gasse, Ch{\'{e}}telat, Ferroni, Charlin, and
  Lodi]{Gas+2019}
M.~Gasse, D.~Ch{\'{e}}telat, N.~Ferroni, L.~Charlin, and A.~Lodi.
\newblock Exact combinatorial optimization with graph convolutional neural
  networks.
\newblock In \emph{NeurIPS}, 2019.

\bibitem[Geerts and Reutter(2022)]{geerts2022}
F.~Geerts and J.~L. Reutter.
\newblock Expressiveness and approximation properties of graph neural networks.
\newblock In \emph{ICLR}, 2022.

\bibitem[Gilmer et~al.(2017)Gilmer, Schoenholz, Riley, Vinyals, and
  Dahl]{Gil+2017}
J.~Gilmer, S.~S. Schoenholz, P.~F. Riley, O.~Vinyals, and G.~E. Dahl.
\newblock Neural message passing for quantum chemistry.
\newblock In \emph{ICLR}, 2017.

\bibitem[Goldreich(2010)]{Gol2010}
O.~Goldreich.
\newblock Introduction to testing graph properties.
\newblock In \emph{Property Testing}. Springer, 2010.

\bibitem[Gori et~al.(2005)Gori, Monfardini, and Scarselli]{gori05}
M.~Gori, G.~Monfardini, and F.~Scarselli.
\newblock A new model for learning in graph domains.
\newblock In \emph{IJCNN}, 2005.

\bibitem[Grohe(2017)]{Gro2017}
M.~Grohe.
\newblock \emph{Descriptive Complexity, Canonisation, and Definable Graph
  Structure Theory}.
\newblock Cambridge University Press, 2017.

\bibitem[Grohe(2021)]{Gro+2021}
M.~Grohe.
\newblock The logic of graph neural networks.
\newblock In \emph{LICS}, 2021.

\bibitem[Gui et~al.(2023)Gui, Liu, Li, Luo, and Ji]{DBLP:conf/nips/GuiLLLJ23}
S.~Gui, M.~Liu, X.~Li, Y.~Luo, and S.~Ji.
\newblock Joint learning of label and environment causal independence for graph
  out-of-distribution generalization.
\newblock In \emph{NeurIPS}, 2023.

\bibitem[Hamilton et~al.(2017)Hamilton, Ying, and Leskovec]{Hamilton2017}
W.~L. Hamilton, R.~Ying, and J.~Leskovec.
\newblock Inductive representation learning on large graphs.
\newblock \emph{NeurIPS}, 2017.

\bibitem[Herbst and Jegelka(2025)]{herbst25higher}
D.~Herbst and S.~Jegelka.
\newblock Higher-order graphon neural networks: Approximation and cut distance.
\newblock In \emph{Int. Conf. on Learning Representations (ICLR)}, 2025.

\bibitem[Hoeffding(1963)]{Hoeffding1963}
W.~Hoeffding.
\newblock Probability inequalities for sums of bounded random variables.
\newblock \emph{Journal of the American Statistical Association}, 58, 1963.

\bibitem[Hoover(1979)]{Hoover79}
D.~N. Hoover.
\newblock Relations on probability spaces and arrays of random variables.
\newblock \emph{Princeton Institute for Advanced Study (preprint)}, 1979.

\bibitem[Hoover(1982)]{Hoover82}
D.~N. Hoover.
\newblock Row-column exchangeability and a generalized model for probability.
\newblock \emph{Exchangeability in Probability and Statistics}, pages 281--291,
  1982.

\bibitem[Jegelka(2022)]{jegelka2022theory}
S.~Jegelka.
\newblock Theory of graph neural networks: Representation and learning.
\newblock \emph{arXiv preprint}, 2022.

\bibitem[Jin et~al.(2018)Jin, Barzilay, and Jaakkola]{Jin2018}
W.~Jin, R.~Barzilay, and T.~Jaakkola.
\newblock Junction tree variational autoencoder for molecular graph generation.
\newblock In \emph{ICML}, 2018.

\bibitem[Jin et~al.(2019)Jin, Yang, Barzilay, and Jaakkola]{Jin2019}
W.~Jin, K.~Yang, R.~Barzilay, and T.~Jaakkola.
\newblock Learning multimodal graph-to-graph translation for molecular
  optimization.
\newblock \emph{ICLR}, 2019.

\bibitem[Ju et~al.(2023)Ju, Li, Sharma, and Zhang]{Ju+2023}
H.~Ju, D.~Li, A.~Sharma, and H.~R. Zhang.
\newblock Generalization in graph neural networks: Improved {PAC-Bayesian}
  bounds on graph diffusion.
\newblock \emph{arXiv preprint}, 2023.

\bibitem[Karczewski et~al.(2024)Karczewski, Souza, and Garg]{Kar+2024}
R.~Karczewski, A.~H. Souza, and V.~Garg.
\newblock On the generalization of equivariant graph neural networks.
\newblock In \emph{ICML}, 2024.

\bibitem[Karpinski and Macintyre(1997)]{Kar+1997}
M.~Karpinski and A.~Macintyre.
\newblock Polynomial bounds for {VC} dimension of sigmoidal and general
  {P}faffian neural networks.
\newblock \emph{Journal of Computer and System Sciences}, 54\penalty0
  (1):\penalty0 169--176, 1997.

\bibitem[Kawaguchi et~al.(2022)Kawaguchi, Deng, Luh, and Huang]{Kaw+2022}
K.~Kawaguchi, Z.~Deng, K.~Luh, and J.~Huang.
\newblock Robustness implies generalization via data-dependent generalization
  bounds.
\newblock In \emph{ICML}, 2022.

\bibitem[Keriven et~al.(2020)Keriven, Bietti, and
  Vaiter]{keriven2020convergence}
N.~Keriven, A.~Bietti, and S.~Vaiter.
\newblock Convergence and stability of graph convolutional networks on large
  random graphs.
\newblock In \emph{NeurIPS}, 2020.

\bibitem[Khovanskiǐ(1991)]{khovanskii1991fewnomials}
A.~Khovanskiǐ.
\newblock \emph{Fewnomials}, volume~88 of \emph{Translations of Mathematical
  Monographs}.
\newblock American Mathematical Society, 1991.

\bibitem[Kingma and Ba(2015)]{Kin+2015}
D.~P. Kingma and J.~Ba.
\newblock Adam: {A} method for stochastic optimization.
\newblock In \emph{ICLR}, 2015.

\bibitem[Kipf and Welling(2017)]{Kip+2017}
T.~N. Kipf and M.~Welling.
\newblock Semi-supervised classification with graph convolutional networks.
\newblock In \emph{ICLR}, 2017.

\bibitem[Koltchinskii(2001)]{DBLP:journals/tit/Koltchinskii01}
V.~Koltchinskii.
\newblock Rademacher penalties and structural risk minimization.
\newblock \emph{{IEEE} Transaction on Information Theory}, 47\penalty0
  (5):\penalty0 1902--1914, 2001.

\bibitem[Kriege et~al.(2018)Kriege, Morris, Rey, and Sohler]{Kri+2018}
N.~M. Kriege, C.~Morris, A.~Rey, and C.~Sohler.
\newblock A property testing framework for the theoretical expressivity of
  graph kernels.
\newblock In \emph{IJCAI}, 2018.

\bibitem[Kriege et~al.(2020)Kriege, Johansson, and Morris]{Kri+2019}
N.~M. Kriege, F.~D. Johansson, and C.~Morris.
\newblock A survey on graph kernels.
\newblock \emph{Applied Network Science}, 5\penalty0 (1):\penalty0 6, 2020.

\bibitem[Lam et~al.(2023)Lam, Sanchez-Gonzalez, Willson, Wirnsberger,
  Fortunato, Alet, Ravuri, Ewalds, Eaton-Rosen, Hu, Merose, Hoyer, Holland,
  Vinyals, Stott, Pritzel, Mohamed, and Battaglia]{Lam+2023}
R.~Lam, A.~Sanchez-Gonzalez, M.~Willson, P.~Wirnsberger, M.~Fortunato, F.~Alet,
  S.~Ravuri, T.~Ewalds, Z.~Eaton-Rosen, W.~Hu, A.~Merose, S.~Hoyer, G.~Holland,
  O.~Vinyals, J.~Stott, A.~Pritzel, S.~Mohamed, and P.~Battaglia.
\newblock Learning skillful medium-range global weather forecasting.
\newblock \emph{Science}, 382\penalty0 (6677):\penalty0 1416--1421, 2023.

\bibitem[Langford and Shawe{-}Taylor(2002)]{DBLP:conf/nips/LangfordS02}
J.~Langford and J.~Shawe{-}Taylor.
\newblock {PAC-Bayes} {\&} margins.
\newblock In S.~Becker, S.~Thrun, and K.~Obermayer, editors, \emph{NIPS}, 2002.

\bibitem[Le and Jegelka(2023)]{le23transfer}
T.~Le and S.~Jegelka.
\newblock Limits, approximation and size transferability for gnns on sparse
  graphs via graphops.
\newblock In \emph{NeurIPS}, 2023.

\bibitem[Lee et~al.(2024)Lee, Hwang, and Whang]{DBLP:conf/icml/LeeHW24}
J.~Lee, M.~Hwang, and J.~J. Whang.
\newblock Pac-bayesian generalization bounds for knowledge graph representation
  learning.
\newblock In \emph{ICML}, 2024.

\bibitem[Levie(2023)]{Lev+2023}
R.~Levie.
\newblock A graphon-signal analysis of graph neural networks.
\newblock In \emph{NeurIPS}, 2023.

\bibitem[Levie et~al.(2019)Levie, Isufi, and
  Kutyniok]{DBLP:journals/corr/abs-1901-10524}
R.~Levie, E.~Isufi, and G.~Kutyniok.
\newblock On the transferability of spectral graph filters.
\newblock \emph{arXiv preprint}, 2019.

\bibitem[{Levie} et~al.(2021){Levie}, Huang, Bucci, Bronstein, and
  Kutyniok]{Levie+2022}
R.~{Levie}, W.~Huang, L.~Bucci, M.~Bronstein, and G.~Kutyniok.
\newblock Transferability of spectral graph convolutional neural networks.
\newblock \emph{Journal of Machine Learning Research}, 22\penalty0
  (272):\penalty0 1--59, 2021.

\bibitem[Li et~al.(2022{\natexlab{a}})Li, Wang, Liu, Chen, and
  Xiong]{DBLP:conf/icml/LiWLCX22}
H.~Li, M.~Wang, S.~Liu, P.~Chen, and J.~Xiong.
\newblock Generalization guarantee of training graph convolutional networks
  with graph topology sampling.
\newblock In \emph{ICML}, 2022{\natexlab{a}}.

\bibitem[Li et~al.(2022{\natexlab{b}})Li, Wang, Zhang, and
  Zhu]{DBLP:journals/corr/abs-2202-07987}
H.~Li, X.~Wang, Z.~Zhang, and W.~Zhu.
\newblock Out-of-distribution generalization on graphs: {A} survey.
\newblock \emph{arXiv preprint}, 2022{\natexlab{b}}.

\bibitem[Li et~al.(2024{\natexlab{a}})Li, Geerts, Kim, and Wang]{Li+2024}
S.~Li, F.~Geerts, D.~Kim, and Q.~Wang.
\newblock Towards bridging generalization and expressivity of graph neural
  networks.
\newblock \emph{arXiv preprint}, 2024{\natexlab{a}}.

\bibitem[Li et~al.(2024{\natexlab{b}})Li, Gui, Luo, and
  Ji]{DBLP:conf/icml/LiGLJ24}
X.~Li, S.~Gui, Y.~Luo, and S.~Ji.
\newblock Graph structure extrapolation for out-of-distribution generalization.
\newblock In \emph{ICML}, 2024{\natexlab{b}}.

\bibitem[Liao et~al.(2019)Liao, Li, Song, Wang, Hamilton, Duvenaud, Urtasun,
  and Zemel]{Liao+2019}
R.~Liao, Y.~Li, Y.~Song, S.~Wang, W.~L. Hamilton, D.~Duvenaud, R.~Urtasun, and
  R.~S. Zemel.
\newblock Efficient graph generation with graph recurrent attention networks.
\newblock In \emph{NeurIPS}, 2019.

\bibitem[Liao et~al.(2021)Liao, Urtasun, and Zemel]{Lia+2021}
R.~Liao, R.~Urtasun, and R.~S. Zemel.
\newblock A {PAC-Bayesian} approach to generalization bounds for graph neural
  networks.
\newblock In \emph{ICML}, 2021.

\bibitem[Lin(2004)]{Lin2004}
Y.~Lin.
\newblock A note on margin-based loss functions in classification.
\newblock \emph{Statistics and Probability Letters}, 68, 2004.

\bibitem[Louizos et~al.(2017)Louizos, Shalit, Mooij, Sontag, Zemel, and
  Welling]{DBLP:conf/nips/LouizosSMSZW17}
C.~Louizos, U.~Shalit, J.~M. Mooij, D.~A. Sontag, R.~S. Zemel, and M.~Welling.
\newblock Causal effect inference with deep latent-variable models.
\newblock In \emph{NeurIPS}, 2017.

\bibitem[Lov{\'a}sz and Szegedy(2007)]{Szemeredi_analyst}
L.~M. Lov{\'a}sz and B.~Szegedy.
\newblock Szemer{\'e}di’s lemma for the analyst.
\newblock \emph{GAFA Geometric And Functional Analysis}, 2007.

\bibitem[Lovász(2012)]{Lov+2012}
L.~Lovász.
\newblock \emph{Large Networks and Graph Limits}.
\newblock American Mathematical Society, 2012.

\bibitem[Lv(2021)]{DBLP:journals/corr/abs-2102-10234}
S.~Lv.
\newblock Generalization bounds for graph convolutional neural networks via
  rademacher complexity.
\newblock \emph{arXiv preprint}, 2021.

\bibitem[Ma et~al.(2021)Ma, Deng, and Mei]{Ma2021}
J.~Ma, J.~Deng, and Q.~Mei.
\newblock Subgroup generalization and fairness of graph neural networks.
\newblock In \emph{NeurIPS}, 2021.

\bibitem[Maehara and NT(2019)]{Mae+2019}
T.~Maehara and H.~NT.
\newblock A simple proof of the universality of invariant/equivariant graph
  neural networks.
\newblock \emph{arXiv preprint}, 2019.

\bibitem[Maron et~al.(2019)Maron, Ben{-}Hamu, Shamir, and Lipman]{Mar+2019c}
H.~Maron, H.~Ben{-}Hamu, N.~Shamir, and Y.~Lipman.
\newblock Invariant and equivariant graph networks.
\newblock In \emph{ICLR}, 2019.

\bibitem[Maskey et~al.(2022)Maskey, Lee, Levie, and Kutyniok]{Mas+2022}
S.~Maskey, Y.~Lee, R.~Levie, and G.~Kutyniok.
\newblock Generalization analysis of message passing neural networks on large
  random graphs.
\newblock In \emph{NeurIPS}, 2022.

\bibitem[Maskey et~al.(2023)Maskey, Levie, and
  Kutyniok]{maskey2021transferability}
S.~Maskey, R.~Levie, and G.~Kutyniok.
\newblock Transferability of graph neural networks: An extended graphon
  approach.
\newblock \emph{Applied and Computational Harmonic Analysis}, 63:\penalty0
  48--83, 2023.

\bibitem[Maskey et~al.(2024)Maskey, Kutyniok, and
  Levie]{maskey2024generalization}
S.~Maskey, G.~Kutyniok, and R.~Levie.
\newblock Generalization bounds for message passing networks on mixture of
  graphons.
\newblock \emph{arXiv preprint}, 2024.

\bibitem[McAllester(1999)]{DBLP:conf/colt/McAllester99}
D.~A. McAllester.
\newblock {PAC-Bayesian} model averaging.
\newblock In \emph{COLT}, 1999.

\bibitem[McAllester(2003)]{DBLP:conf/colt/McAllester03}
D.~A. McAllester.
\newblock Simplified pac-bayesian margin bounds.
\newblock In \emph{COLT}, 2003.

\bibitem[McDiarmid(1989)]{McDiarmid1989}
C.~McDiarmid.
\newblock On the method of bounded differences.
\newblock In \emph{Surveys in Combinatorics 1989}, pages 148--188. Cambridge
  University Press, Cambridge, 1989.

\bibitem[Merkwirth and Lengauer(2005)]{merkwirth05}
C.~Merkwirth and T.~Lengauer.
\newblock Automatic generation of complementary descriptors with molecular
  graph networks.
\newblock \emph{Journal of Chemical Information and Modeling}, 45\penalty0
  (5):\penalty0 1159--1168, 2005.

\bibitem[Mohri et~al.(2018)Mohri, Rostamizadeh, and
  Talwalkar]{MohriRostamizadehTalwalkar18}
M.~Mohri, A.~Rostamizadeh, and A.~Talwalkar.
\newblock \emph{Foundations of Machine Learning}.
\newblock The MIT Press, Cambridge, MA, 2018.

\bibitem[Morris et~al.(2019{\natexlab{a}})Morris, Ritzert, Fey, Hamilton,
  Lenssen, Rattan, and Grohe]{Mor+2019}
C.~Morris, M.~Ritzert, M.~Fey, W.~L. Hamilton, J.~E. Lenssen, G.~Rattan, and
  M.~Grohe.
\newblock Weisfeiler and {L}eman go neural: Higher-order graph neural networks.
\newblock In \emph{{AAAI}}, 2019{\natexlab{a}}.

\bibitem[Morris et~al.(2019{\natexlab{b}})Morris, Ritzert, Fey, Hamilton,
  Lenssen, Rattan, and Grohe]{Morris2019}
C.~Morris, M.~Ritzert, M.~Fey, W.~L. Hamilton, J.~E. Lenssen, G.~Rattan, and
  M.~Grohe.
\newblock Weisfeiler and {Leman} go neural: Higher-order graph neural networks.
\newblock \emph{AAAI}, 2019{\natexlab{b}}.

\bibitem[Morris et~al.(2020{\natexlab{a}})Morris, Kriege, Bause, Kersting,
  Mutzel, and Neumann]{Mor+2020}
C.~Morris, N.~M. Kriege, F.~Bause, K.~Kersting, P.~Mutzel, and M.~Neumann.
\newblock {TUDataset:} {A} collection of benchmark datasets for learning with
  graphs.
\newblock \emph{arXiv preprint}, 2020{\natexlab{a}}.

\bibitem[Morris et~al.(2020{\natexlab{b}})Morris, Rattan, and
  Mutzel]{Morris2020b}
C.~Morris, G.~Rattan, and P.~Mutzel.
\newblock {Weisfeiler and Leman} go sparse: Towards higher-order graph
  embeddings.
\newblock In \emph{NeurIPS}, 2020{\natexlab{b}}.

\bibitem[Morris et~al.(2021)Morris, L., Maron, Rieck, Kriege, Grohe, Fey, and
  Borgwardt]{Mor+2022}
C.~Morris, Y.~L., H.~Maron, B.~Rieck, N.~M. Kriege, M.~Grohe, M.~Fey, and
  K.~Borgwardt.
\newblock {Weisfeiler and Leman} go machine learning: The story so far.
\newblock \emph{arXiv preprint}, 2021.

\bibitem[Morris et~al.(2023)Morris, Geerts, T{\"{o}}nshoff, and
  Grohe]{Mor+2023}
C.~Morris, F.~Geerts, J.~T{\"{o}}nshoff, and M.~Grohe.
\newblock {WL} meet {VC}.
\newblock In \emph{ICML}, 2023.

\bibitem[Morris et~al.(2024)Morris, Frasca, Dym, Maron, Ceylan, Levie, Lim,
  Bronstein, Grohe, and Jegelka]{Mor+2024b}
C.~Morris, F.~Frasca, N.~Dym, H.~Maron, {\.I}.~{\.I}. Ceylan, R.~Levie, D.~Lim,
  M.~M. Bronstein, M.~Grohe, and S.~Jegelka.
\newblock Future directions in foundations of graph machine learning.
\newblock \emph{arXiv preprint}, 2024.

\bibitem[Mukherjee et~al.(2006)Mukherjee, Niyogi, Poggio, and
  Rifkin]{Mukherjee2006}
S.~Mukherjee, P.~Niyogi, T.~Poggio, and R.~Rifkin.
\newblock Learning theory: Stability is sufficient for generalization and
  necessary and sufficient for consistency of empirical risk minimization.
\newblock \emph{Advances in Computational Mathematics}, 25, 2006.

\bibitem[Neyshabur et~al.(2018)Neyshabur, Bhojanapalli, and
  Srebro]{Neyshabur2018}
B.~Neyshabur, S.~Bhojanapalli, and N.~Srebro.
\newblock A {PAC-Bayesian} approach to spectrally-normalized margin bounds for
  neural networks.
\newblock \emph{ICLR}, 2018.

\bibitem[Oono and Suzuki(2020)]{Oono2020}
K.~Oono and T.~Suzuki.
\newblock Optimization and generalization analysis of transduction through
  gradient boosting and application to multi-scale graph neural networks.
\newblock In \emph{NeurIPS}, 2020.

\bibitem[Pechyony(2008)]{DBLP:phd/il/Pechyony08}
D.~Pechyony.
\newblock \emph{Theory and practice of transductive learning}.
\newblock PhD thesis, Technion - Israel Institute of Technology, Israel, 2008.

\bibitem[Pellizzoni et~al.(2024)Pellizzoni, Schulz, Chen, and
  Borgwardt]{Pel+2024}
P.~Pellizzoni, T.~Schulz, D.~Chen, and K.~M. Borgwardt.
\newblock On the expressivity and sample complexity of node-individualized
  graph neural networks.
\newblock In \emph{NeurIPS}, 2024.

\bibitem[Qian et~al.(2023)Qian, Ch{\'{e}}telat, and Morris]{Qia+2023}
C.~Qian, D.~Ch{\'{e}}telat, and C.~Morris.
\newblock Exploring the power of graph neural networks in solving linear
  optimization problems.
\newblock \emph{arXiv preprint}, 2023.

\bibitem[Rauchwerger et~al.(2024)Rauchwerger, Jegelka, and Levie]{Rac+2024}
L.~Rauchwerger, S.~Jegelka, and R.~Levie.
\newblock Generalization, expressivity, and universality of graph neural
  networks on attributed graphs.
\newblock \emph{arXiv preprint}, 2024.

\bibitem[Ruiz et~al.(2020)Ruiz, Chamon, and Ribeiro]{Ruisz+2020}
L.~Ruiz, L.~F.~O. Chamon, and A.~Ribeiro.
\newblock Graphon neural networks and the transferability of graph neural
  networks.
\newblock In \emph{NeurIPS}, 2020.

\bibitem[Ruiz et~al.(2021{\natexlab{a}})Ruiz, Chamon, and Ribeiro]{RuizI}
L.~Ruiz, L.~F.~O. Chamon, and A.~Ribeiro.
\newblock Graphon signal processing.
\newblock \emph{IEEE Transactions on Signal Processing}, pages 4961--4976,
  2021{\natexlab{a}}.

\bibitem[Ruiz et~al.(2021{\natexlab{b}})Ruiz, Wang, and Ribeiro]{Ruisz21}
L.~Ruiz, Z.~Wang, and A.~Ribeiro.
\newblock Graphon and graph neural network stability.
\newblock In \emph{ICASSP}, 2021{\natexlab{b}}.

\bibitem[Scarselli et~al.(2009)Scarselli, Gori, Tsoi, Hagenbuchner, and
  Monfardini]{Sca+2009}
F.~Scarselli, M.~Gori, A.~C. Tsoi, M.~Hagenbuchner, and G.~Monfardini.
\newblock The graph neural network model.
\newblock \emph{IEEE Transactions on Neural Networks}, 20\penalty0
  (1):\penalty0 61--80, 2009.

\bibitem[Scarselli et~al.(2018)Scarselli, Tsoi, and
  Hagenbuchner]{scarselli2018vapnik}
F.~Scarselli, A.~C. Tsoi, and M.~Hagenbuchner.
\newblock The {V}apnik--{C}hervonenkis dimension of graph and recursive neural
  networks.
\newblock \emph{Neural Networks}, 108:\penalty0 248--259, 2018.

\bibitem[Shi et~al.(2024)Shi, Pan, Hu, and Dokmanić]{Shi2024}
C.~Shi, L.~Pan, H.~Hu, and I.~Dokmanić.
\newblock Homophily modulates double descent generalization in graph
  convolution networks.
\newblock \emph{Proceedings of the National Academy of Sciences of the United
  States of America}, 121, 2024.

\bibitem[Sun and Lin(2024)]{DBLP:journals/corr/abs-2402-04038}
T.~Sun and J.~Lin.
\newblock Pac-bayesian adversarially robust generalization bounds for graph
  neural network.
\newblock \emph{arXiv}, 2024.

\bibitem[Suzuki et~al.(2023)Suzuki, Nitanda, Suzuki, Wang, Tian, and
  Yamanishi]{DBLP:conf/icml/SuzukiNS00Y23}
A.~Suzuki, A.~Nitanda, T.~Suzuki, J.~Wang, F.~Tian, and K.~Yamanishi.
\newblock Tight and fast generalization error bound of graph embedding in
  metric space.
\newblock In \emph{ICML}, 2023.

\bibitem[Tang and Liu(2023)]{Tan+2023}
H.~Tang and Y.~Liu.
\newblock Towards understanding generalization of graph neural networks.
\newblock In \emph{ICML}, 2023.

\bibitem[Vapnik(1998)]{Vap+1998}
V.~Vapnik.
\newblock \emph{Statistical learning theory}.
\newblock Wiley, 1998.

\bibitem[Vapnik(2006)]{DBLP:books/sp/Vapnik06}
V.~Vapnik.
\newblock \emph{Estimation of Dependences Based on Empirical Data, Second
  Editiontion}.
\newblock Springer, 2006.

\bibitem[Vapnik(1995)]{Vap+95}
V.~N. Vapnik.
\newblock \emph{The Nature of Statistical Learning Theory}.
\newblock Springer, 1995.

\bibitem[Vapnik and Chervonenkis(1964)]{Vap+1964}
V.~N. Vapnik and A.~Chervonenkis.
\newblock A note on one class of perceptrons.
\newblock \emph{Avtomatika i Telemekhanika}, 24\penalty0 (6):\penalty0
  937–945, 1964.

\bibitem[Vasileiou et~al.(2024)Vasileiou, Finkelshtein, Geerts, Levie, and
  Morris]{Vas+2024}
A.~Vasileiou, B.~Finkelshtein, F.~Geerts, R.~Levie, and C.~Morris.
\newblock Covered forest: Fine-grained generalization analysis of graph neural
  networks.
\newblock \emph{arXiv preprint}, 2024.

\bibitem[Verma and Zhang(2019)]{Ver+2019}
S.~Verma and Z.~Zhang.
\newblock Stability and generalization of graph convolutional neural networks.
\newblock In \emph{KDD}, 2019.

\bibitem[von Luxburg and Schölkopf(2011)]{Luxburg2011}
U.~von Luxburg and B.~Schölkopf.
\newblock \emph{Statistical Learning Theory: Models, Concepts, and Results},
  volume~10.
\newblock Handbook of the History of Logic, 2011.

\bibitem[Weisfeiler(1976)]{Wei+1976}
B.~Weisfeiler.
\newblock \emph{On Construction and Identification of Graphs}.
\newblock Springer, 1976.

\bibitem[Weisfeiler and Leman(1968)]{Wei+1968}
B.~Weisfeiler and A.~Leman.
\newblock The reduction of a graph to canonical form and the algebra which
  appears therein.
\newblock \emph{Nauchno-Technicheskaya Informatsia}, 2\penalty0 (9):\penalty0
  12--16, 1968.

\bibitem[Wong et~al.(2023)Wong, Zheng, Valeri, Donghia, Anahtar, Omori, Li,
  Cubillos-Ruiz, Krishnan, Jin, Manson, Friedrichs, Helbig, Hajian, Fiejtek,
  Wagner, Soutter, Earl, Stokes, Renner, and Collins]{Won+2023}
F.~Wong, E.~J. Zheng, J.~A. Valeri, N.~M. Donghia, M.~N. Anahtar, S.~Omori,
  A.~Li, A.~Cubillos-Ruiz, A.~Krishnan, W.~Jin, A.~L. Manson, J.~Friedrichs,
  R.~Helbig, B.~Hajian, D.~K. Fiejtek, F.~F. Wagner, H.~H. Soutter, A.~M. Earl,
  J.~M. Stokes, L.~D. Renner, and J.~J. Collins.
\newblock Discovery of a structural class of antibiotics with explainable deep
  learning.
\newblock \emph{Nature}, 2023.

\bibitem[Wu et~al.(2023{\natexlab{a}})Wu, Sun, Zhang, Xie, and
  Cui]{DBLP:journals/csur/WuSZXC23}
S.~Wu, F.~Sun, W.~Zhang, X.~Xie, and B.~Cui.
\newblock Graph neural networks in recommender systems: {A} survey.
\newblock \emph{{ACM} Computing Surveys}, 55\penalty0 (5):\penalty0
  97:1--97:37, 2023{\natexlab{a}}.

\bibitem[Wu et~al.(2023{\natexlab{b}})Wu, Bojchevski, and Huang]{Wu2023}
Y.~Wu, A.~Bojchevski, and H.~Huang.
\newblock Adversarial weight perturbation improves generalization in graph
  neural networks.
\newblock \emph{AAAI}, 2023{\natexlab{b}}.

\bibitem[Xu and Mannor(2012)]{Xu+2012}
H.~Xu and S.~Mannor.
\newblock Robustness and generalization.
\newblock \emph{Machine Learning}, 86\penalty0 (3):\penalty0 391--423, 2012.

\bibitem[Xu et~al.(2018)Xu, Li, Tian, Sonobe, Kawarabayashi, and
  Jegelka]{Xu+2018}
K.~Xu, C.~Li, Y.~Tian, T.~Sonobe, K.~Kawarabayashi, and S.~Jegelka.
\newblock Representation learning on graphs with jumping knowledge networks.
\newblock In \emph{ICLR}, 2018.

\bibitem[Xu et~al.(2019{\natexlab{a}})Xu, Hu, Leskovec, and Jegelka]{Xu+2018b}
K.~Xu, W.~Hu, J.~Leskovec, and S.~Jegelka.
\newblock How powerful are graph neural networks?
\newblock In \emph{ICLR}, 2019{\natexlab{a}}.

\bibitem[Xu et~al.(2019{\natexlab{b}})Xu, Jegelka, Hu, and Leskovec]{Xu2019}
K.~Xu, S.~Jegelka, W.~Hu, and J.~Leskovec.
\newblock How powerful are graph neural networks?
\newblock \emph{ICLR}, 2019{\natexlab{b}}.

\bibitem[Xu et~al.(2020)Xu, Li, Zhang, Du, Kawarabayashi, and
  Jegelka]{xu20reasoning}
K.~Xu, J.~Li, M.~Zhang, S.~Du, K.~Kawarabayashi, and S.~Jegelka.
\newblock What can neural networks reason about?
\newblock In \emph{ICLR}, 2020.

\bibitem[Xu et~al.(2021)Xu, Zhang, Li, Du, Kawarabayashi, and
  Jegelka]{xu2021how}
K.~Xu, M.~Zhang, J.~Li, S.~S. Du, K.-I. Kawarabayashi, and S.~Jegelka.
\newblock How neural networks extrapolate: From feedforward to graph neural
  networks.
\newblock In \emph{ICLR}, 2021.

\bibitem[Ye et~al.(2021)Ye, Xie, Cai, Li, Li, and
  Wang]{DBLP:conf/nips/YeXCLLW21}
H.~Ye, C.~Xie, T.~Cai, R.~Li, Z.~Li, and L.~Wang.
\newblock Towards a theoretical framework of out-of-distribution
  generalization.
\newblock In \emph{NeurIPS}, 2021.

\bibitem[Yehudai et~al.(2021)Yehudai, Fetaya, Meirom, Chechik, and
  Maron]{Yeh+2021}
G.~Yehudai, E.~Fetaya, E.~A. Meirom, G.~Chechik, and H.~Maron.
\newblock From local structures to size generalization in graph neural
  networks.
\newblock In \emph{ICML}, 2021.

\bibitem[Zhang et~al.(2023)Zhang, Feng, Du, He, and Wang]{Zha+2023b}
B.~Zhang, G.~Feng, Y.~Du, D.~He, and L.~Wang.
\newblock A complete expressiveness hierarchy for subgraph {GNNs} via subgraph
  {W}eisfeiler-{L}ehman tests.
\newblock \emph{arXiv preprint}, 2023.

\bibitem[Zhang et~al.(2020)Zhang, Wang, Liu, Chen, and Xiong]{Zhang2020}
S.~Zhang, M.~Wang, S.~Liu, P.~Y. Chen, and J.~Xiong.
\newblock Fast learning of graph neural networks with guaranteed
  generalizability: One-hidden-layer case.
\newblock In \emph{ICML}, 2020.

\bibitem[Zhou and Wang(2021)]{Zhou2021}
X.~Zhou and H.~Wang.
\newblock The generalization error of graph convolutional networks may enlarge
  with more layers.
\newblock \emph{Neurocomputing}, 424, 2021.

\end{thebibliography}

\appendix
\addcontentsline{toc}{section}{Appendix}
\addtocontents{toc}{\protect\setcounter{tocdepth}{-1}}

\section{The \texorpdfstring{$1$}{1}-dimensional Weisfeiler--Leman algorithm}\label{subsec:1WL} The \wlone{} or \new{color refinement} is a well-studied heuristic for the graph isomorphism problem, originally proposed by~\citet{Wei+1968}.\footnote{Strictly speaking, the \wlone{} and color refinement are two different algorithms. That is, the \wlone{} considers neighbors and non-neighbors to update the coloring, resulting in a slightly higher expressive power when distinguishing vertices in a given graph; see~\cite {Gro+2021} for details. Following customs in the machine learning literature, we consider both algorithms to be equivalent.} Intuitively, the algorithm determines if two graphs are non-isomorphic by iteratively coloring or labeling vertices. Given an initial coloring or labeling of the vertices of both graphs, e.g., their degree or application-specific information, in each iteration, two vertices with the same label get different labels if the number of identically labeled neighbors is unequal. These labels induce a vertex partition, and the algorithm terminates when, after some iteration, the algorithm does not refine the current partition, i.e., when a \new{stable coloring} or \new{stable partition} is obtained. Then, if the number of vertices annotated with a specific label is different in both graphs, we can conclude that the two graphs are not isomorphic. It is easy to see that the algorithm cannot distinguish all non-isomorphic graphs~\citep{Cai+1992}. However, it is a powerful heuristic that can successfully decide isomorphism for a broad class of graphs~\citep{Arv+2015,Bab+1979}.

Formally, let $G = (V(G),E(G),\ell)$ be a labeled graph. In each iteration, $t > 0$, the \wlone{} computes a vertex coloring $C^1_t \colon V(G) \to \Nb$, depending on the coloring of the neighbors. That is, in iteration $t>0$, we set
\begin{equation*}
	C^1_t(v) \coloneq \REL\Big(\!\big(C^1_{t-1}(v),\oms C^1_{t-1}(u) \mid u \in N(v)  \cms \big)\! \Big),
\end{equation*}
for all vertices $v \in V(G)$, where $\REL$ injectively maps the above pair to a unique natural number, which has not been used in previous iterations. In iteration $0$, the coloring $C^1_{0}\coloneq \ell$ is used. To test whether two graphs $G$ and $H$ are non-isomorphic, we run the above algorithm in ``parallel'' on both graphs. If the two graphs have a different number of vertices colored $c \in \Nb$ at some iteration, the \wlone{} \new{distinguishes} the graphs as non-isomorphic. Moreover, if the number of colors between two iterations, $t$ and $(t+1)$, does not change, i.e., the cardinalities of the images of $C^1_{t}$ and $C^1_{i+t}$ are equal, or, equivalently,
\begin{equation*}
	C^1_{t}(v) = C^1_{t}(w) \iff C^1_{t+1}(v) = C^1_{t+1}(w),
\end{equation*}
for all vertices $v$ and $w$ in $V(G\,\dot\cup H)$, then the algorithm terminates. For such $t$, we define the \new{stable coloring} $C^1_{\infty}(v) = C^1_t(v)$, for $v \in V(G\,\dot\cup H)$. The stable coloring is reached after at most $\max \{ |V(G)|,|V(H)| \}$ iterations~\citep{Gro2017}.  We define the \emph{color complexity}
of a graph $G$ as the number of colors computed by the \wlone{} after $|V(G)|$ iterations on $G$.

\section{Pfaffian functions}
\label{sec:pfaff}
\emph{Pfaffian functions} have been first introduce in \citet{khovanskii1991fewnomials} to extend Bezout’s theorem, stating that the number of complex solutions of a set of polynomial equations can be estimated based on their degree. For different works exploiting properties of Pfaffian functions, see \citep{Kar+1997, DBLP:journals/tnn/BianchiniS14}.

A Pfaffian chain of order $\ell \geq0 $ and degree $\alpha \geq 1$, in an open domain $U \subset \mathbb{R}^n$, is a sequence of analytic functions $f_1,f_2,,\ldots,f_{\ell}$ over $U$, satisfying the differential equations
\begin{equation*}
	\frac{d f_j(\vec{x})}{d x_i} = \sum_{1 \leq i \leq n} g_{ij}(\vec{x},f_1(\vec{x}),\ldots,f_j(\vec{x})), \ \ \ 1\leq j \leq \ell.
\end{equation*}
Here, $g_{ij}(\vec{x},y_1,\ldots,y_j)$ are polynomials in $\vec{x} \in U$ and $y_1,\ldots y_j \in \mathbb{R}$ of degree not exceeding $\alpha$. A function $f(\vec{x}) = P(\vec{x},f_1(\vec{x}),\ldots,f_{\ell}(\vec{x}))$ where $P(\vec{x},y_1,\ldots,y_j)$ is a polynomial of degree not exceeding $\beta$, is called a \emph{Pfaffian function of format} $(\alpha, \beta, \ell)$. Most of the functions with continuous derivatives are \emph{Pfaffian functions}. In particular, the \emph{arcatangent}, \emph{logistic sigmoid}, and the \emph{hyperbolic tangent} are \emph{Pfaffian functions}.

\section{Graph convolutional neural networks}
\label{sec:kGCNs}
Let $G \in \mathcal{G}_{n,d}$ be an unlabeled graph with adjacency matrix $\vec{A}$, Laplacian $\vec{L}$, and node features $\vec{x}_v \in \mathbb{R}^d$, for $v \in V(G)$. For the $K$-class graph classification problem, we use the following matrix notation of an $L$-layer GCN,
\begin{align*}
	H_k = \sigma(\vec{L}\vec{H}_{k-1}\vec{W}_k)  \quad       & \text{($k$-th Graph Convolution Layer)} \\
	H_L = \frac{1}{n} \vec{1}_n \vec{H}_{L-1}\vec{W}_L \quad & \text{(Readout Layer)},
\end{align*}
where, $k \in [L-1]$, $\vec{H}_k \in \mathbb{R}^{n \times h_k}$, $\vec{1}_n \in \mathbb{R}^{1 \times n}$ is the all-one vector, $\vec{W}_j$ is the weight matrix of the $j$-th layer, and $\vec{H}_0 \in \mathbb{R}^{n \times d}$ is the initial node representation matrix, having rows equal to the initial node features $\vec{x}_u$, for $u \in V(G)$.

\section{Extended graphon-theory}
\label{sec:extended_graphon-theory}

Graphons are utilized to analyze the generalization capabilities of MPNNs, either as models of the data distribution or as analytic tools for generic data distributions. In the former approach, graphs are assumed to be randomly sampled from graphons. In the latter, the non-complete space of graphs is embedded into the compact metric space of graphons, leveraging compactness to derive generalization bounds or universal approximation results for arbitrary data distributions. In what follows, we present the most fundamental properties of graphons and their relevance to graph theory. For simplicity, we focus on the main results concerning the space of graphons rather than graphon signals. However, when necessary, we reference works that extend these results to graphon signals.

\subsection{Basics on graphon theory}

Here, we review the basics on graphon theory.

\paragraph{Definitions} We begin by recalling the definition of a graphon, which is a symmetric, measurable function $W \colon [0,1]^2 \to [0,1]$. The set of all graphons is denoted by $\cW$. Graphons generalize graphs in the following way. Every graph $G$ of order $n$ can be represented as a graphon $W_G$. This is achieved by partitioning $[0,1]$ into $n$ intervals $(I_v)_{v \in V(G)}$, each of measure $\sfrac{1}{n}$, and defining $W_G(x, y)$ for $x \in I_u$ and $y \in I_v$ to be one if $\{u, v\} \in E(G)$ and zero otherwise. We refer to $W_G$ as the graphon induced by the graph $G$.

\paragraph{Message-passing graph neural networks on graphons}
For a graphon $W \in \cW$, an $L$-layer MPNN initializes a feature $\vec{h}_x^{(0)} \coloneq \varphi_0 \in \Rb^{d_0}$, for $x \in [0,1]$. Then, for each layer $t \in [L]$, the features $\vec{h}_x^{(t)} \colon [0,1] \to \Rb^{d_t}$ are computed iteratively. After $L$ layers, a single graphon-level feature $\vec{h}_W \in \Rb^{1 \times d}$ is computed as follows,
\begin{equation*}
	\label{def:graphonmpnns}
	\vec{h}_x^{(t)} \coloneq \varphi_t \left( \int_{[0,1]} W(x,y) \vec{h}_y^{(t-1)} \, d\lambda(y) \right),
	\quad \text{and} \quad
	\vec{h}_W \coloneq \psi \left( \int_{[0,1]} \vec{h}_x^{(L)} \, d\lambda(x) \right).
\end{equation*}
Here, $(\varphi_t)_{t=1}^L$ and $\psi$ are the functions defined in \cref{def:unlabeledmpnnsgraphs}. This definition generalizes architectures following \cref{def:unlabeledmpnnsgraphs} to graphons. Specifically, we have the following result:

\begin{lemma}[{\citep[Theorem 9]{Boe+2023}}]
	\label{mpnnsgraphonsismpnngraphs}
	Let $G \in \cG_n$ for some $n \in \Nb$, let $(\varphi_t)_{t=1}^L$ define an $L$-layer MPNN model, and let $\psi$ be Lipschitz. Then,
	\begin{equation*}
		\vec{h}_G = \vec{h}_{W_G}.\tag*{\qed}
	\end{equation*}
\end{lemma}

The above definition of MPNNs (and \cref{mpnnsgraphonsismpnngraphs}) can be extended to graphon-signals $(W, \vec{f})$ by initializing $\vec{h}_x^{(0)} = \vec{f}(x)$ for $x \in [0,1]$, as described in \citet{Rac+2024}.

\subsection{Graphons as generative graph models} A graphon can serve as a generative model of graphs. Given a graphon $W$, a random graph of $n$ nodes is generated by sampling uniformly and independently points $\{X_i\}_{i=1}^n$ from  $[0,1]$, and connecting each pair $X_i,X_j$ in probability $W(X_i,X_j)$ to obtain the edges of the graph. Most works consider graphs as generative models of graphs to model the data distributions. These include papers that study \emph{transferability}. Here, one studies a setting where an MPNN is trained on a given graph, assumed to be sampled from a graphon, and the goal is to show that the repercussion of the MPNN on another graph, sampled from the same graphon, is close to the repercussion on the original graph. This aims at modeling the ability to transfer MPNNs between different graphs of the same ``type'' \citep{Levie+2022,RuizI,keriven2020convergence, maskey2021transferability,Ruisz21}. Other works \citep{Mas+2022,maskey2024generalization} use graphons to model the data distribution in graph classification or regression settings. Here, one assumes that there is a finite set of template graphons, such that each graph from the dataset is sampled independently from one of these graphons; see \citep{maskey2024generalization} for more details.

\subsection{Graphons as the completion of the space of graphs}
When analyzing properties like generalization or universal approximation of a model, one usually has to choose a metric or a topology for the input space. The compactness of the input space is a requirement both for using the Stone--Weierstrass theorem for universality analysis and for proving generalization bounds via a covering number analysis. In the context of graphs, when the size or degree $n$  is not assumed to be bounded from above, the space of graphs is typically non-complete (and hence non-compact) concerning standard graph metrics. A standard tool for obtaining a complete space from a separable space is \emph{completion}. For example, the real numbers---a complete space---is the completion of the rationals---a non-complete space. In the context of graphs, as explained above, the space of graphons is the completion of the space of graphs concerning the cut distance. Hence, the universality and generalization analysis are developed for the \say{nice} compact space of graphons, and the results are then also valid for graphs as a special case. Such an analysis is presented in \citet{Lev+2023} for generalization and in \citet{Boe+2023} for expressivity and universal approximation analysis.

\paragraph{Cut distance}

For any $W\in L^2[0,1]^2$, we define the cut norm of $W$ as
\[\|W\|_{\square} =\sup_{U,V\subset [0,1] } \Big|\int_{U\times V} W(x,y)dx dy\Big|\]
where $U,V\subset [0,1]$ are measurable.
The cut metric between two graphons $W,W'\in\mathcal{W}$ is defined to be
\[d_{\square}(W,W') = \norm{W-W'}_{\square}.\]
This is interpreted as the difference between the average edge weights of $W$ and $W'$ on the cut $U\times V$ that maximizes this difference.

Let $W'=W_{G'}$ be an SBM induced from a graph of $m$ nodes, and $W_G$ induced from a graph of $n=jm$ nodes, where $m,j\in\mathbb{N}$. In this case, $d_{\square}(W_G,W_{G'})$ measures how much the statistics/densities of the edge weights of $G$ behave like the expected number of edges of graphs randomly sampled from the SBM $W_{G'}$.
Hence, the cut metric has a pseudo-probabilistic interpretation in this case. How much the deterministic graph $G$ looks like it was randomly sampled from the SBM $G'$. This interpretation can be extended to any pair of graphs $G,G'$, where now $d_{\square}(W_G,W_{G'})$ is interpreted as how much $G$ and $G'$ look like they were sampled from the same SBM.

The cut metric is sensitive to the nodes' specific ``ordering'' in $[0,1]$. To obtain a ``permutation invariant'' metric, one defines the \emph{cut distance}
\[\delta_{\square}(W',W) = \inf_{\psi \in S_{[0,1]}} d_{\square}(W',W^{\psi}), \]
where $S_{[0,1]}$ is the space of all measure preserving bijections -- the analogue to permutations in general measure spaces.

The cut distance is a pseudo-metric.
By introducing an equivalence relation $W\sim W'$ iff $\delta_{\square}(W,W')=0$, and the quotient space $\widetilde{\mathcal{W}}:=\mathcal{W}/\sim$,  the space  $\widetilde{\mathcal{W}}$ is a metric space with a metric $\delta_{\square}$
defined by  $\delta_{\square}([W],[W'])=\delta_{\square}(W,W')$ where $W\in[W]$ and $W'\in[W']$ are two arbitrary representatives.  By abuse of terminology, we call elements of $\widetilde{\mathcal{W}}$ also graphons.

\paragraph{The weak regularity lemma}

One can extend the above interpretation of cut metric to distance between general graphons and SBMs. Namely, $d_{\square}(W,W_{G_n})$ quantifies how much the graphon $W$ looks like it was randomly sampled from the SBM $G_n$ of $n$ nodes. The \emph{weak regularity lemma} \citep{weakReg,Szemeredi_analyst} states that for every graphon $W$ there is some SBM $W_{G_n}$ such that $d_{\square}(W,W_{G_n}) \leq \frac{1}{\sqrt{c\log(n)}}$, where $c>0$ is a universal constant. This means that the number of required blocks $n^2$ for approximating graphons by SBMs depends only on the desired error tolerance and not on any property of the graphon $W$. Hence, any graph looks like it was sampled from a SBM, where the number of blocks of the SBM does not depend on the graphon.

As a corollary of the weak regularity lemma, one can show that $\widetilde{\mathcal{W}}$ is a compact space undercut distance \citep{Szemeredi_analyst}. Moreover, it follows that the space of graphs, embedded in $\widetilde{\mathcal{W}}$ via induced graphons, is dense in $\widetilde{\mathcal{W}}$. Hence, completing the space of graphs with cut distance is the compact space $\widetilde{\mathcal{W}}$.

\section{Tree Mover's distance}
\label{sec:TMD}

The following introduces the \new{Tree Mover’s Distance} (TMD), defined through the optimal transport problem. Following the notation from the original paper by \citet{chuang2022tree}, we introduce the optimal transport problem and the Wasserstein distance. We then define an optimal transport problem between the sets of unrolled computation trees corresponding to the vertices of two attributed graphs. Next, we define a distance between labeled rooted trees and extend this definition to the space of attributed graphs, using the optimal transport problem and the unrolling trees of graphs' vertices.

\paragraph{The optimal transport problem} We begin with a brief introduction to \new{optimal transport} (OT) and the \new{Wasserstein distance}. Let $ X = \{x_i\}_{i=1}^m $ and $ Y = \{y_j\}_{j=1}^m $ be two multisets of $m$ elements each. Let $\vec{C} \in \Rb^{m \times m} $ be the transportation cost for each pair, i.e., $ C_{ij} = d(x_i, y_j) $, where $d$ is a distance function between $X$ and $Y$. The Wasserstein distance is defined through the following minimization problem,
\begin{equation*}
	\text{OT}^*_d(X, Y) \coloneq \min_{\vec{\gamma} \in \Gamma(X, Y)} \frac{\langle \vec{C}, \vec{\gamma} \rangle}{m}, \quad \Gamma(X, Y) = \{\gamma \in \Rb^{m \times m}_+ \mid \vec{\gamma} \mathbf{1}_m = \vec{\gamma}^\top \mathbf{1}_m = \mathbf{1}_m \},
\end{equation*}
where $\langle \cdot, \cdot \rangle$ is defined as the sum of the elements resulting from the entry-wise multiplication of two matrices with equal dimensions, and $\mathbf{1}_m$ denotes the all ones $m$-dimensional column vector.
In the following, we use the unnormalized version of the Wasserstein distance,
\begin{equation*}
	\text{OT}_d(X, Y)  \coloneq \min_{\vec{\gamma} \in \Gamma(X, Y)} \langle \vec{C}, \vec{\gamma} \rangle = m \cdot \text{OT}^*_d(X, Y).
\end{equation*}

\paragraph{Distance between rooted trees via hierarchical OT}
Let $ (T,r)$ denote a rooted tree. We further let $ \cT_v $ be the multiset of subtrees of $T$ consisting of subtrees rooted at the children of $v$. Determining whether two trees are similar requires iteratively examining whether the subtrees in each level are similar. By recursively computing the optimal transportation cost between their subtrees, we define the distance between two rooted trees $(T_a,r_a), (T_b,r_b)$. However, the number of subtrees could be different for $r_a$ and $r_b$, i.e., $|\cT_{r_a}| \neq |\cT_{r_b}|$. To compute the OT between sets with different sizes, we augment the smaller set with \new{blank trees}.

\begin{definition}
	A blank tree $T_0$ is a tree (graph) that contains a single vertex and no edge, where the vertex feature is the zero vector $\mathbf{0}_p \in \Rb^p$, and $ T_0^n $ denotes a multiset of $n$ blank trees.
\end{definition}

\begin{definition}
	Given two multisets of trees $\cT_u, \cT_v$, define $\rho$ to be a function that augments a pair of trees with blank trees as follows,
	\begin{equation*}
		\rho \colon (\cT_v, \cT_u) \to \mleft( \cT_v \cup T_0^{\max(|\cT_u| - |\cT_v|, 0)}, \, \cT_u \cup T_0^{\max(|\cT_v| - |\cT_u|, 0)} \mright).
	\end{equation*}
\end{definition}

We now recursively define a distance between rooted trees using the optimal transport problem.

\begin{definition}
	The distance between two trees rooted $(T_a,r_a), (T_b,r_b)$ is defined recursively as:
	\begin{equation*}
		\text{TD}_\omega(T_a, T_b)  \coloneq
		\begin{cases}
			\|\vec{x}_{r_a} - \vec{x}_{r_b}\| + \omega(L) \cdot \text{OT}_{\text{TD}_w}(\rho(T_{r_a}, T_{r_b})) & \text{if } L > 1, \\
			\|\vec{x}_{r_a} -
			\vec{x}_{r_b}\|                                                                                     & \text{otherwise},
		\end{cases}
	\end{equation*}
	where $\norm{\cdot}$ is a norm on the feature space, $ L = \max(\text{Depth}(T_a), \text{Depth}(T_b)) $ and $ \omega \colon \mathbb{N} \to \mathbb{R}^+ $ is a depth-dependent weighting function.
\end{definition}

\paragraph{Extension to attributed graphs}
Below, we extend the above distance between rooted trees to the TMD on attributed graphs by calculating the optimal transportation cost between the graphs’ unrolling trees.

\begin{definition}
	Given two graphs $G, H$ and $L>0$, the tree mover’s distance between $G$ and $H$ is defined as
	\begin{equation*}
		\text{TMD}_\omega^L(G, H) = \text{OT}_{\text{TD}_\omega}(\rho(\cT^L_{G}, \cT^L_{H})),
	\end{equation*}
	where $\cT^L_{G}$ and $\cT^L_{H}$ are multisets of the depth-$L$ unrolling trees of graphs $G$ and $H$, respectively.
\end{definition}

For completeness below we recall the defintion of the depth $L$ unrolling of a labeled graph $(G,\ell_G)$ following \citet{Morris2020b}
\paragraph{Graph unrollings}
Given an $n$-order labeled graph $(G,\ell_G)$, we define the \new{unrolling tree} of depth $L \in \Nb_0$ for a vertex $u \in V(G)$, denoted as $\text{UNR}(G,u,L)$, inductively as follows.
\begin{enumerate}
	\item For $L=0$, we consider the trivial tree as an isolated vertex labeled $\ell_G(u)$.
	\item For $L>0$, we consider the root vertex with label $\ell_G(u)$ and, for $v \in N(u)$, we attach the subtree $\text{UNR}(G,v,L-1)$ under the root.
\end{enumerate}
Note that above unrolling tree construction characterizes the $\wlone$ algorithm through the following lemma implying that the TMD distance is equivalent to $\wlone$ algorithm in terms of distinguishing non-isomorphic graphs.
\begin{lemma}[Folklore, see, e.g.,~\citet{Mor+2020}]
	\label{wlone:unrollingschar}
	The following are equivalent for $L \in \Nb_0$, given a labeled graph $(G,\ell_G)$ and vertices $u,v \in V(G)$.
	\begin{enumerate}
		\item The vertices $u$ and $v$ have the same color after $L$ iterations of the \wlone.
		\item The unrolling trees $\text{UNR}(G,u,L)$ and $\text{UNR}(G,v,L)$ are isomorphic.
	\end{enumerate}
\end{lemma}

\end{document}